\documentclass{article}
\usepackage{iclr2027_conference,times}
\usepackage[T1]{fontenc}
\usepackage{amsmath,amssymb,booktabs,graphicx,subcaption}
\usepackage{url}

\usepackage[english]{babel}

\usepackage{amsthm}
\usepackage{enumitem}
\usepackage{makecell}

\usepackage{url}
\usepackage{graphicx}
\usepackage{xcolor}

\usepackage{float}
\usepackage{soul}
\usepackage[ruled,vlined,linesnumbered]{algorithm2e}
\SetAlCapNameFnt{\small}
\SetAlCapFnt{\small}
\usepackage[font=small]{caption}
\definecolor{LightRed}{HTML}{f8e8e8}
\sethlcolor{LightRed}

\newcommand{\rmv}[1]{}

\expandafter\def\expandafter\normalsize\expandafter{%
    \normalsize% % Display equations
    \setlength{\abovedisplayskip}{3pt}%
    \setlength{\belowdisplayskip}{3pt}%
    \setlength{\abovedisplayshortskip}{1.5pt}%
    \setlength{\belowdisplayshortskip}{1.5pt}%
}

\usepackage{titlesec}
\usepackage{titletoc}
\usepackage{hyperref}
\titlespacing{\section}{0pt}{3pt}{1pt}
\titlespacing{\subsection}{0pt}{2pt}{1pt}
\titlespacing{\subsubsection}{0pt}{1pt}{1pt}

\title{Learning Chaos Without Seeing Chaos:\\
Extrapolation of Global Dynamics\\ in Autoregressive Transformers}
\author{Yilun Liu$^{1,3}$ \quad
Yi Zhang$^{2}$ \quad
Ganyu Wu$^{1,3}$ \quad
Sikuan Yan$^{1,3}$ \quad
Mengyue Wang${^2}$\\
\textbf{Alois Knoll}$^{2}$ \quad
\textbf{Volker Tresp}$^{1,3}$ \quad
\textbf{Yunpu Ma}$^{1,3}$\\
$^1$Ludwig Maximilian University of Munich \quad
$^2$Technical University of Munich\\
$^3$Munich Center for Machine Learning\\
\texttt{yilun.liu@campus.lmu.de}}
\iclrfinalcopy % Uncomment for camera-ready version, but NOT for submission.
\begin{document}

\maketitle

\begin{abstract}
Autoregressive models are trained to predict a system's behavior one step at a time, and recursive generation allows the learned dynamics to unfold over long horizons.
To what extent can such dynamics learned from local observations recover broader organization of an underlying system that was only partially observed during training?
Here we study small autoregressive transformers trained from scratch on trajectories sampled from restricted parameter regimes of several non-linear dynamical systems, including logistic and sine maps, the Lorenz system, and the generalized Hopf system, with control parameters and state trajectories represented as sequences of continuous tokens.
Under closed-loop evaluation at parameters far outside the training distribution, the models can recover self-similar period-doubling cascades, chaotic dynamics, and attractor structures with remarkable visual and numerical fidelity.
For the logistic map, a transformer reproduces successive period doublings up to period 128, yielding a finite-order scaling ratio of 4.6687, matching the Feigenbaum constant to within 5$\times$10\textsuperscript{-4}.
We further investigate how these structures emerge over the course of training, and reveal with causal interventions how control-parameter information is processed through attention into state prediction and shapes the resulting closed-loop dynamics.
These results suggest that a surprisingly narrow window into a system's local behavior may suffice for autoregressive transformers to generalize to its unseen global dynamical organization.
\end{abstract}

\section{Introduction}
Autoregressive models of dynamical systems learn locally from partial observations by construction.
They are trained to predict the next observation from those available so far, often on trajectories that cover only a restricted range of system conditions.
From this limited evidence, the model is expected to internalize an account of how the underlying system actually evolves.
At inference, each prediction is fed back as the next input, turning into an autonomous closed-loop dynamical system whose behavior can unfold horizons and parameter regimes that were never represented during training.
There is no guarantee that these unfolded dynamics remain faithful to the underlying system.
Recursive generation shifts the input distribution from observed states to model-generated ones~\citep{bengio2015scheduled,ross2011reduction}, while extrapolation in the parameter distributions requires extending the learned transition beyond the conditions constraining it during training.

Can a model trained only on local observations give rise to parameter-dependent closed-loop operators that are sufficiently accurate and robust to preserve aspects of ground-truth global organization of the underlying dynamics when extrapolated beyond the regime it has seen?
Prior work confirms that learned dynamical models can reproduce unseen behavior in some parameterized systems~\citep{pathak2018,vlachas2018,koglmayr2024,kim2021global,huh2025,vantegelen2025},
while recent results with transformers occupy an ambiguous position~\citep{srivastava2023beyond, casert2024}.
Dedicated state-space function approximators provide a direct interface for learning transitions or vector fields~\citep{narendra1990identification,chen2018neural}.
Here, we ask whether a small autoregressive transformer can realize through generic sequence computation over separate continuous parameter and state tokens a parameter-conditioned transition that, when recursively applied, give rise to the correct dynamical organization absent from the training regime.

Bifurcations provide a particularly informative test of this question.
Unlike ordinary out-of-distribution (OOD) prediction, extrapolating across a bifurcation requires the learned dynamics to cross a qualitative change in long-term behavior, where the trivial, stable fixed-point trajectories give way to periodic orbits and eventually chaos.
In the quadratic universality class, successive stable cycles of periods $2,4,8,\ldots$ emerge through nested sequences of bifurcations, whose parameter intervals asymptotically approach the Feigenbaum scaling ratio~\citep{feigenbaum1978}.
Such properties give us increasingly stringent notions of extrapolation, whether a model trained only on stable-point trajectories simply mimics the superficial pattern, whether it captures a coherent sequence of period doublings, whether it places those transitions correctly, and ultimately whether their relative scaling follows the same universality structure.

We experiment with small decoder-only transformers trained autoregressively from scratch on trajectories sampled only from restricted parameter regimes of several nonlinear systems, including the logistic and sine maps with period-doubling structures, and the Lorenz systems and generalized Hopf systems, which provides multidimensional chaotic attractors and multistable behaviors.
Control parameter scalars and the observed states are represented directly as independent continuous tokens, and training uses only local next-state prediction, without any other labels, stability annotations, governing equations, or long-time dynamical summaries.
At test time, we recursively apply the learned transition at parameter values far outside the training range and examine the resulting closed-loop dynamics.

\begin{figure}[t!]
\centering
\begin{subfigure}[b]{.25\textwidth}
\centering
\includegraphics[height=2.45cm]{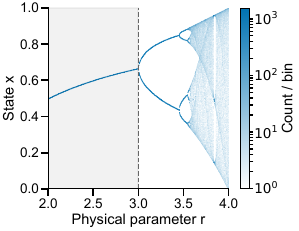}
\caption{Ground truth}
\end{subfigure}\hfill
\begin{subfigure}[b]{.71\textwidth}
\centering
\includegraphics[height=2.45cm]{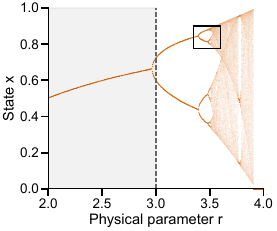}\hfill
\includegraphics[height=2.45cm]{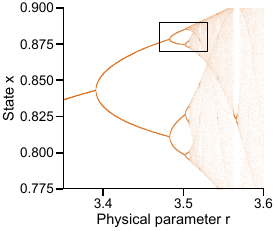}\hfill
\includegraphics[height=2.45cm]{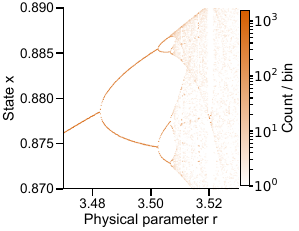}
\caption{Closed-loop dynamics of an autoregressive transformer}
\end{subfigure}
\vspace{-1ex}
\caption{\textbf{Autoregressive Transformer trained only on pre-bifurcation trajectories of the logistic map recovers self-similar period-doubling structure beyond the domain it was trained on.}
The shaded areas mark the regime $[2,3]$ used for training.
Closed-loop results are generated by transformer of seed 42 trained after 10,000 steps with one state token, where the boxes mark zoomed-in areas in the successive panels.
Quantitative comparisons with the reference are detailed in Section~\ref{sec:verified-cascade} and Appendices~\ref{app:cascade-results} and~\ref{app:results}.
}%; the sine counterpart is shown in Appendix~\ref{app:training}.}
\label{fig:overview}
\vspace{-1ex}
\end{figure}

Across these systems, autoregressive transformers extrapolate qualitatively correct global dynamics far beyond the parameter regimes observed during training.
For one-dimensional maps, models trained before the onset of period doubling generate corresponding cascades and chaotic behavior; in the logistic family, seven successive flips up to period $128$ are numerically verified, with scaling ratio closely matching the reference Feigenbaum constant.
The cascade persists under narrow and substantially shifted training windows and disappears when the correspondence between parameters and trajectories is disrupted, linking the extrapolated behavior to learned parameter-conditioned dynamics.
Beyond period doubling, models on the Lorenz and generalized Hopf system also successfully recover unseen attractor behavior.
Causal interventions on attention further reveal how control-parameter information shapes the learned dynamics.
Together, these results suggest that local next-state supervision within trivial dynamics of a system can give rise to autonomous dynamics that preserve nontrivial global organization well beyond the training distribution.

\section{Background and Preliminaries}
\label{sec:background}
\subsection{From system identification to learned scientific models}

Learning scientific dynamics from data targets different objects.
Classical neural system identification seeks learning nonlinear transition laws directly from observations~\citep{narendra1990identification}, while automated scientific-discovery methods seek explicit symbolic descriptions of the underlying system, including symbolic reconstruction of nonlinear dynamics~\citep{bongard2007automated}, free-form natural-law discovery~\citep{schmidt2009distilling}, sparse identification of governing equations~\citep{brunton2016discovering}, equation-learning networks~\citep{sahoo2018learning}, and physics-informed symbolic regression~\citep{udrescu2020aifeynman}.
More recent approaches use learned representations as intermediate scientific models, for example by extracting symbolic relations from neural networks~\citep{cranmer2020symbolic} or by designing interpretable function approximators~\citep{liu2025kan}.
A complementary family of methods learns predictive models without requiring an explicit symbolic law.
Koopman-inspired models seek representations in which nonlinear evolution becomes approximately linear~\citep{lusch2018deep}; neural differential equations parameterize continuous-time dynamics directly~\citep{chen2018neural}; and recurrent or reservoir models can reproduce autonomous attractor-level behavior from observed trajectories~\citep{lu2018attractor,pathak2018}.
Related ideas appear in scientific surrogate modeling, where reduced-order models and neural operators learn parameterized families of physical solutions~\citep{hesthaven2018nonintrusive,li2021fourier}, and in \emph{world models}, where learned predictive dynamics support internal simulation and planning~\citep{ha2018worldmodels}.
These approaches differ in what they aim to recover: an explicit governing law, a useful representation, a solution operator, or a predictive dynamical surrogate.
Our setting concerns rather than whether the exact governing equation is identified, but whether a generic learned predictor can preserve structural consequences of that equation outside the conditions from which it was learned.

\subsection{Parameterized dynamical systems and bifurcations}
We consider a family of dynamical systems whose evolution depends on control parameter $r\in\mathcal{R}$.
For a discrete-time system, we have
\begin{equation}
 x_{t+1}=f_r(x_t),
 \label{eq:discrete}
\end{equation}
where $x_t\in\mathcal{X}$ is the state.
For a continuous-time system
\begin{equation}
\dot x=v_r(x)
\end{equation}
we denote by $\varPhi_r^{\Delta t}$ its flow map over an observation interval $\Delta t$, such that
\begin{equation}
 x_{t+\Delta t}=\varPhi_r^{\Delta t}(x_t).
 \label{eq:flowmap}
\end{equation}
This allows discrete maps and discretely observed flows be viewed through a common notion of repeated state transition, and we denote $F_r$ for either form when the distinction is unimportant.

Given a learned predictor $g_\theta$, recursive application likewise induces a parameterized autonomous dynamical system, which we denote by $F_{\theta,r}$ when the learned state is Markovian.
Our analysis concerns selected properties of the dynamics $F_{\theta,r}$ induces under iteration.
We examine \emph{structural extrapolation} by asking whether the learned system $F_{\theta,r}$ preserves aspects of the organization of $F_r$ outside the parameter support observed during training, including properties of the qualitative or quantitative organization of the induced dynamics, such as attractor type, periodicity, bifurcation structure, scaling relations, chaotic instability, etc.

Bifurcations are qualitative changes in the dynamics of a parameterized system as a control parameter varies~\citep{kuznetsov2004}. The relevant stability object depends on the system and invariant set under consideration.
For a scalar discrete map $F_r$, let $x$ belong to an orbit of minimal period-$p$  $x,F_r(x),\ldots,F_r^{p-1}(x)$, so that the orbit and its cycle multiplier $\Lambda_p(r,x)$ satisfy
\begin{equation}
    F_r^p(x)=x,\qquad
 \Lambda_p(r,x)=\left(F_r^p\right)'(x)=\prod_{j=0}^{p-1}F_r'\!\left(F_r^j(x)\right).
 \label{eq:multiplier}
\end{equation}
The orbit is locally asymptotically stable when $|\Lambda_p|<1$.
Under the standard transversality and nondegeneracy conditions, a flip, or period-doubling, bifurcation occurs when the multiplier passes through \(-1\), creating a nearby period-\(2p\) branch.
We combine orbit closure, the multiplier condition, and two-sided stability continuation jointly to quantitatively verify such transitions numerically.

Successive flip bifurcations can organize into a period-doubling cascade.
Let $r_1,r_2,\cdots$ denotes successive bifurcation parameters along a principal cascade, with attracting periods \(1,2,4,\ldots\).
For smooth unimodal maps in the quadratic universality class, the finite-order ratios of successive parameter intervals $\{\delta_n\}$
converge asymptotically to the Feigenbaum constant
\begin{equation}
    \delta=\lim_{n\rightarrow\infty}\delta_n\approx4.6692016091,\qquad
 \delta_n=\frac{r_n-r_{n-1}}{r_{n+1}-r_n},
 \label{eq:ratio}
\end{equation}
under the usual assumptions defining this universality class~\citep{feigenbaum1978}.
This universality is a statement about the asymptotic organization of the cascade, with different maps in the same universality class having different bifurcation locations \(r_n\) while sharing the same limiting scaling ratio.
We distinguish this throughout the paper between bifurcation placement, measured by the physical values of \(r_n\), from cascade scaling, measured by the finite-order ratios \(\delta_n\).

The continuous-time experiments involve other forms of bifurcation mechanisms.
For example, a generic Hopf bifurcation occurs when a complex-conjugate pair of Jacobian eigenvalues crosses the imaginary axis under the corresponding transversality and nondegeneracy conditions~\citep{kuznetsov2004}. We elaborate system-specific stability and attractor diagnostics for the flow experiments in the repsective sections.

\subsection{Learning dynamics across parameter-conditioned regimes}

Data-driven models can reproduce autonomous behavior that extends beyond short-horizon trajectory prediction.
Recurrent and reservoir models can reproduce long-run or attractor-level behavior of chaotic systems~\citep{vlachas2018,pathak2018,lu2018attractor,gauthier2021}, while parameterized scientific surrogates and neural operators generalize solution maps across physical conditions~\citep{hesthaven2018nonintrusive,li2021fourier}.
These settings motivate studying a learned predictor itself as a dynamical system, but generalization of a solution family does not necessarily imply recovery of qualitative regime transitions.

Closer to our setting are models that explicitly condition evolution on a control parameter.
Recurrent networks can infer control-dependent transformations and bifurcation structure from local examples and extrapolate them beyond the observed control range~\citep{kim2021global}.
Parameter-aware reservoir methods have been used to extrapolate tipping transitions and post-transition dynamics~\citep{koglmayr2024}, and dynamical analysis of such a reservoir shows that the learned autonomous model can itself undergo a corresponding bifurcation near that of the target system~\citep{sisodia2024dynamical}.
Parameterized Neural ODEs similarly extrapolate bifurcation structure through learned parameter-dependent vector fields~\citep{vantegelen2025}.

Specifically, transformers~\citep{vaswani2017} have also been studied under several related formulations.
Autonomously generated trajectories can exhibit emergent behavior at physical conditions absent from training~\citep{casert2024}; other work predicts bifurcation diagrams directly from trajectory observations, making bifurcation structure itself the supervised object~\citep{zhornyak2024}; and recent experiments report failures to extrapolate unseen collapse transitions despite successful parameter-aware reservoir models~\citep{zhai2026transformers}.
These results suggest that regime extrapolation is not guaranteed by architecture or low local prediction error, but is an empirical property of the autonomous dynamics induced by the learned transition.
Beyond behavioral evaluation, causal-intervention methods provide tools for testing whether internal representations or computational pathways mediate model outputs~\citep{geiger2021,meng2022rome}.
We adopt this perspective to investigate how control-parameter information shapes the learned process, such that local predictive signal induces the appropriate dynamical organization globally.

\section{Methodology and Experiments}
\label{sec:method}

\subsection{Systems, Data, and Training Objective}
\label{sec:scalar-training}
We first study two one-parameter families of scalar maps on $x\in[0,1]$,
\begin{equation}
F_r(x)=rx(1-x)
\qquad\text{and}\qquad
F_r(x)=r\sin(\pi x),
\label{eq:maps}
\end{equation}
corresponding to the logistic and sine families.
For the logistic map, training parameters are sampled uniformly from
$\mathcal R_{\mathrm{train}}=[2,3]$ and evaluation covers $2,4$.
For the sine map, training uses $0.1,0.71$ and evaluation covers $0.1,1$.
The first period-doubling bifurcations of the reference maps occur at $r=3$ and $r\approx0.719961683$, respectively, so neither training distribution contains an interval beyond the first period doubling.
For each trajectory, the control parameter and initial state are sampled independently and uniformly.%, and the trajectory is retained from initialization without discarding transients.
%The training data therefore contain local transitions along the approach to the training-regime attractors rather than only samples from their asymptotic states.

Given a trajectory $x_0,\ldots,x_T$ generated at parameter $r$, the models are trained with teacher-forced next-state autoregression,
\begin{equation}
 \mathcal L(\theta)=\mathbb E\left[\frac1T\sum_{t=0}^{T-1}
 \bigl\|g_\theta(r,x_{0:t})-x_{t+1}\bigr\|_2^2\right],
 \label{eq:loss}
\end{equation}
where every prediction is conditioned on the reference prefix $x_{0:t}$ during training, and model-generated states enter only during autonomous closed-loop evaluation.
Sequence lengths are drawn from sampling per batch, and no burn-in samples is removed from the training dataset.
Appendix~\ref{app:protocol} provides the complete sampling, optimization, and checkpoint protocols.

\subsection{Parameterized Transformer and Induced Dynamics}
\label{sec:induced-map}

The control parameter and states are represented as separate continuous tokens. In the scalar setting,
\begin{equation}
 h_r^{(0)}=W_r r+b_r+v_r,\qquad
 h_t^{(0)}=W_x x_t+b_x+v_x,
\end{equation}
where $v_r$ and $v_x$ are learned type embeddings distinguishing parameter from state tokens. These are linear embeddings of continuous numerical values rather than discrete one-hot vocabulary embeddings, and the parameter token(s) is appended preceding all state tokens in the causal sequence.

The model is a decoder-only Transformer with residual-connected layers. Each layer applies causal multi-head self-attention followed by a gated feed-forward block, with $\operatorname{SiLU}(z)=z\sigma(z)$,
\begin{align}
 u^{(\ell)}=h^{(\ell)}+\operatorname{Attn}_{\ell}(h^{(\ell)}),\qquad
 h^{(\ell+1)}=u^{(\ell)}+
 W_o[\operatorname{SiLU}(W_g u^{(\ell)})\odot W_v u^{(\ell)}],
\end{align}
where $\operatorname{SiLU}(z)=z\sigma(z)$. A linear output head predicts the next state at every state-token position.
The configuration for scalar map experiments has hidden width 16, two attention heads, feed-forward width 64, two untied residual layers, and 8,304 trainable parameters.
It uses no positional embeddings, layer normalization, input normalization, nor dropout. Details in Appendix~\ref{app:protocol}.

Because feed-forward computation is purely token-wise, information from the separate parameter token enters state-token computation through attention and can then be combined with state information by subsequent nonlinear transformations.
his representation makes the route of parameter influence accessible to our interventions in Section~\ref{sec:intervention}, linking token-level computation to the induced dynamics.
Appendix~\ref{app:parameter-computation} further discusses the distinction between parameter routing and the arithmetic operations that realize the transition in detail.
%This does not directly imply that attention weights themselves implement the analytic parameter dependence of the physical maps, nor that successful trajectories identify a particular internal arithmetic algorithm; Appendix~\ref{app:parameter-computation} addresses these representational possibilities separately.

At evaluation time the trained parameters $\theta$ are frozen and predictions are returned to the model as inputs. For a context containing at most $C$ state observations,
\begin{equation}
 \hat x_{t+1}=\Pi_{[0,1]}g_\theta\bigl(r,\hat x_{\max(0,t-C+1):t}\bigr),
 \label{eq:rollout}
\end{equation}
where $\Pi_{[0,1]}$ clips the scalar predictions to the physically allowed state interval and the available prefix is used before the context reaches length $C$. For $C=1$, recursive prediction defines a smooth, parameterized neural transition
and the corresponding rollout map
\begin{equation}
F_{\theta,r}(x)=\Pi_{[0,1]}
\widetilde F_{\theta,r}(x), \qquad \widetilde F_{\theta,r}(x)=g_\theta(r,x).
\label{eq:induced-map}
\end{equation}
The local bifurcation calculations below concern periodic orbits in the interior of the state domain, where the clamp is inactive and the derivatives of $F_{\theta,r}$ coincide with those of the smooth neural transition $\widetilde F_{\theta,r}$. This separates the differentiable map used for local stability analysis from the range constraint used during autonomous rollout.

For $C>1$ the generated process is not a one-dimensional map even though each observation is scalar. Once the context is full, define
\begin{equation}
X_t=
(\hat x_{t-C+1},\ldots,\hat x_t)
\in[0,1]^C.
\end{equation}
The induced autonomous update is then the delay-state map
\begin{equation}
 \mathcal F_{\theta,r}(X_t)=\bigl(\hat x_{t-C+2},\ldots,\hat x_t,\Pi_{[0,1]}g_\theta(r,X_t)\bigr).
 \label{eq:delay-map}
\end{equation}
We consequently restrict scalar cycle-multiplier and root-continuation calculations to $C=1$, while trajectory- and distribution-level comparisons are applied to all evaluated contexts.

\subsection{Extension to Continuous-Time Flows}
\label{sec:flow-method}
For continuously evolving systems observed at a fixed interval $\Delta t$, we retain the same separation between parameter and state tokens but predict a finite-time state increment instead of the next state directly.
Denoting the learned increment as $\Delta_\theta$, the resulting discrete transition is
\begin{equation}
\widehat\varPhi_{\theta,r}^{\Delta t}(x) =
x+\Delta_\theta(r,x),
\label{eq:flow-predictor}
\end{equation}
which approximates the physical flow map $\varPhi_r^{\Delta t}$ introduced in Section~\ref{sec:background}. The model is therefore trained on discretely observed flow transitions; the continuous vector field $v_r$ itself is neither supplied nor explicitly reconstructed.
All state, parameter, and increment normalization statistics are estimated from the corresponding training data only, and the flow rollouts are not clipped.

We apply this construction to Lorenz and generalized Hopf systems.
The Lorenz experiments include both a one-parameter setting and a joint setting in which $(\rho,\sigma)$ are embedded in a single parameter token with $\beta=8/3$ fixed; the state token contains the Cartesian state $(x,y,z)$.
In the joint experiment, training parameters are drawn only from trajectories that numerically satisfy a numerical fixed-destination criterion, with their finite-time transients retained.
For the generalized Hopf system, the parameter token contains $(\mu_1,\mu_2)$ and the state token contains Cartesian coordinates $(x,y)$.
Training is restricted to a region in which every sampled nonzero trajectory contracts radially, so no stable periodic attractor appears in the training data.
The detailed parameters, sampling, normalization, and optimization settings are specified in Appendices~\ref{app:lorenz}, \ref{app:generalized-hopf}, and~\ref{app:joint-lorenz}.

\subsection{Intervention on Parameter Conditioning}
\label{sec:intervention}
To test whether parameter information transmitted through attention causally affects the state prediction, we patch activations while holding the state history fixed. Let $H$ denote a fixed state history and let $r$ and $r'$ be recipient and donor parameters, with $y_{\mathrm{rec}}=g_\theta(r,H)$ and $y_{\mathrm{don}}=g_\theta(r',H)$. Let $a_\ell(r,H)$ be the complete attention output at the final state position in layer $\ell$, immediately before its residual addition. Replacing it by $a_\ell(r',H)$ in the recipient computation produces a patched prediction $y_\ell^{\mathrm{patch}}$, and the fraction of the donor effect transferred by the intervention is
\begin{equation}
 R_\ell=1-
 \frac{\mathbb E[(y_\ell^{\mathrm{patch}}-y_{\mathrm{don}})^2]}
 {\mathbb E[(y_{\mathrm{rec}}-y_{\mathrm{don}})^2]}.
 \label{eq:patch}
\end{equation}
Thus $R_\ell=1$ corresponds to recovery of the donor model output and $R_\ell=0$ to no reduction in donor discrepancy relative to the unmodified recipient.
%The donor target is explicitly a counterfactual output of the learned model, not the reference physical transition.
We adopt sham replacement that reuses the recipient activation, and a random control that uses a per-example random direction of equal norm as the donor--recipient activation difference.
We also complement this single-step intervention with closed-loop ones.
In a selected layer, all attention logits from state queries to the parameter key are set to $-\infty$ at every generated step, and softmax is recomputed over the remaining admissible keys.
This removes the selected parameter-to-state attention edge. Appendix~\ref{app:interventions} reports the resulting trajectories together with the native model, sham and equal-norm random controls.

\subsection{Measuring Structural Extrapolation}
\label{sec:measuring}
We evaluate properties of the autonomous system induced by the frozen predictor.
The scalar measurements separately examine long-run state distributions, periodic organization, bifurcation placement, and cascade scaling, along with longitudinal measurements comparing these quantities with local transition accuracy during training.
The multidimensional flow systems use diagnostics appropriate to their higher-dimensional attractors and stability structure. More details in Appendix~\ref{app:measure}.

% Section 4 reports outcomes only. Protocol, metrics and architecture are defined
% in Section~3 and in the appendices, so those definitions are not repeated here.

\begin{figure}[t]
\centering
\begin{subfigure}[t]{.31\textwidth}
\centering
\includegraphics[width=\linewidth]{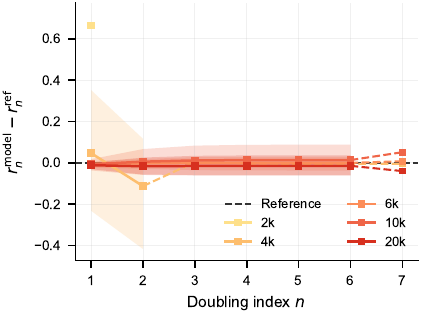}
\caption{Flip parameter displacement}
\end{subfigure}\hfill
\begin{subfigure}[t]{.31\textwidth}
\centering
\includegraphics[width=\linewidth]{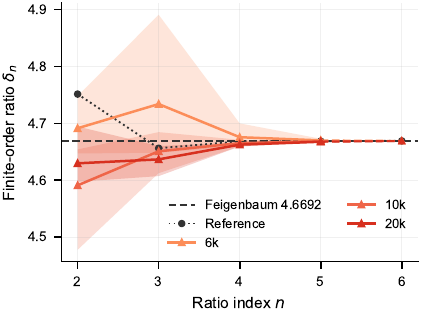}
\caption{Finite-order scaling ratios}
\end{subfigure}\hfill
\begin{subfigure}[t]{.31\textwidth}
\centering
\includegraphics[width=\linewidth]{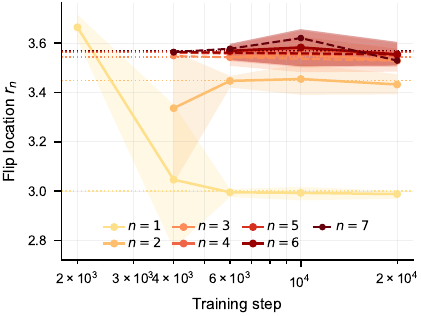}
\caption{Flip parameters during training}
\end{subfigure}
\vspace{-1ex}
\caption{\textbf{Emergence of period-doubling cascades beyond the observed regime can be numerically verified via the reconstructed scaling patterns of period-doubling cascades.}
Results for the Logistic map, averaged across seeds with shaded min--max ranges; solid segments are evidence supported by all seeds, dashed segments only partially supported by those that had reached that level.
$r_n$ is %the parameter value of the $n$-th period-doubling (flip) bifurcation of the one-state logistic map,
verified as detailed in Section~\ref{sec:verification}.}
\label{fig:scaling}
\vspace{-1ex}
\end{figure}

\section{Results and Analyses}
\label{sec:results}

Here we organize results concerning whether and how global structure emerges beyond the observed regime during training, how much observation it requires, what kind of structure is recovered, how parameter information supports it, and whether the same phenomenon extends beyond scalar maps.

\subsection{Emergence of Global Structure Beyond the Observed Regime}
\label{sec:verified-cascade}
The learned logistic maps develop a high-order period-doubling cascade entirely outside the training interval.
At the final checkpoint, up to 7 consecutive flips can be numerically verified, reaching period-$128$.
The deepest finite-order ratio shared across seeds is \(4.668686\), compared with \(4.669132\) for the reference cascade at the same order.
The learned sine model also exhibits a qualitative period-doubling cascade; applying the verification pipeline to the sine reference gives a deepest finite-order ratio of $4.668805$.
These results establish that the learned closed-loop maps can recover the self-similar successive bifurcations and the precise scaling structure organizing the it.
Detailed results are provided in Appendix~\ref{app:cascade-results}, \ref{app:training}, and \ref{app:visual-atlas}.

\begin{figure*}[t]
\centering
\setlength{\tabcolsep}{0pt}
\renewcommand{\arraystretch}{0.8}
\noindent\begin{tabular*}{\textwidth}{@{\extracolsep{\fill}}*{5}{c}@{}}
\footnotesize Init. & \footnotesize 500 & \footnotesize 1k & \footnotesize 2k & \footnotesize 3k \\[-2pt]
\includegraphics[width=0.18\textwidth]{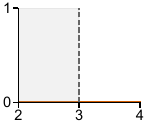} & \includegraphics[width=0.18\textwidth]{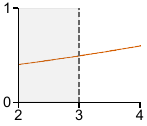} & \includegraphics[width=0.18\textwidth]{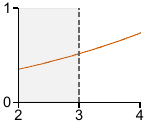} & \includegraphics[width=0.18\textwidth]{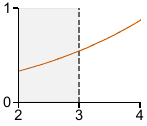} & \includegraphics[width=0.18\textwidth]{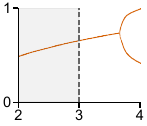} \\[1pt]
\footnotesize 4k & \footnotesize 6k & \footnotesize 10k & \footnotesize 20k & \footnotesize Ground truth \\[-2pt]
\includegraphics[width=0.18\textwidth]{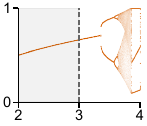} & \includegraphics[width=0.18\textwidth]{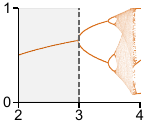} & \includegraphics[width=0.18\textwidth]{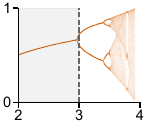} & \includegraphics[width=0.18\textwidth]{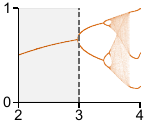} & \includegraphics[width=0.18\textwidth]{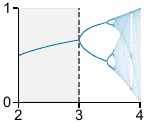} \\[1pt]
\end{tabular*}
\vspace{-2ex}
\caption{\textbf{Emergence of period-doubling cascades beyond the observed regime can be visually examined via the similar patterns reconstructed.}
The shaded areas mark the regime $[2,3]$ used for training.
Results for the Logistic map, with seed 42, $C=1$, on the same 513-point grid used in Figure~\ref{fig:overview}.}
\label{fig:learning-evolution}
\vspace{-1ex}
\end{figure*}

\label{sec:learning}
We can also observe from how the patterns emerge across training steps.
In the logistic model, no doubling can be found by 2,000 updates, until the first one appears at 3,000, three by 4,000, etc.
The sine model develops the same qualitative organization even earlier, with two doublings by 1,000 updates and three by 2,000.
In Appendix~\ref{app:training} we provide additional results showing that how local prediction and closed-loop structures evolve differently, with between 6,000 and 10,000 logistic updates, OOD one-step MSE decreases from \(1.10\times10^{-3}\) to \(2.41\times10^{-4}\), while closed-loop OOD \(W_1\) increases from \(0.0296\) to \(0.0867\).
This further confirms that structural organization cannot be reduced to monotonic improvement in one-step prediction accuracy.
Appendix~\ref{app:cascade-results} and \ref{app:results} report further details about the distributional comparisons, numerical verifications, per-seed measurements, and comparisons with alternative transition representations.

%\subsection{Emergence Survives Severely Restricted Observation Windows}
\label{sec:window}
We experimented varying the parameter interval available during training while keeping the remaining protocol fixed. Training on \([2,3]\), \([2.5,3]\), \([2.9,3]\), or the substantially lower interval \([2,2.5]\) still produces the same qualitative cascade organization.
At least six consecutive flips can be numerically verified under each training window, and the coarse detector identifies a cascade in 11 of the 12 window--seed combinations.
This confirms that emergence of the cascade does not require training observations to be even immediately adjacent to the transition. Additional details are provided in Appendix~\ref{app:window}.

%\subsection{Recovery of Ground Truth Dynamical Organizations}
The closed-loop predictor can reproduce global structural consequences of the underlying dynamics without explicitly recovering the analytic governing equation.
Here we use the recovered Feigenbaum scaling to study a particular structural property of the learned map.
The logistic Transformer has a fitted critical exponent \(z=2.00\), and the sine Transformer \(z\approx2.10\), consistent with a quadratic critical point, which can also be visually examined in cobweb plot results provided in Appendix~\ref{app:cobweb} and detailed in \ref{app:universality}.
Their autonomous cascades therefore exhibit the organization expected of the quadratic universality class, confirming a recovery of the ground truth global dynamical organizations.

\subsection{Parameter Conditioning Causally Shapes the Learned Dynamics}
\label{sec:parameter-route}
Counterfactual activation patching shows that parameter dependence is mediated through learned attention routes.
Replacing the final-query attention output while holding the state history fixed identifies where parameter information enters the state computation.
For the logistic map, a second-layer patch transfers almost the full donor effect ($R_1>0.99999$) whereas a first-layer patch is negligible ($R_0=0.000151$).
The equal-norm random control increases donor error and sham replacement leaves the prediction unchanged.
The dominant mediating layer varies across seeds, confirming that the route is indeed learned during training rather than hard-coded by the architecture.
Closed-loop blocking connects these local interventions to global autonomoous dynamics.
For logistic seed 42 with one state token, blocking the relevant second-layer attention to the parameter token raises OOD \(W_1\) from \(0.03497\) to \(0.60605\) and eliminates the two consecutive flips detected on the intervention grid.
The parameter--trajectory correspondence is also critically required during learning.
When parameter labels are permuted across training examples, all runs lose coarse-grid flips and periodic agreement, with OOD error rising sharply.
The extrapolated organization therefore depends on learning how the control parameter changes the transition itself.
Details in Appendix~\ref{app:interventions}.

\subsection{Structural Extrapolation in Continuous Flows}
\label{sec:flows}

\begin{figure*}[!t]
\centering

% ---------- Lorenz ----------
\begin{minipage}[t]{0.49\textwidth}
\centering
\begin{subfigure}[t]{0.48\linewidth}
    \includegraphics[width=\linewidth]{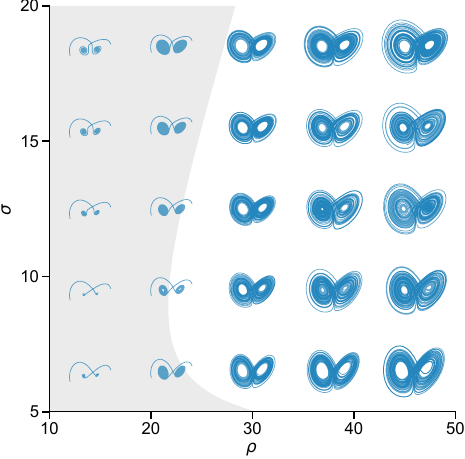}
    \caption{Ground truth}
\end{subfigure}
\hfill
\begin{subfigure}[t]{0.48\linewidth}
    \includegraphics[width=\linewidth]{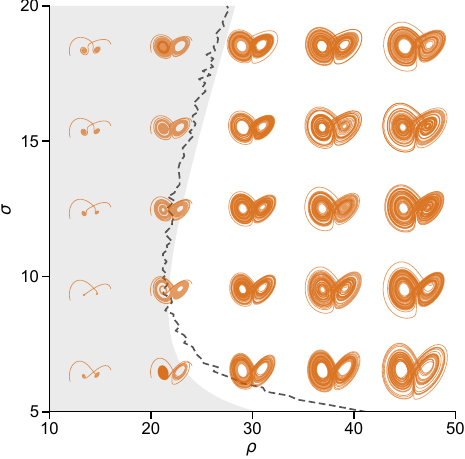}
    \caption{Transformer}
\end{subfigure}
\vspace{-0.5ex}
\caption{
\textbf{Extrapolation in the Lorenz system.}
Training is confined to the shaded fixed-destination region, where trajectories converge to fixed states. Beyond it, the model recovers chaotic double-wing dynamics and the transition boundary between the two regimes, marked as the dashed curve in (b). %Trajectories use matched parameter pairs and initial conditions.
}
\label{fig:lorenz-extension}
\vspace{-1ex}
\end{minipage}
\hfill
% ---------- Generalized Hopf ----------
\begin{minipage}[t]{0.49\textwidth}
\centering
\begin{subfigure}[t]{0.48\linewidth}
    \includegraphics[width=\linewidth]{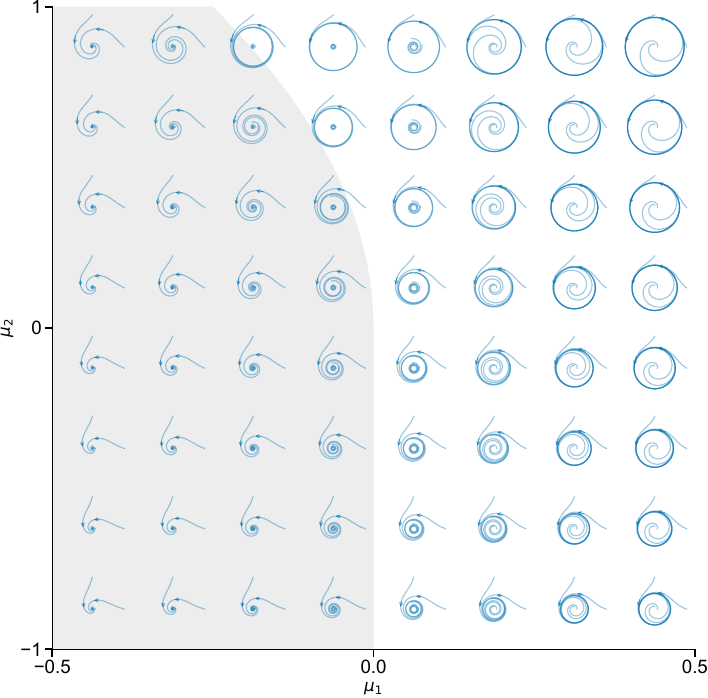}
    \caption{Ground truth}
\end{subfigure}
\hfill
\begin{subfigure}[t]{0.48\linewidth}
    \includegraphics[width=\linewidth]{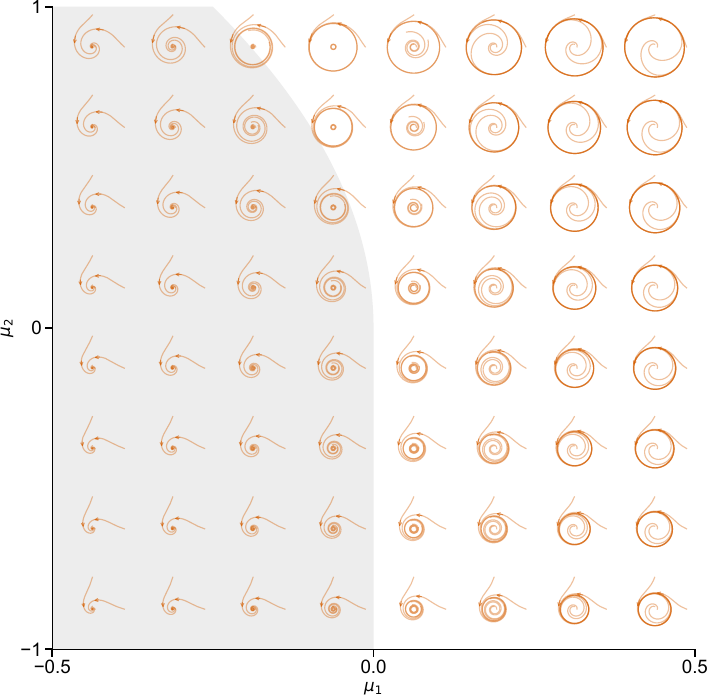}
    \caption{Transformer}
\end{subfigure}

\vspace{-0.5ex}
\caption{
\textbf{Extrapolation in the generalized Hopf system.}
Training is confined to the shaded radial-contraction region, where no stable cycle is observed. Beyond it, the model recovers stable-point, stable-cycle, and coexisting-attractor behavior from matched initial conditions.
}
\label{fig:generalized-hopf}
\vspace{-1ex}
\end{minipage}

\end{figure*}

\label{sec:lorenz-extension}
We next test whether the same token-based formulation extends from scalar maps to coupled multidimensional flows. We consider the Lorenz system
\begin{equation}
\dot x=\sigma(y-x),\qquad
\dot y=x(\rho-z)-y,\qquad
\dot z=xy-\beta z,
\label{eq:lorenz}
\end{equation}
with $\beta=8/3$ fixed and $(\rho,\sigma)$ varied jointly.
The model observes trajectories consisting of Cartesian states $(x,y,z)$ together with a parameter token $(\rho,\sigma)$.
In our experiments, the learned transition generates unseen chaotic dynamics and recovers the organization of fixed and sustained-motion regimes.
%In the joint-parameter Lorenz system, the learned transition generates unseen chaotic dynamics and recovers the organization of fixed and sustained-motion regimes.
%The same parameter-conditioned formulation recovers nontrivial regime organization in a multidimensional chaotic flow.
Across the joint \((\rho,\sigma)\) evaluation region, the model trained only on short observations of fixed-destination trajectories agrees with the reference fixed/nonfixed classification in \(593/625\) cells and reproduces repeated wing switching in \(362/375\) reference cells exhibiting such motion.
Figure~\ref{fig:lorenz-extension} show the corresponding change of regime at matched parameter pairs.
Inside the shaded region that the model has seen during training, trajectories spiral onto a fixed point, and outside they extrapolate into correct sustained double-wing motion whose amplitude grows with $\rho$.
The model also gives an accurate reconstructed boundary, tracking the reference to a median absolute displacement of $0.95$ in $\rho$ with a maximum of $10.7$ only at the steep low-$\sigma$ end.
Appendix~\ref{app:lorenz-conventional} provides details for the boundary and distributional results and the rendering protocols.
The recovered nonfixed dynamics are genuinely unstable rather than merely visually double-winged. At \((\rho,\sigma)=(28,10)\), the learned system has largest Lyapunov exponent \(1.0404\), compared with \(0.9046\) for the reference under the same estimator, and the sign agrees at all ten tested parameter pairs, as detailed in Appendix~\ref{app:flow-lyapunov}.

\label{sec:generalized-hopf}
The two-parameter generalized Hopf normal form~\citep{kuznetsov2004}
\begin{equation}
\dot r=r(\mu_1+\mu_2r^2-r^4),\qquad \dot\theta=1,
\end{equation}
tests a qualitatively different structure: coexistence of attracting states.
The model observes only Cartesian states $(x,y)$ trajectories together with a parameter token $(\mu_1,\mu_2)$.
Training is restricted to a region in which every sampled nonzero trajectory contracts radially, so the model observes only trajectories approaching the origin and never a stable periodic attractor.
Figure~\ref{fig:generalized-hopf} displays the resulting transient geometry on both sides of this curved training boundary.
The model correctly reproduces the reference point-versus-cycle organization across all nine tested parameter pairs.
All $36$ trajectories remain finite through $t=300$ and match the reference long-time outcome, with $14$ approaching fixed points and $22$ approaching cycles. At coexistence parameters, different initial radii converge to different attractors; for example, at $(\mu_1,\mu_2)=(-0.1,1)$ the predicted outer-cycle radius is $0.948$, compared with $0.942$ for the reference. Because each parameter pair is evaluated from multiple initial radii and phases, the coexistence is generated by the learned dynamics rather than inferred from a single trajectory.
Appendix~\ref{app:generalized-hopf} reports additional detailed results. %long-horizon basin tests, parameter slices, and learned stability boundaries.

Taken together, the scalar, Lorenz, and generalized Hopf results point to the same phenomenon across distinct dynamical mechanisms, where local parameter-conditioned prediction can give rise to autonomous systems that recover nontrivial unseen chaotic, periodic, and multistable long-time organization beyond the regimes represented during training.

\section{Discussion}
\label{sec:discussion}

Our experiments reveal a recurring pattern across several nonlinear systems: local next-state learning can organize a learned predictor into an autonomous dynamical system with global structure that was never represented in its training regime. In the scalar maps, the learned systems develop an extended hierarchy of period doublings whose finite-order scaling approaches the Feigenbaum constant. In Lorenz and generalized Hopf systems, the same formulation gives rise to unseen chaotic and multistable long-time behavior.
The verification recipe is system-agnostic, with a blind coarse scan proposes candidates, paired burn-in budgets with phase agreement confirm them, and a closure-plus-multiplier solve certifies each flip, yielding flip parameters and ratios without reference to known bifurcation values (Appendices~\ref{app:protocol} and~\ref{app:evaluation}). The same three ingredients, a parameter token, short local observations and closed-loop evaluation, carry over from scalar maps to continuous flows (Appendices~\ref{app:lorenz}, \ref{app:joint-lorenz} and~\ref{app:generalized-hopf}).

One of the most striking observations is that structural organization appears early in training and follows a trajectory distinct from one-step prediction accuracy.
In the logistic system, a recognizable cascade emerges early in training, while subsequent improvements in one-step OOD prediction do not translate monotonically into improved closed-loop fidelity.
This suggests that recursive application exposes an organizing structure of the learned transition that is not captured by local prediction error alone.
The same principle appears beyond one-dimensional period doubling. In the joint Lorenz system, a model trained on fixed-destination trajectories recovers a structured transition to sustained double-wing motion across two jointly varying parameters and produces positive Lyapunov exponents in chaotic regimes.
In the generalized Hopf system, training trajectories only contract radially, yet the learned dynamics later generate stable cycles and reproduce coexistence between point and cycle attractors from different initial conditions.
The common thread is the emergence of global regime organization from restricted local predictive experience.

Transformer represents parameters and states separately and combines them through attention.
Our experiments address two linked questions: whether a parameter-conditioned transition can emerge through generic token-based autoregressive computation, and how control-parameter information is routed through that computation. The verified cascade and flow results establish the dynamical structure produced under recursive execution, while counterfactual patches and closed-loop blocking identify learned routes through which parameter information shapes these dynamics (Section~\ref{sec:parameter-route}).
Comparisons with direct transition models in Appendix~\ref{app:results} distinguish the structural phenomena shared across representations from the parameter-routing mechanisms examined here.
Together, these analyses connect the realization of a transition through token interactions to its autonomous global dynamical consequences produced under recursive feedback.

%\paragraph{Structure, universality class, and physical alignment.}
%Finite-order scaling describes relations among transitions and need not determine the parameter values at which they occur~\citep{feigenbaum1978}, nor identify the family: the learned map is quadratic-critical even when the training family has a quartic maximum (Appendix~\ref{app:universality}). The closed loop therefore generates a genuine cascade \emph{of the quadratic class}, and where it places that cascade follows the training data, as the logistic displacement and the training-stage measurements both show (Section~\ref{sec:learning}).

These findings suggest a complementary perspective on data-driven scientific modeling.
A learned predictor can encode scientifically meaningful organization not only in its pointwise predictions or internal representations, but also in the dynamical system obtained by recursively applying it.
Bifurcation structure, universality scaling, chaotic instability, and attractor selection therefore become objects for probing what a model has learned about an underlying physical family.
This points toward learned autonomous dynamics as a potentially useful substrate for scientific discovery, especially when the goal is to facilitate understanding global consequences of local observations.

\section{Conclusion}
Autoregressive Transformers trained only on restricted local observations can generate unseen global dynamical organization. In scalar maps, the learned closed-loop systems develop numerically verified high-order period-doubling cascades with Feigenbaum scaling; in Lorenz and generalized Hopf systems, the same parameter-conditioned formulation produces unseen chaotic and multistable long-time behavior. Causal interventions further connect these global dynamics to learned pathways through which control parameters influence state computation.
%These results suggest that a surprisingly narrow window into a system's local behavior may suffice for autoregressive transformers to generalize to its unseen global dynamics.
These results suggest that a surprisingly narrow window into a system's local behavior can give rise to parameter-conditioned dynamics through continuous-token autoregressive computation, producing global structure beyond the observed regime and exposing how control information shapes its recursive evolution.

\section*{Acknowledgments}

The authors gratefully acknowledge the support from the Munich Center for Machine Learning (MCML) and the Program of China Scholarship Council (Grant No.202508080292).
The funding bodies had no role in the design of the study, the development of methodology, the conduct of experiments, the analysis or interpretation of results, or the preparation of the manuscript.

\section*{AI Use Statement}
Generative AI tools were used to assist with language editing and polishing of the manuscript, and to support retrieval and discovery of potentially relevant literature. All retrieved references were independently checked by the authors for relevance and correctness. Generative AI was not involved in the design of methodology and experiments, the analysis and interpretation of results, generating synthetic datasets, or proving mathematical claims. All scientific content, analyses, results, and final wording were reviewed and approved by the authors.

\section*{Reproducibility Statement}
The appendices have documented all detailed descriptions of the training procedures, evaluation protocols, intervention settings and all corresponding implementation details  and experimental results necessary to reproduce all our claims made in this paper.

\label{sec:main-end}
\bibliographystyle{iclr2027_conference}
\bibliography{structural_references}

\vspace{5ex}
\appendix
% Appendix organized by protocol, scalar evidence, representations, mechanisms,
% continuous systems, and reproducibility. Existing labels remain stable.
\startcontents[appendices]
\begingroup
\small
\setlength{\parskip}{3pt}
\printcontents[appendices]{}{1}{\section*{Contents of the Appendices}\setcounter{tocdepth}{2}}
\endgroup
\medskip

\section{Experimental Setup and Training Protocols}
\label{app:protocol}
The scalar comparison uses a fixed 12-run neural experiment set and six polynomial model fits. High-order local verification and intervention experiments reuse the final trained checkpoints. The flow experiments have separate training supports and budgets, recorded in Appendices~\ref{app:lorenz}, \ref{app:generalized-hopf}, and \ref{app:joint-lorenz}.

\subsection{Data Generation and Sampling}

For each batch, we draw a shared sequence length $T$, then 512 independent control parameters uniformly in the training interval and 512 initial states uniformly in $[0.01,0.99]$. The targets are generated by iterating the physical map in float32. No burn-in is discarded from training sequences. Let $G$ have a geometric distribution with success probability $1-\exp(-0.5)$ on $1,2,\ldots$. We set $T=4+G-1$; if this exceeds 64, we redraw $T$ uniformly from the integers 4 through 64. Realized means for seeds 42--44 are 5.5405, 5.52695, and 5.54985, with maxima 22, 22, and 28. Both neural architectures use the same stream seed for each paired comparison.

\subsection{Transformer Architecture}
The Transformer has width 16, two untied residual layers, two attention heads, feed-forward width 64, and maximum total context 256. Input projections have bias; attention projections, feed-forward projections, and the output head have no bias. Learned parameter/state type embeddings distinguish token roles.
\subsection{Optimization and Checkpoint Selection}
Both neural architectures use teacher forcing at every state position. Training uses AdamW with its PyTorch default momentum coefficients $(0.9,0.999)$ and $\epsilon=10^{-8}$, 20,000 updates, learning rate $5\times10^{-4}$, weight decay $10^{-5}$, 500 linear warm-up steps, and cosine decay to zero. The saved final checkpoint supplies the fixed-grid comparisons, high-order verification, and interventions. Training-stage figures additionally show intermediate checkpoints; the main overview and matched trajectory/cobweb examples use the explicitly labeled 10,000-update stage.

The direct neural and polynomial transition models are specified in
Appendix~\ref{app:results}; software and execution devices are recorded in
Appendix~\ref{app:artifacts}.

\section{Evaluation and Numerical Verification}
\label{app:measure}
\label{app:evaluation}
This section defines the measurements and verification procedures used throughout
the scalar experiments. Detailed numerical results appear in
Appendix~\ref{app:scalar}; system-specific flow protocols appear in
Appendices~\ref{app:lorenz-results} and~\ref{app:generalized-hopf}.

\subsection{Distributional and Local Prediction Metrics}
\label{sec:distributional}
Uniform-grid evaluations use 513 parameter values over $[2,4]$ or $[0.1,1]$. Three initial states are sampled once from $[0.05,0.95]$ using NumPy generator seed 20260912 and shared by every model and reference. We discard 1,024 iterations and retain the next 512. Neural inference uses float32 and reference inference uses float64. Predicted states are clipped to $[0,1]$. At each parameter, $W_1$ is computed over all $3\times512$ retained observations. OOD averages include parameters strictly above the training upper endpoint.

For parameter $r$ and retained tail length $N=512$, we define the empirical distributional measures
\begin{equation}
 \widehat\mu_r=\frac{1}{3N}\sum_{i=1}^{3}\sum_{t=1}^{N}\delta_{\hat x_t^{(i)}},
 \qquad
 \mu_r=\frac{1}{3N}\sum_{i=1}^{3}\sum_{t=1}^{N}\delta_{x_t^{(i)}}
 \label{eq:measures}
\end{equation}
for the learned and reference systems, respectively. We compare them with one-dimensional Wasserstein-1 distance $W_1(\widehat\mu_r,\mu_r)$. %The reported OOD distributional score averages this quantity over evaluated parameter values strictly outside the boundary of the training interval.

Distributional agreement and periodic organization measure complementary properties
of the autonomous dynamics. Fixed one-step probes and longitudinal measurements
are specified in Appendix~\ref{app:training}.

\subsection{Period Detection}
\label{app:period-detection}
For a retained scalar sequence $z_t$, the base tolerance is $\epsilon=10^{-5}+10^{-4}(\max z-\min z)$. Candidate periods are tested in increasing order, up to $\min(64,\lfloor N/6\rfloor)$, with at least 64 samples. The 0.99 quantile of $|z_{t+p}-z_t|/\epsilon$ must be at most one. Values are then arranged by cycle phase. Within-phase deviations from phase medians and the change between early and late phase medians must also satisfy the tolerance. For $p>1$, the phase tolerance is capped at one percent of the minimum separation between distinct phase medians. This additional relative check resolves small-amplitude oscillations close to a flip. An unaccepted sequence is unresolved, which includes aperiodicity, long transients, and insufficient resolution; it is not classified as chaos.

Parameter-level cascade detection requires at least 80\% initial-condition agreement, which requires all three initial states here. Both sides of a transition need three consecutive grid points. The maximum bracket width is 3\% of the scanned parameter span. A primary cascade must proceed consecutively from period one without jumping a contradictory periodic region. Period agreement in the main comparisons is evaluated at individual parameter--initial-state pairs, over all evaluated parameters for which the reference period is accepted. It is distinct from the OOD-only $W_1$ average.
\subsection{Adaptive Refinement and Flip Verification}
\label{sec:verification}

A finite-resolution bifurcation diagram might be visually misleading near high-order transitions because of sparse parameter sampling, long transients, and unrelated periodic windows.
We therefore separate candidate detection from numerical verification.
A blind trajectory scan searches for consecutive $1\to2\to4\to\cdots$ transitions without initializing the search at known reference bifurcation locations.
Candidate transitions must be supported on both sides of the boundary and satisfy the prescribed consensus across initial conditions.
Each accepted bracket, together with the next unresolved boundary, is then adaptively refined on a local parameter grid.
At every refinement stage, trajectories are evaluated with two burn-in budgets $B$ and $2B$; a candidate period must persist under the longer budget, and corresponding cycle phases from the two evaluations must agree up to cyclic alignment.
%This guards against slowly decaying transients being mistaken for stable high-order cycles.
For a candidate flip whose parent orbit has period $p$, we then independently solve
\begin{equation}
 F_r^p(x)-x=0,\qquad \Lambda_p(r,x)=-1,
 \label{eq:flip}
\end{equation}
with the cycle multiplier $\Lambda_p$ of Equation~\ref{eq:multiplier}. For the learned system this calculation uses the one-state map $F_{\theta,r}$ of Equation~\ref{eq:induced-map}, with derivatives obtained by automatic differentiation, whereas analytic derivatives are used for the reference scalar maps. %Local root calculations are performed in float64 without modifying the learned weights.
A transition is reported as numerically verified only if the joint solution lies inside the independently trajectory-derived bracket and continuation of the parent orbit to parameters on both sides confirms the corresponding stable-to-unstable crossing.
%Thus neither the known reference bifurcation locations nor the Feigenbaum constant are used to locate learned transitions.
For the verified flip parameters $r_1,\ldots,r_K$ we compute the finite-order interval ratios $\delta_n$ of Equation~\ref{eq:ratio}, and we report the physical locations $r_n$ and the ratios $\delta_n$ separately. The former measure alignment of the learned and reference parameterizations, whereas the latter measure the relative geometry of the cascade.

The high-order analysis uses up to period 128, three fixed initial states $(0.123456789,0.371239,0.817321)$, and 33-point local grids. The initial uniform scans locate candidates without using theoretical bifurcation values. The reference scans use 2,049 parameters; neural scans reuse their main 513-point arrays. Each local stage tests burn-in budgets $B$ and $2B$ and checks phase-center agreement up to a cyclic shift, in addition to period labels. Previously retained tails allow reanalysis of convergence criteria without changing learned weights.

The root calculations for parent periods $p=1,2,4,\ldots,64$ use float64 arithmetic without modifying the learned weights.

\subsection{Evaluation of Continuous Flows}

\label{sec:flow-evaluation}
For the continuous-time systems, the relevant long-run structures are not adequately described by scalar period labels, so we use system-specific diagnostics on the frozen discrete flow predictor of Equation~\ref{eq:flow-predictor}.

For the joint Lorenz experiment, the primary coarse comparison classifies each parameter pair as fixed or nonfixed from the long-time trajectory, using the same numerical criterion that defines the fixed-destination training parameters.
On nonfixed trajectories we additionally measure marginal state distributions and repeated motion between the two wings. Because repeated wing switching alone does not establish chaos, to test chaotic instability directly, we estimate the largest Lyapunov exponent of the frozen learned map by propagating a tangent vector through successive learned updates and repeatedly renormalizing it. A positive estimated largest exponent provides evidence of exponential separation along the generated attractor under the specified finite-time estimator.
The wider two-parameter Lorenz analysis additionally identifies resolved periodic behavior through recurrence of crossings of a fixed Poincar\'e-section. Minimal recurrence lags are tested over a prescribed range with fixed crossing-count and recurrence-distance criteria.

For the generalized Hopf system, the principal asymptotic distinction is between attraction to the origin and attraction to a periodic orbit. We therefore evaluate long-time radius and angular motion from multiple initial radii and phases. Point outcomes require sufficiently small mean radius, while cycle outcomes require sustained nonzero radius together with low radial variation and angular speed close to the reference rotation rate. Multiple initial conditions at the same parameter value allow coexistence of attracting states to be tested directly rather than inferred from a single trajectory. The operational thresholds and evaluation windows are fixed in advance and specified in Appendix~\ref{app:generalized-hopf}.

The reference and learned systems are evaluated using the same estimator. For more details, see Appendix~\ref{app:flow-lyapunov}.

\section{Supplementary Results and Diagnostics for Scalar Maps}
\label{app:scalar}
\subsection{Verified Cascades and Scaling}
\label{app:cascade-results}
The following tables collect the adaptive evaluation budgets and numerically
verified roots obtained with the procedure in Appendix~\ref{sec:verification}.
The shared root table also supplies the structural comparison in
Appendix~\ref{app:structural-comparisons}.
\begin{table}[!htbp]
\centering
\small
\caption{Final adaptive budgets. Each level compares two burn-in budgets; ``maximum'' is the longer final budget. Neural maps use logistic seed 42 and one current state.}
\vspace{-1ex}
\begin{tabular}{lrrr}
\toprule
Map & Levels & Max. burn & Tail\\
\midrule
Logistic reference & 4 & 65,536 & 2,048\\
Sine reference & 4 & 65,536 & 2,048\\
MLP & 5 & 131,072 & 2,048\\
Transformer & 4 & 32,768 & 1,024\\
\bottomrule
\end{tabular}
\end{table}

Continuation to either side checks the parent cycle's loss of stability. All reported solutions lie inside the independently detected trajectory brackets. Table~\ref{tab:allroots} lists the resulting seven flip parameters, and under the same definition the sine reference ladder reaches a final finite-order ratio of $4.668805$. The bracket widths represent spatial/recognition resolution, not statistical confidence intervals. Tests of the detector cover analytic first flips, known periods, unresolved tails, missing boundaries, later periodic islands, and slowly decaying oscillations. The reported sequences require cross-budget phase agreement, not only matching period labels.

\begin{table*}[t]
\centering
\small
\caption{All numerically verified flip parameters. Learned maps use logistic seed 42. The Transformer receives one state token and a parameter token.}\label{tab:allroots}
\vspace{-1ex}
\begin{tabular}{rcccc}\toprule
Index & Logistic reference & Sine reference & Logistic MLP & Logistic Transformer\\\midrule
1 & 3.0000000000 & 0.7199616830 & 2.9879977074 & 2.9692234640\\
2 & 3.4494897428 & 0.8332663537 & 3.4523234903 & 3.3927447143\\
3 & 3.5440903596 & 0.8586090599 & 3.5573408693 & 3.4848703460\\
4 & 3.5644072661 & 0.8640841737 & 3.5802157502 & 3.5048663661\\
5 & 3.5687594195 & 0.8652589607 & 3.5851355903 & 3.5091579915\\
6 & 3.5696916098 & 0.8655106638 & 3.5861901963 & 3.5100776679\\
7 & 3.5698912594 & 0.8655645755 & 3.5864161045 & 3.5102746561\\
\bottomrule\end{tabular}\end{table*}

\subsubsection{Precision of the Verified Cascade}

\label{app:float32}
The roots above are solved on float64 maps while the headline rollouts and comparisons use float32
inference. Re-evaluating the seven verified logistic flips of the seed-42 Transformer inside the
float32 model gives closure residuals between $1.2\times10^{-7}$ and $2.1\times10^{-5}$ and
multipliers within $2\%$ of $-1$ for all seven flips. One-sided period probes placed at the midpoint
of each inter-flip window reproduce the expected parent or daughter period in six of seven cases in
float32 and in six of seven cases in float64; the unmatched cases are the period-64 and period-128
windows, whose width is about $2\times10^{-4}$. The verified structure is therefore not an artifact
of float64 evaluation, but the two highest-order transitions are not resolvable at deployment
precision.

\subsubsection{Detailed Cascade Geometry}
\label{app:visual-atlas}
The local geometry and progressive magnifications below complement the numerical
verification. Full seed and context comparisons appear in
Appendix~\ref{app:seed-context}, and alternative-model comparisons in
Appendix~\ref{app:results}.

The local cascade views use a separate 1,025-point uniform parameter grid, three initial states $(0.123456789,0.371239,0.817321)$, 8,192 burn-in steps, and 512 retained steps. Reference and seed-42 one-state Transformer rollouts use float64; trained weights are unchanged. The logistic window is $[3.35,3.62]$ and the sine window is $[0.80,0.92]$. These magnifications reveal local shape and displacement and do not replace the fixed-grid accuracy measurements or the independent flip-verification procedure.

\begin{figure}[!htbp]
\centering
\begin{subfigure}[t]{.4\textwidth}
\centering
\includegraphics[width=\linewidth]{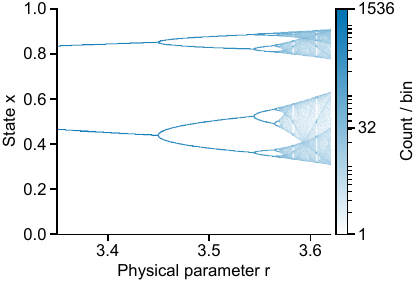}
\caption{Logistic reference}
\end{subfigure}
\quad
\begin{subfigure}[t]{.4\textwidth}
\centering
\includegraphics[width=\linewidth]{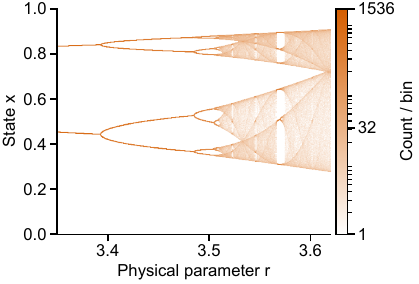}
\caption{Logistic Transformer, seed 42}
\end{subfigure}
\par\medskip
\begin{subfigure}[t]{.4\textwidth}
\centering
\includegraphics[width=\linewidth]{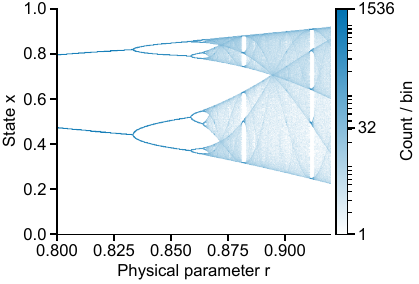}
\caption{Sine reference}
\end{subfigure}
\quad
\begin{subfigure}[t]{.4\textwidth}
\centering
\includegraphics[width=\linewidth]{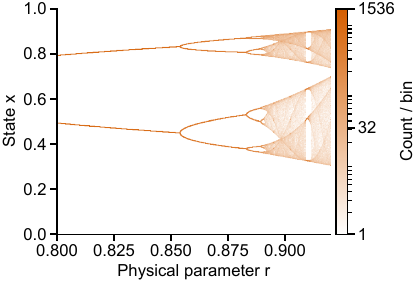}
\caption{Sine Transformer, seed 42}
\end{subfigure}
\caption{\textbf{Dense local views resolve the cascade shape.} Reference and Transformer panels share the same physical parameter window within each system. All $1{,}025\times3\times512$ retained samples enter the histograms. The parameter positions are not aligned or rescaled between reference and learned maps.}
\label{fig:local-cascades}
\end{figure}

\subsubsection{Progressive Magnification of Verified Flips}

The following views use the saved blind adaptive trajectories and numerically verified roots. At duplicate parameter values, the longest available burn-in is retained. Each view spans $r_n\pm0.16(r_n-r_{n-1})$ and magnifies the orbit branch nearest $x=1/2$. The parent phase nearest $x=1/2$ is isolated using 45\% of the distance to its nearest neighboring phase just below the root. The vertical crop includes the observed branch and its daughters across sampled parameters, with a 10\% margin on each side; all observations within the crop are plotted. Dots occupy only evaluated parameter values: gaps are not interpolated. Dashed lines mark verified roots. Physical coordinates are retained, so these panels expose both repeated local splitting and displacement. The source manifest records exact crops, sample counts, and burn-in budgets. Sine panels here calibrate the reference map; the verified neural high-order panels concern logistic seed 42.

\begin{figure}[!htbp]
\centering
\begin{subfigure}[t]{.4\textwidth}
\centering
\includegraphics[width=\linewidth]{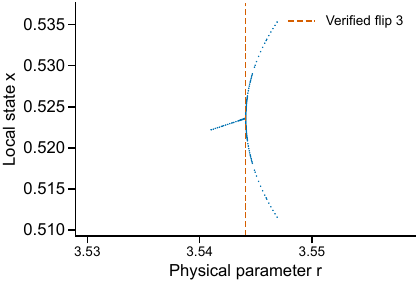}
\caption{Logistic reference}
\end{subfigure}
\quad
\begin{subfigure}[t]{.4\textwidth}
\centering
\includegraphics[width=\linewidth]{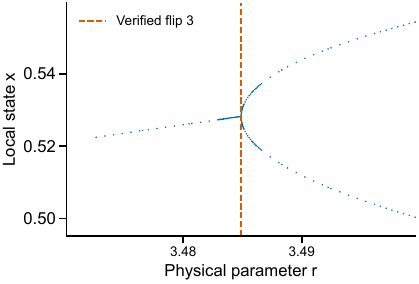}
\caption{Logistic Transformer, seed 42}
\end{subfigure}
\par\medskip
\begin{subfigure}[t]{.4\textwidth}
\centering
\includegraphics[width=\linewidth]{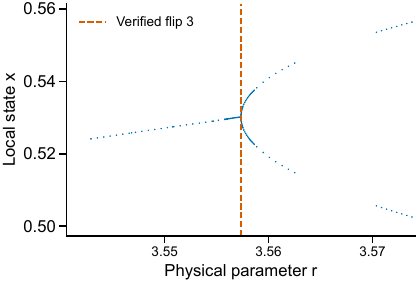}
\caption{Logistic MLP, seed 42}
\end{subfigure}
\quad
\begin{subfigure}[t]{.4\textwidth}
\centering
\includegraphics[width=\linewidth]{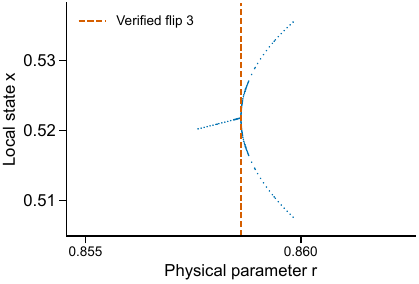}
\caption{Sine reference}
\end{subfigure}
\caption{\textbf{Local geometry at doubling 3.} The verified parent period is $4$ and the emerging doubled period is $8$. Each panel uses its own physical root and explicitly labeled local state range. Dense clouds near a root can include slow transients; root verification additionally uses cycle and multiplier checks.}
\label{fig:local-flip3}
\end{figure}

\begin{figure}[!htbp]
\centering
\begin{subfigure}[t]{.4\textwidth}
\centering
\includegraphics[width=\linewidth]{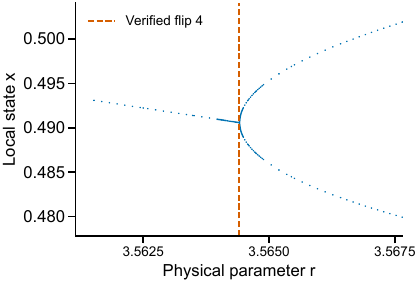}
\caption{Logistic reference}
\end{subfigure}
\quad
\begin{subfigure}[t]{.4\textwidth}
\centering
\includegraphics[width=\linewidth]{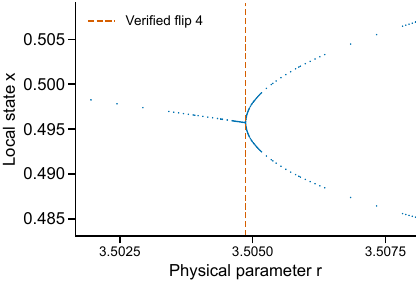}
\caption{Logistic Transformer, seed 42}
\end{subfigure}
\par\medskip
\begin{subfigure}[t]{.4\textwidth}
\centering
\includegraphics[width=\linewidth]{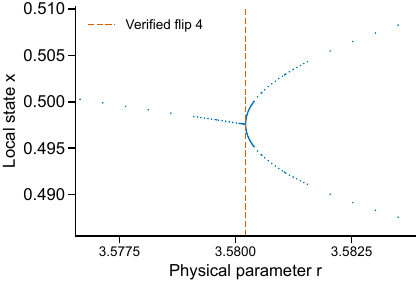}
\caption{Logistic MLP, seed 42}
\end{subfigure}
\quad
\begin{subfigure}[t]{.4\textwidth}
\centering
\includegraphics[width=\linewidth]{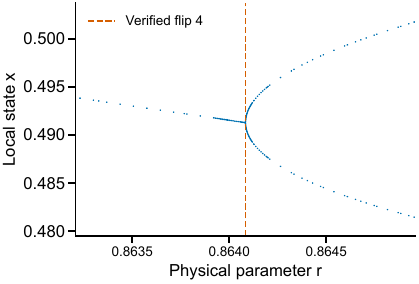}
\caption{Sine reference}
\end{subfigure}
\caption{\textbf{Local geometry at doubling 4.} The verified parent period is $8$ and the emerging doubled period is $16$. Each panel uses its own physical root and explicitly labeled local state range. Dense clouds near a root can include slow transients; root verification additionally uses cycle and multiplier checks.}
\label{fig:local-flip4}
\end{figure}

\begin{figure}[!htbp]
\centering
\begin{subfigure}[t]{.4\textwidth}
\centering
\includegraphics[width=\linewidth]{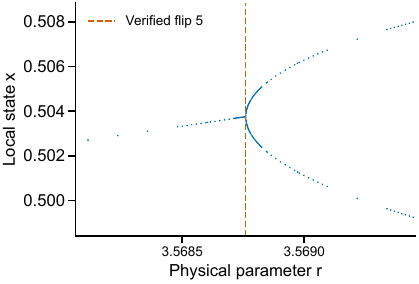}
\caption{Logistic reference}
\end{subfigure}
\quad
\begin{subfigure}[t]{.4\textwidth}
\centering
\includegraphics[width=\linewidth]{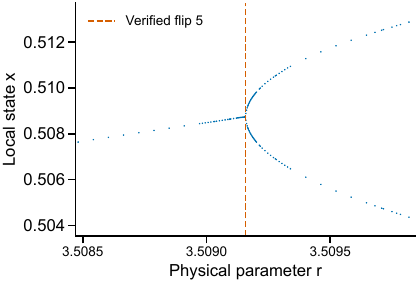}
\caption{Logistic Transformer, seed 42}
\end{subfigure}
\par\medskip
\begin{subfigure}[t]{.4\textwidth}
\centering
\includegraphics[width=\linewidth]{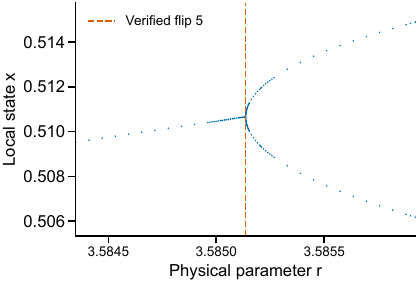}
\caption{Logistic MLP, seed 42}
\end{subfigure}
\quad
\begin{subfigure}[t]{.4\textwidth}
\centering
\includegraphics[width=\linewidth]{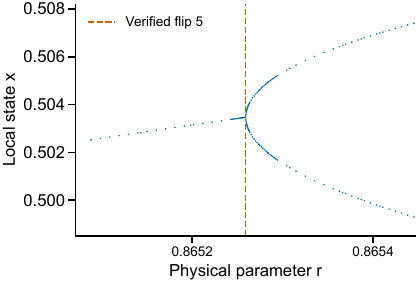}
\caption{Sine reference}
\end{subfigure}
\caption{\textbf{Local geometry at doubling 5.} The verified parent period is $16$ and the emerging doubled period is $32$. Each panel uses its own physical root and explicitly labeled local state range. Dense clouds near a root can include slow transients; root verification additionally uses cycle and multiplier checks.}
\label{fig:local-flip5}
\end{figure}

\begin{figure}[!htbp]
\centering
\begin{subfigure}[t]{.4\textwidth}
\centering
\includegraphics[width=\linewidth]{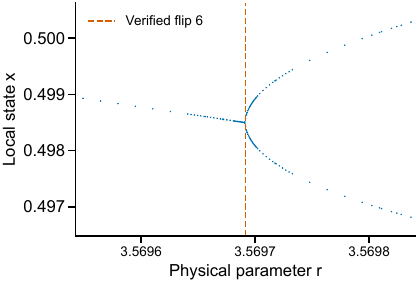}
\caption{Logistic reference}
\end{subfigure}
\quad
\begin{subfigure}[t]{.4\textwidth}
\centering
\includegraphics[width=\linewidth]{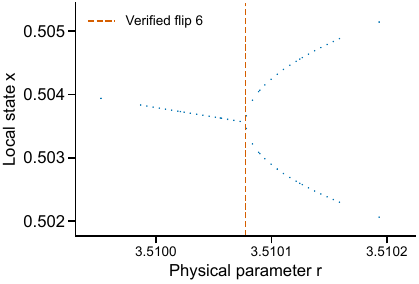}
\caption{Logistic Transformer, seed 42}
\end{subfigure}
\par\medskip
\begin{subfigure}[t]{.4\textwidth}
\centering
\includegraphics[width=\linewidth]{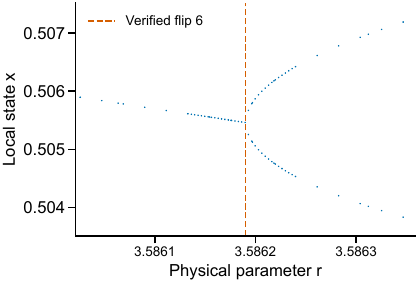}
\caption{Logistic MLP, seed 42}
\end{subfigure}
\quad
\begin{subfigure}[t]{.4\textwidth}
\centering
\includegraphics[width=\linewidth]{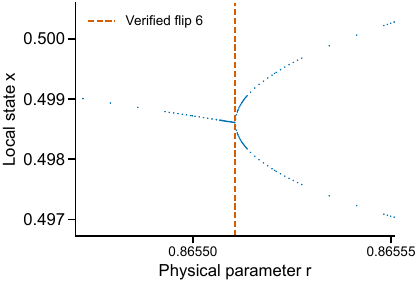}
\caption{Sine reference}
\end{subfigure}
\caption{\textbf{Local geometry at doubling 6.} The verified parent period is $32$ and the emerging doubled period is $64$. Each panel uses its own physical root and explicitly labeled local state range. Dense clouds near a root can include slow transients; root verification additionally uses cycle and multiplier checks.}
\label{fig:local-flip6}
\end{figure}

\begin{figure}[!htbp]
\centering
\begin{subfigure}[t]{.4\textwidth}
\centering
\includegraphics[width=\linewidth]{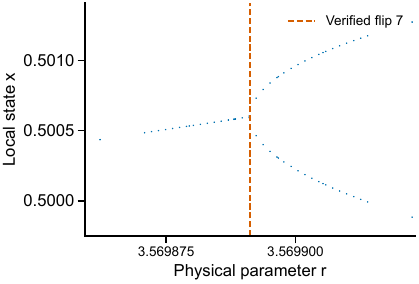}
\caption{Logistic reference}
\end{subfigure}
\quad
\begin{subfigure}[t]{.4\textwidth}
\centering
\includegraphics[width=\linewidth]{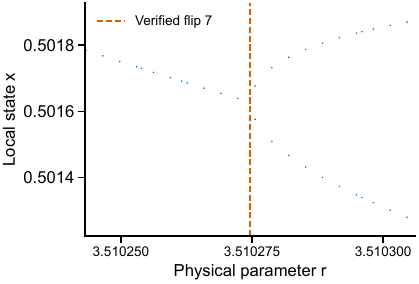}
\caption{Logistic Transformer, seed 42}
\end{subfigure}
\par\medskip
\begin{subfigure}[t]{.4\textwidth}
\centering
\includegraphics[width=\linewidth]{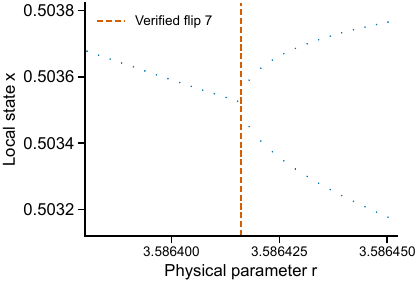}
\caption{Logistic MLP, seed 42}
\end{subfigure}
\quad
\begin{subfigure}[t]{.4\textwidth}
\centering
\includegraphics[width=\linewidth]{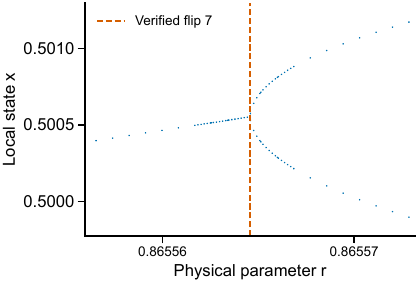}
\caption{Sine reference}
\end{subfigure}
\caption{\textbf{Local geometry at doubling 7.} The verified parent period is $64$ and the emerging doubled period is $128$. Each panel uses its own physical root and explicitly labeled local state range. Dense clouds near a root can include slow transients; root verification additionally uses cycle and multiplier checks.}
\label{fig:local-flip7}
\end{figure}

\subsection{Emergence during Training}
\label{app:training}
\label{sec:local-vs-autonomous}

We record training dynamics for seed 42 on each system using the original 20,000-update optimization protocol. Weight snapshots are saved every 250 updates, including initialization, giving 81 stages per system. Training minibatch loss is recorded at each saved update. After training, every snapshot is evaluated on the same fixed 2,048 ID and 2,048 OOD one-state examples: parameters are uniform within each region and states are uniform on $[0.01,0.99]$. These pointwise diagnostics are distinct from autonomous trajectory fidelity.

Closed-loop evaluations use all 23 predeclared stages: initialization, 250 and 500 updates, and every 1,000 updates through 20,000. Each uses the original 513-point physical grid, the same three initial states, 1,024 burn-in steps, 512 retained states, and $C=1$. One-step predictions are unclipped; autonomous rollouts use the $[0,1]$ clamp. The figures below display every evaluated stage. Error and structure curves connect these fixed stages; connecting lines do not resolve intermediate events. Period matching is evaluated only where reference periodicity is confirmed. The coarse flip count uses the trajectory detector on this uniform grid and is separate from adaptive high-order verification.

The final replayed weights match the original reported checkpoint exactly, tensor by tensor, on both systems. These longitudinal views therefore end at the same learned maps used in the main comparisons.

\subsubsection{Structural Emergence and Physical Alignment}
\begin{figure}[!htbp]
\centering
\begin{subfigure}[t]{.35\textwidth}
\centering
\includegraphics[width=\linewidth]{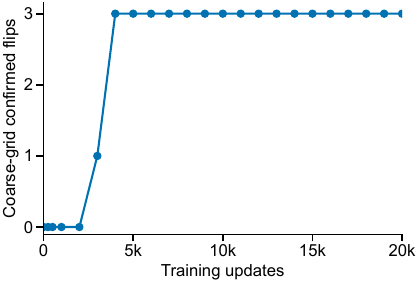}
\caption{Logistic}
\end{subfigure}
\quad
\begin{subfigure}[t]{.35\textwidth}
\centering
\includegraphics[width=\linewidth]{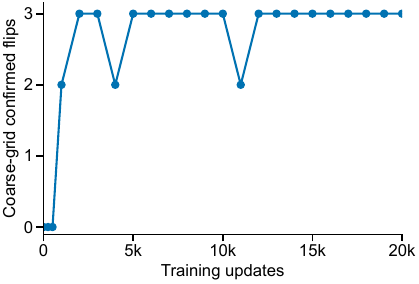}
\caption{Sine}
\end{subfigure}
\caption{\textbf{Coarse structural detection across training.} Each point counts consecutive confirmed period doublings on the fixed 513-point grid. Zero denotes no cascade confirmed at this budget. Adaptive high-order verification uses the separate local protocol.}
\label{fig:training-flips}
\end{figure}

\begin{figure}[!htbp]
\centering
\begin{subfigure}[t]{.35\textwidth}
\centering
\includegraphics[width=\linewidth]{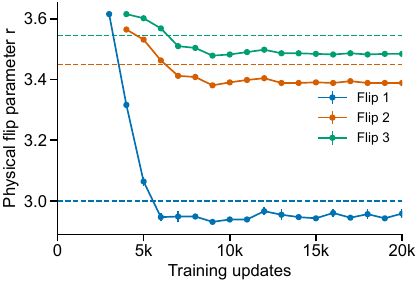}
\caption{Logistic}
\end{subfigure}
\quad
\begin{subfigure}[t]{.35\textwidth}
\centering
\includegraphics[width=\linewidth]{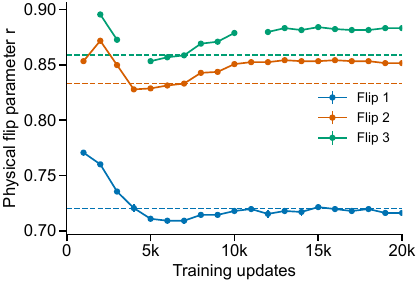}
\caption{Sine}
\end{subfigure}
\caption{\textbf{Flip parameters move during training.} Points show the first three detected flip parameters and vertical bars show their trajectory brackets. Dashed horizontal lines mark the reference flip parameters, solved independently of the trajectory detector. Unresolved detections appear as gaps. These curves track geometric displacement separately from the number of detected doublings.}
\label{fig:training-locations}
\end{figure}

The training records separate the appearance of a cascade from its quantitative alignment with the physical system (Figures~\ref{fig:training-flips} and~\ref{fig:training-locations}). In the logistic replay, the fixed-grid detector confirms no doubling through the 2,000-update recording, one at 3,000, and three at 4,000. At 4,000 updates, however, the first detected transition is near $r=3.32$, compared with the physical value $r=3$. A recognizable cascade therefore appears while its location is still moving. The sine replay already has two detected doublings at the 1,000-update recording and three at 2,000; subsequent coarse counts and locations continue to vary.

\begin{figure}[!htbp]
\centering
\begin{subfigure}[t]{.35\textwidth}
\centering
\includegraphics[width=\linewidth]{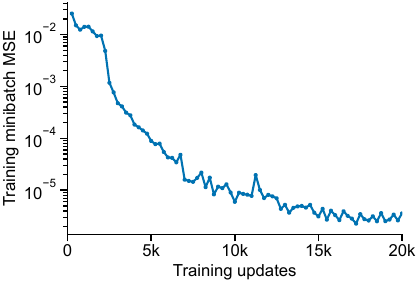}
\caption{Training minibatch loss}
\end{subfigure}
\quad
\begin{subfigure}[t]{.35\textwidth}
\centering
\includegraphics[width=\linewidth]{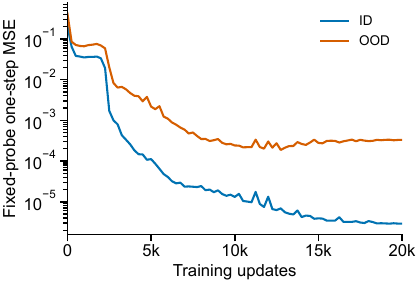}
\caption{Fixed ID and OOD one-step probes}
\end{subfigure}
\par\medskip
\begin{subfigure}[t]{.35\textwidth}
\centering
\includegraphics[width=\linewidth]{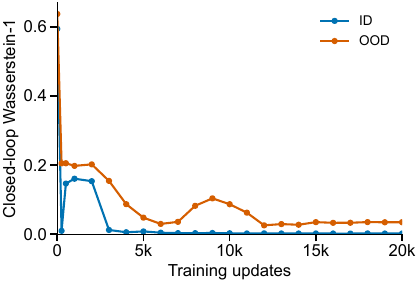}
\caption{Autonomous distributional fidelity}
\end{subfigure}
\quad
\begin{subfigure}[t]{.35\textwidth}
\centering
\includegraphics[width=\linewidth]{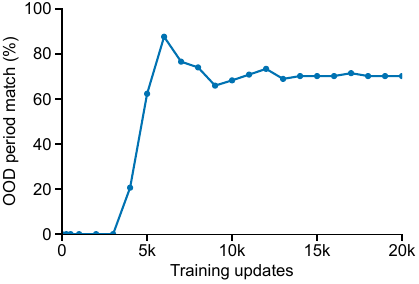}
\caption{OOD periodic-structure agreement}
\end{subfigure}
\caption{\textbf{Logistic learning curves.} Training and probe measurements are recorded every 250 updates; autonomous diagnostics use the 23 fixed stages. One-step probes use 2,048 samples per region. Period matching uses confirmed periodic reference trajectories, with unresolved model outputs counted as mismatches. All curves concern seed 42 and one state token.}
\label{fig:training-curves-logistic}
\end{figure}

One-step fitting and autonomous fidelity follow different trajectories (Figure~\ref{fig:training-curves-logistic}). For logistic, the fixed-probe OOD one-step MSE decreases from $1.10\times10^{-3}$ at 6,000 updates to $2.41\times10^{-4}$ at 10,000, while closed-loop OOD $W_1$ rises from $0.0296$ to $0.0867$. Across these stages, three doublings remain confirmed on the coarse grid. The learned update becomes more accurate on the sampled one-step task while its generated long-run distribution moves non-monotonically. Together, the stage diagrams, location curves, and fidelity curves distinguish learning a period-doubling structure from reproducing its precise physical realization.

\begin{figure}[!htbp]
\centering
\begin{subfigure}[t]{.35\textwidth}
\centering
\includegraphics[width=\linewidth]{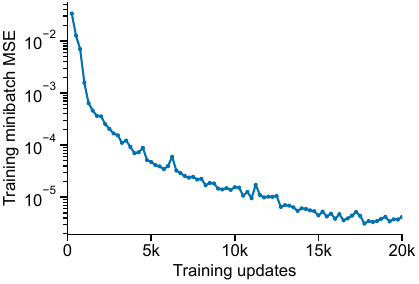}
\caption{Training minibatch loss}
\end{subfigure}
\quad
\begin{subfigure}[t]{.35\textwidth}
\centering
\includegraphics[width=\linewidth]{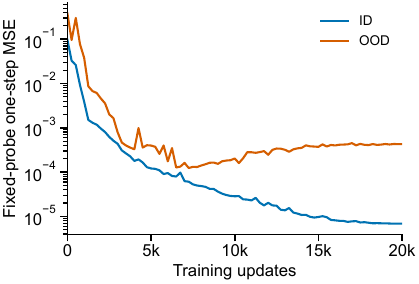}
\caption{Fixed ID and OOD one-step probes}
\end{subfigure}
\par\medskip
\begin{subfigure}[t]{.35\textwidth}
\centering
\includegraphics[width=\linewidth]{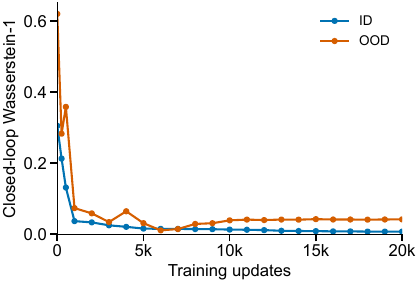}
\caption{Autonomous distributional fidelity}
\end{subfigure}
\quad
\begin{subfigure}[t]{.35\textwidth}
\centering
\includegraphics[width=\linewidth]{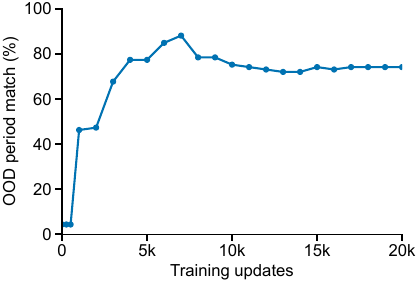}
\caption{OOD periodic-structure agreement}
\end{subfigure}
\caption{\textbf{Sine learning curves.} Training and probe measurements are recorded every 250 updates; autonomous diagnostics use the 23 fixed stages. One-step probes use 2,048 samples per region. Period matching uses confirmed periodic reference trajectories, with unresolved model outputs counted as mismatches. All curves concern seed 42 and one state token.}
\label{fig:training-curves-sine}
\end{figure}

\subsubsection{Logistic Attractor Evolution}

\begin{figure}[H]
\centering
\setlength{\tabcolsep}{0pt}
\renewcommand{\arraystretch}{0.8}
\noindent\begin{tabular*}{\textwidth}{@{\extracolsep{\fill}}*{4}{c}@{}}
\footnotesize Init. & \footnotesize 250 & \footnotesize 500 & \footnotesize 1k \\[-2pt]
\includegraphics[width=0.24\textwidth]{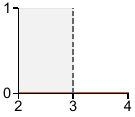} & \includegraphics[width=0.24\textwidth]{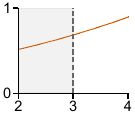} & \includegraphics[width=0.24\textwidth]{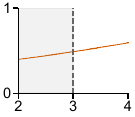} & \includegraphics[width=0.24\textwidth]{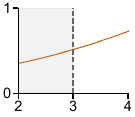} \\[1pt]
\footnotesize 2k & \footnotesize 3k & \footnotesize 4k & \footnotesize 5k \\[-2pt]
\includegraphics[width=0.24\textwidth]{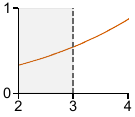} & \includegraphics[width=0.24\textwidth]{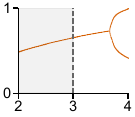} & \includegraphics[width=0.24\textwidth]{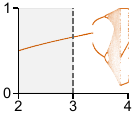} & \includegraphics[width=0.24\textwidth]{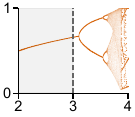} \\[1pt]
\footnotesize 6k & \footnotesize 7k & \footnotesize 8k & \footnotesize 9k \\[-2pt]
\includegraphics[width=0.24\textwidth]{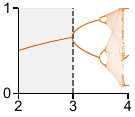} & \includegraphics[width=0.24\textwidth]{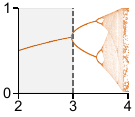} & \includegraphics[width=0.24\textwidth]{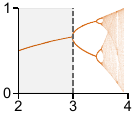} & \includegraphics[width=0.24\textwidth]{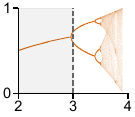} \\[1pt]
\footnotesize 10k & \footnotesize 11k & \footnotesize 12k & \footnotesize 13k \\[-2pt]
\includegraphics[width=0.24\textwidth]{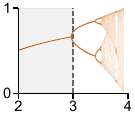} & \includegraphics[width=0.24\textwidth]{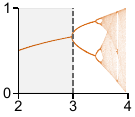} & \includegraphics[width=0.24\textwidth]{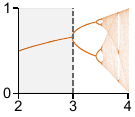} & \includegraphics[width=0.24\textwidth]{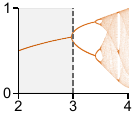} \\[1pt]
\footnotesize 14k & \footnotesize 15k & \footnotesize 16k & \footnotesize 17k \\[-2pt]
\includegraphics[width=0.24\textwidth]{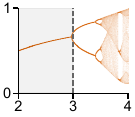} & \includegraphics[width=0.24\textwidth]{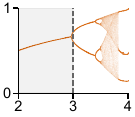} & \includegraphics[width=0.24\textwidth]{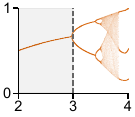} & \includegraphics[width=0.24\textwidth]{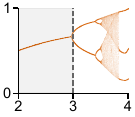} \\[1pt]
\footnotesize 18k & \footnotesize 19k & \footnotesize 20k & \footnotesize Ground truth \\[-2pt]
\includegraphics[width=0.24\textwidth]{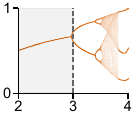} & \includegraphics[width=0.24\textwidth]{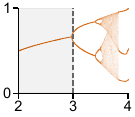} & \includegraphics[width=0.24\textwidth]{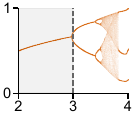} & \includegraphics[width=0.24\textwidth]{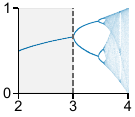} \\[1pt]
\end{tabular*}
\caption{\textbf{Logistic attractor evolution.} All 23 closed-loop stages plus the ground-truth reference, arranged as four columns and six rows. Panels use the same compact style as Figure~\ref{fig:learning-evolution}: no colorbars or axis titles, ticks only at the range endpoints and training boundary, and gray shading for the training interval. Orange denotes the Transformer and blue the reference.}
\label{fig:training-logistic-atlas}
\end{figure}

\subsubsection{Sine Attractor Evolution}

\begin{figure}[H]
\centering
\setlength{\tabcolsep}{0pt}
\renewcommand{\arraystretch}{0.8}
\noindent\begin{tabular*}{\textwidth}{@{\extracolsep{\fill}}*{4}{c}@{}}
\footnotesize Init. & \footnotesize 250 & \footnotesize 500 & \footnotesize 1k \\[-2pt]
\includegraphics[width=0.24\textwidth]{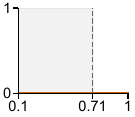} & \includegraphics[width=0.24\textwidth]{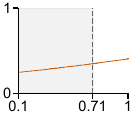} & \includegraphics[width=0.24\textwidth]{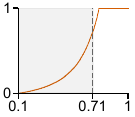} & \includegraphics[width=0.24\textwidth]{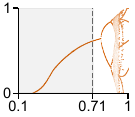} \\[1pt]
\footnotesize 2k & \footnotesize 3k & \footnotesize 4k & \footnotesize 5k \\[-2pt]
\includegraphics[width=0.24\textwidth]{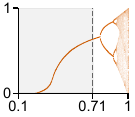} & \includegraphics[width=0.24\textwidth]{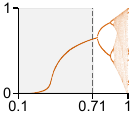} & \includegraphics[width=0.24\textwidth]{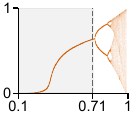} & \includegraphics[width=0.24\textwidth]{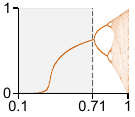} \\[1pt]
\footnotesize 6k & \footnotesize 7k & \footnotesize 8k & \footnotesize 9k \\[-2pt]
\includegraphics[width=0.24\textwidth]{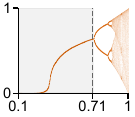} & \includegraphics[width=0.24\textwidth]{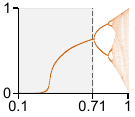} & \includegraphics[width=0.24\textwidth]{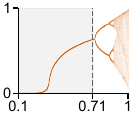} & \includegraphics[width=0.24\textwidth]{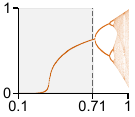} \\[1pt]
\footnotesize 10k & \footnotesize 11k & \footnotesize 12k & \footnotesize 13k \\[-2pt]
\includegraphics[width=0.24\textwidth]{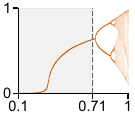} & \includegraphics[width=0.24\textwidth]{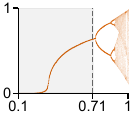} & \includegraphics[width=0.24\textwidth]{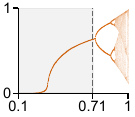} & \includegraphics[width=0.24\textwidth]{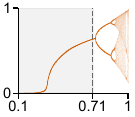} \\[1pt]
\footnotesize 14k & \footnotesize 15k & \footnotesize 16k & \footnotesize 17k \\[-2pt]
\includegraphics[width=0.24\textwidth]{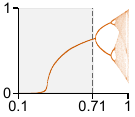} & \includegraphics[width=0.24\textwidth]{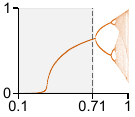} & \includegraphics[width=0.24\textwidth]{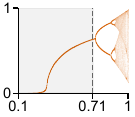} & \includegraphics[width=0.24\textwidth]{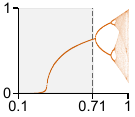} \\[1pt]
\footnotesize 18k & \footnotesize 19k & \footnotesize 20k & \footnotesize Ground truth \\[-2pt]
\includegraphics[width=0.24\textwidth]{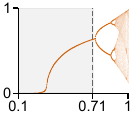} & \includegraphics[width=0.24\textwidth]{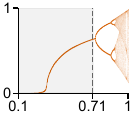} & \includegraphics[width=0.24\textwidth]{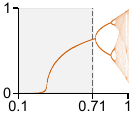} & \includegraphics[width=0.24\textwidth]{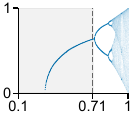} \\[1pt]
\end{tabular*}
\caption{\textbf{Sine attractor evolution.} All 23 closed-loop stages plus the ground-truth reference, arranged as four columns and six rows. Panels use the same compact style as Figure~\ref{fig:learning-evolution}: no colorbars or axis titles, ticks only at the range endpoints and training boundary, and gray shading for the training interval. Orange denotes the Transformer and blue the reference.}
\label{fig:training-sine-atlas}
\end{figure}

\subsubsection{Matched Trajectories and Cobwebs across Training}
\label{app:cobweb}
\begin{figure}[!htbp]
\centering
\begin{subfigure}[t]{.19\textwidth}
\centering
\includegraphics[width=\linewidth]{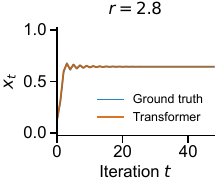}
\caption{Stable}
\end{subfigure}
\hfill
\begin{subfigure}[t]{.19\textwidth}
\centering
\includegraphics[width=\linewidth]{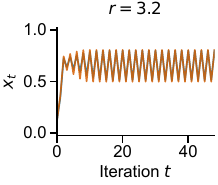}
\caption{Period 2}
\end{subfigure}
\hfill
\begin{subfigure}[t]{.19\textwidth}
\centering
\includegraphics[width=\linewidth]{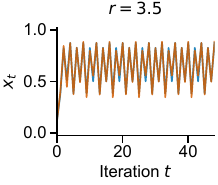}
\caption{Period 4}
\end{subfigure}
\hfill
\begin{subfigure}[t]{.19\textwidth}
\centering
\includegraphics[width=\linewidth]{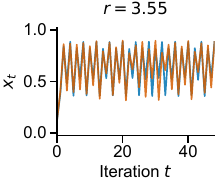}
\caption{Period 8}
\end{subfigure}
\hfill
\begin{subfigure}[t]{.19\textwidth}
\centering
\includegraphics[width=\linewidth]{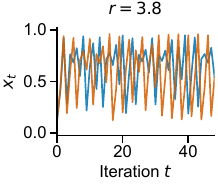}
\caption{Chaotic}
\end{subfigure}
\caption{\textbf{Matched trajectories across regimes.}
Logistic ground truth and the one-state Transformer after 10,000 updates share seed 42, identical parameters, and the same initial state.
Columns progress from a fixed point through period-2, period-4, and period-8 reference regimes to a chaotic reference.
Each panel shows 48 autoregressive updates including transients.
Blue marks ground truth and orange the Transformer.
The chaotic illustration uses $r=3.8$, where both maps have positive finite-orbit Lyapunov estimates; the appendix also shows the final 20k model's period-3 window at $r=3.9$.
Appendix~\ref{app:cobweb} repeats these views across both systems and training stages.}
\label{fig:phases}
\end{figure}

\begin{figure}[!htbp]
\centering
\begin{subfigure}[t]{.19\textwidth}
\centering
\includegraphics[width=\linewidth]{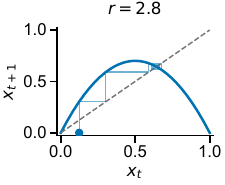}\\[-0.15em]
\includegraphics[width=\linewidth]{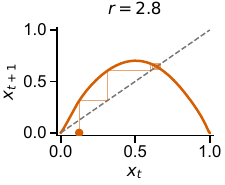}
\caption{Stable}
\end{subfigure}
\hfill
\begin{subfigure}[t]{.19\textwidth}
\centering
\includegraphics[width=\linewidth]{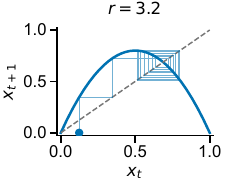}\\[-0.15em]
\includegraphics[width=\linewidth]{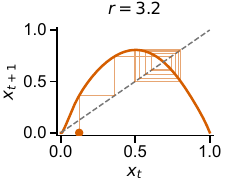}
\caption{Period 2}
\end{subfigure}
\hfill
\begin{subfigure}[t]{.19\textwidth}
\centering
\includegraphics[width=\linewidth]{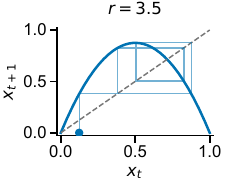}\\[-0.15em]
\includegraphics[width=\linewidth]{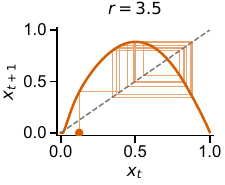}
\caption{Period 4}
\end{subfigure}
\hfill
\begin{subfigure}[t]{.19\textwidth}
\centering
\includegraphics[width=\linewidth]{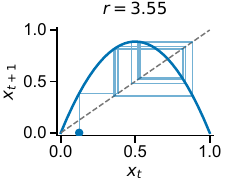}\\[-0.15em]
\includegraphics[width=\linewidth]{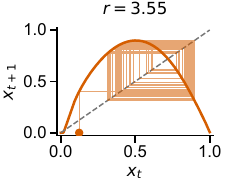}
\caption{Period 8}
\end{subfigure}
\hfill
\begin{subfigure}[t]{.19\textwidth}
\centering
\includegraphics[width=\linewidth]{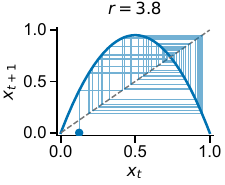}\\[-0.15em]
\includegraphics[width=\linewidth]{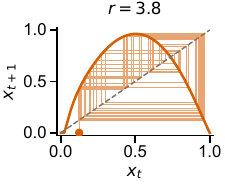}
\caption{Chaotic}
\end{subfigure}
\caption{\textbf{Reference and learned cobwebs for the same regimes.}
Each column stacks the ground-truth cobweb above the Transformer cobweb at the matching parameter after 10,000 updates.
Solid curves show the corresponding scalar map and dashed lines mark the identity.
Blue marks ground truth and orange the Transformer.
Repeated iteration turns the learned transition into stable, periodic, or irregular motion.
The chaotic illustration uses $r=3.8$, where both maps have positive finite-orbit Lyapunov estimates; the appendix also shows the final 20k model's period-3 window at $r=3.9$.}
\label{fig:phases-cobweb}
\end{figure}
\textbf{Chaotic Dynamics and Periodic Windows.}
Figure~\ref{fig:phases} and Figure~\ref{fig:phases-cobweb} use logistic $r=3.8$ to illustrate chaotic behavior in both maps at the 10,000-update checkpoint.
After 2,048 burn-in steps, 2,048 further iterates give Lyapunov estimates of 0.422 for ground truth and 0.431 for the Transformer.
The model derivative uses a centered finite difference through the clamped scalar map.
The atlas keeps $r=3.9$ throughout training: there, the final model converges to period 3 (maximum lag-3 residual below $10^{-9}$ over the last 64 states) with a Lyapunov estimate of $-0.099$, although the reference is chaotic.
This contrast exposes a displaced periodic window rather than presenting every chaotic-reference parameter as successful chaotic reproduction.
These illustrative parameters do not select a checkpoint or alter the full-grid metrics.
These views connect the transition function to its repeated application. All panels use seed 42, one state token ($C=1$), and the same initial state $x_0=0.123456789$. Blue denotes ground truth and orange the Transformer. A cobweb alternates a vertical move to the map and a horizontal move to the identity line; repeated moves show convergence or recurrent motion. Solid curves show the reference or learned map and dashed gray lines show the identity. The learned map includes the same $[0,1]$ output clamp as autonomous evaluation. We compute it directly from each checkpoint on 1,001 state values. No map is fitted to the plotted trajectories.
For each system we fix five physical parameters: logistic $r=2.8,3.2,3.5,3.55,3.9$ and sine $r=0.6,0.8,0.845,0.861,0.99$. Column titles name the \emph{reference} regime. Training covers the stable-reference parameters, including transient trajectories; the periodic and chaotic reference parameters are outside the training interval. We retain all 128 iterates from initialization in the source data. Plots display the first 48 updates without burn-in. Float64 evaluation resolves the scalar geometry; primary aggregate fidelity remains the original float32 evaluation.
We show nine stages for both systems: initialization, 500, 1,000, 2,000, 3,000, 4,000, 6,000, 10,000, and 20,000 updates. Trajectory panels are grouped by system, then cobweb panels. Within each group a single caption closes the figure; row labels mark the stage and column titles mark the regime.

\begin{figure*}[!htbp]
\centering
\resizebox{\textwidth}{!}{%
\setlength{\tabcolsep}{2.0pt}
\renewcommand{\arraystretch}{0.75}
\begin{tabular}{@{}rccccc@{}}
 & \footnotesize Stable & \footnotesize Period 2 & \footnotesize Period 4 & \footnotesize Period 8 & \footnotesize Chaotic \\[-2pt]
\raisebox{0.55cm}{\footnotesize Init.} & \includegraphics[width=0.175\textwidth]{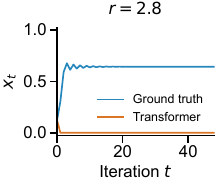} & \includegraphics[width=0.175\textwidth]{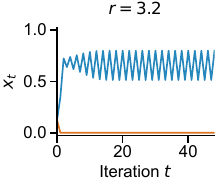} & \includegraphics[width=0.175\textwidth]{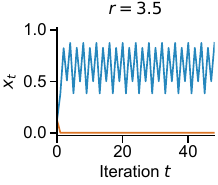} & \includegraphics[width=0.175\textwidth]{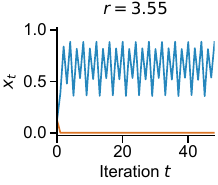} & \includegraphics[width=0.175\textwidth]{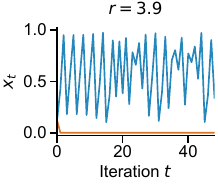} \\[0pt]
\raisebox{0.55cm}{\footnotesize 500} & \includegraphics[width=0.175\textwidth]{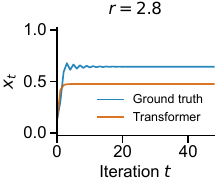} & \includegraphics[width=0.175\textwidth]{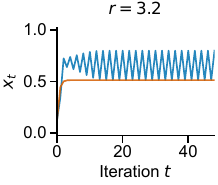} & \includegraphics[width=0.175\textwidth]{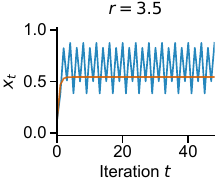} & \includegraphics[width=0.175\textwidth]{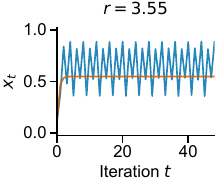} & \includegraphics[width=0.175\textwidth]{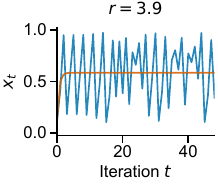} \\[0pt]
\raisebox{0.55cm}{\footnotesize 1k} & \includegraphics[width=0.175\textwidth]{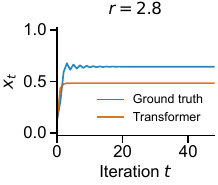} & \includegraphics[width=0.175\textwidth]{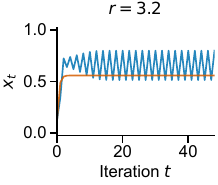} & \includegraphics[width=0.175\textwidth]{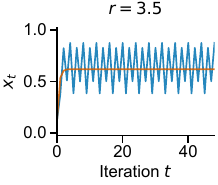} & \includegraphics[width=0.175\textwidth]{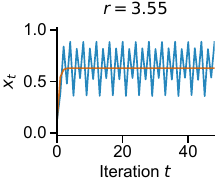} & \includegraphics[width=0.175\textwidth]{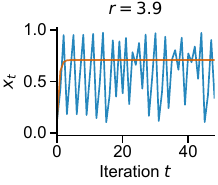} \\[0pt]
\raisebox{0.55cm}{\footnotesize 2k} & \includegraphics[width=0.175\textwidth]{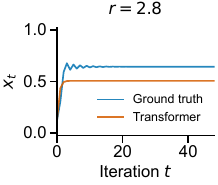} & \includegraphics[width=0.175\textwidth]{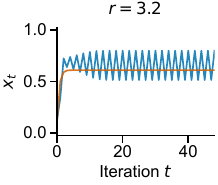} & \includegraphics[width=0.175\textwidth]{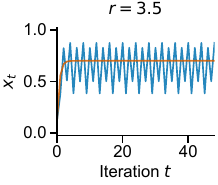} & \includegraphics[width=0.175\textwidth]{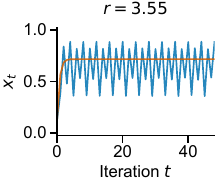} & \includegraphics[width=0.175\textwidth]{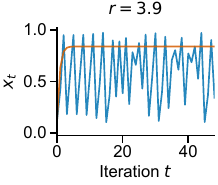} \\[0pt]
\raisebox{0.55cm}{\footnotesize 3k} & \includegraphics[width=0.175\textwidth]{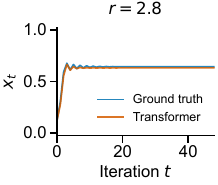} & \includegraphics[width=0.175\textwidth]{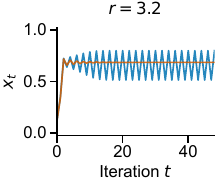} & \includegraphics[width=0.175\textwidth]{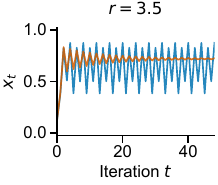} & \includegraphics[width=0.175\textwidth]{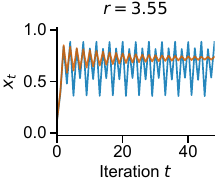} & \includegraphics[width=0.175\textwidth]{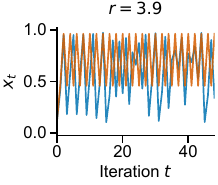} \\[0pt]
\raisebox{0.55cm}{\footnotesize 4k} & \includegraphics[width=0.175\textwidth]{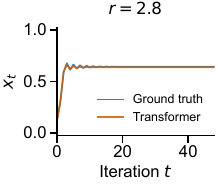} & \includegraphics[width=0.175\textwidth]{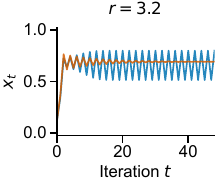} & \includegraphics[width=0.175\textwidth]{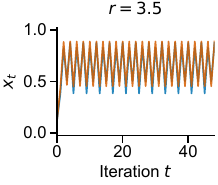} & \includegraphics[width=0.175\textwidth]{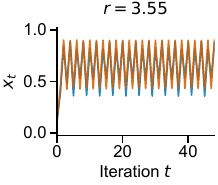} & \includegraphics[width=0.175\textwidth]{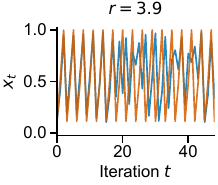} \\[0pt]
\raisebox{0.55cm}{\footnotesize 6k} & \includegraphics[width=0.175\textwidth]{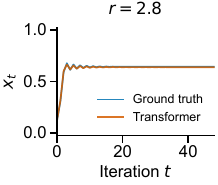} & \includegraphics[width=0.175\textwidth]{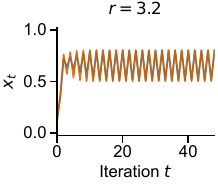} & \includegraphics[width=0.175\textwidth]{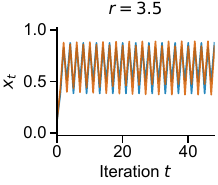} & \includegraphics[width=0.175\textwidth]{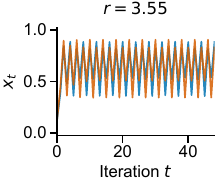} & \includegraphics[width=0.175\textwidth]{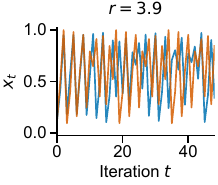} \\[0pt]
\raisebox{0.55cm}{\footnotesize 10k} & \includegraphics[width=0.175\textwidth]{img/cobweb_logistic_010000_trajectory_0.pdf} & \includegraphics[width=0.175\textwidth]{img/cobweb_logistic_010000_trajectory_1.pdf} & \includegraphics[width=0.175\textwidth]{img/cobweb_logistic_010000_trajectory_2.pdf} & \includegraphics[width=0.175\textwidth]{img/cobweb_logistic_010000_trajectory_3.pdf} & \includegraphics[width=0.175\textwidth]{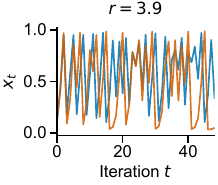} \\[0pt]
\raisebox{0.55cm}{\footnotesize 20k} & \includegraphics[width=0.175\textwidth]{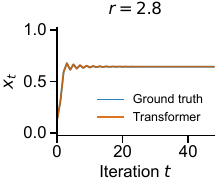} & \includegraphics[width=0.175\textwidth]{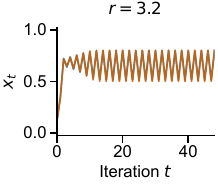} & \includegraphics[width=0.175\textwidth]{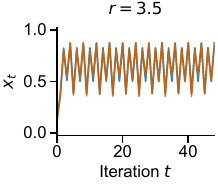} & \includegraphics[width=0.175\textwidth]{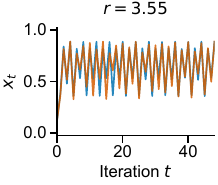} & \includegraphics[width=0.175\textwidth]{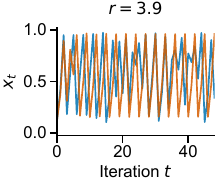} \\[0pt]
\end{tabular}%
}
\caption{\textbf{Logistic trajectories across training.} Each row is one checkpoint; columns are the same five reference regimes. Blue denotes ground truth and orange the Transformer, with the legend only in the top-left panel. Regime names describe the reference, not a classification inferred for the checkpoint.}
\label{fig:cobweb-logistic-traj}
\end{figure*}

\begin{figure*}[!htbp]
\centering
\resizebox{\textwidth}{!}{%
\setlength{\tabcolsep}{2.0pt}
\renewcommand{\arraystretch}{0.75}
\begin{tabular}{@{}rccccc@{}}
 & \footnotesize Stable & \footnotesize Period 2 & \footnotesize Period 4 & \footnotesize Period 8 & \footnotesize Chaotic \\[-2pt]
\raisebox{0.55cm}{\footnotesize Init.} & \includegraphics[width=0.175\textwidth]{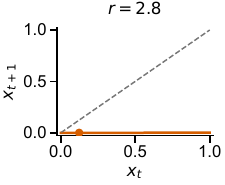} & \includegraphics[width=0.175\textwidth]{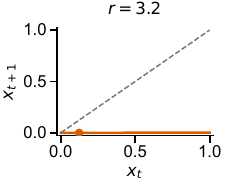} & \includegraphics[width=0.175\textwidth]{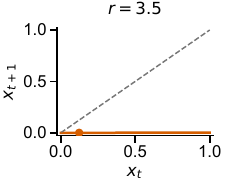} & \includegraphics[width=0.175\textwidth]{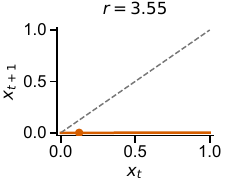} & \includegraphics[width=0.175\textwidth]{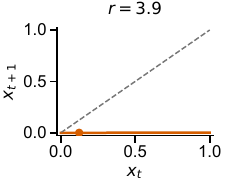} \\[0pt]
\raisebox{0.55cm}{\footnotesize 500} & \includegraphics[width=0.175\textwidth]{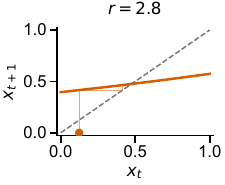} & \includegraphics[width=0.175\textwidth]{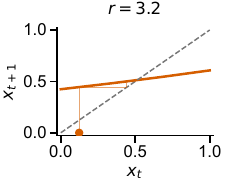} & \includegraphics[width=0.175\textwidth]{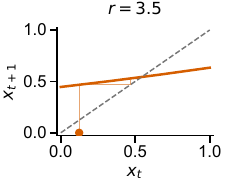} & \includegraphics[width=0.175\textwidth]{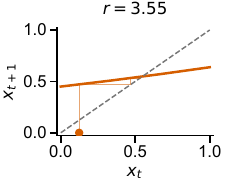} & \includegraphics[width=0.175\textwidth]{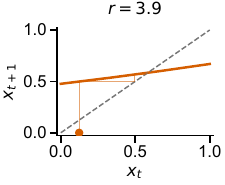} \\[0pt]
\raisebox{0.55cm}{\footnotesize 1k} & \includegraphics[width=0.175\textwidth]{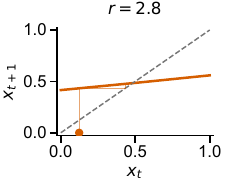} & \includegraphics[width=0.175\textwidth]{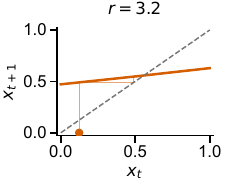} & \includegraphics[width=0.175\textwidth]{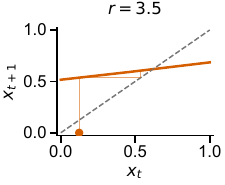} & \includegraphics[width=0.175\textwidth]{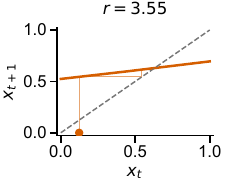} & \includegraphics[width=0.175\textwidth]{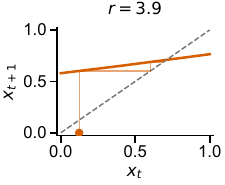} \\[0pt]
\raisebox{0.55cm}{\footnotesize 2k} & \includegraphics[width=0.175\textwidth]{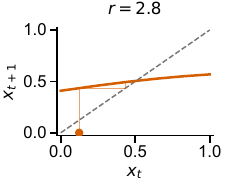} & \includegraphics[width=0.175\textwidth]{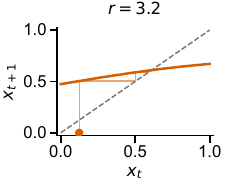} & \includegraphics[width=0.175\textwidth]{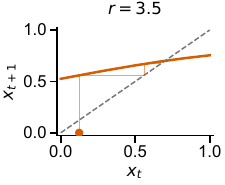} & \includegraphics[width=0.175\textwidth]{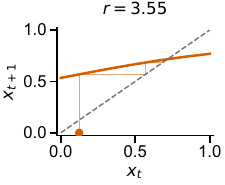} & \includegraphics[width=0.175\textwidth]{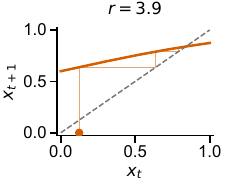} \\[0pt]
\raisebox{0.55cm}{\footnotesize 3k} & \includegraphics[width=0.175\textwidth]{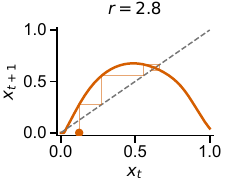} & \includegraphics[width=0.175\textwidth]{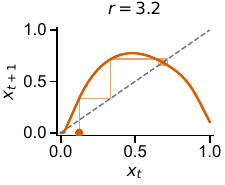} & \includegraphics[width=0.175\textwidth]{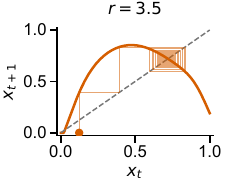} & \includegraphics[width=0.175\textwidth]{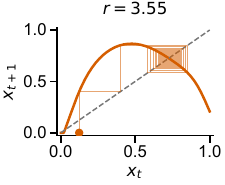} & \includegraphics[width=0.175\textwidth]{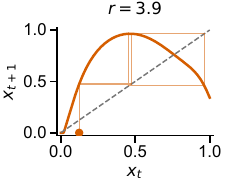} \\[0pt]
\raisebox{0.55cm}{\footnotesize 4k} & \includegraphics[width=0.175\textwidth]{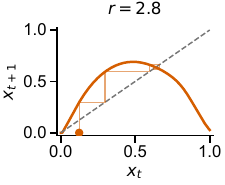} & \includegraphics[width=0.175\textwidth]{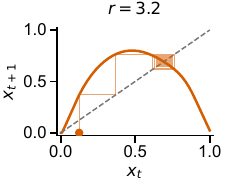} & \includegraphics[width=0.175\textwidth]{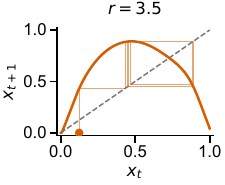} & \includegraphics[width=0.175\textwidth]{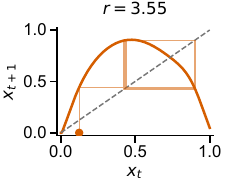} & \includegraphics[width=0.175\textwidth]{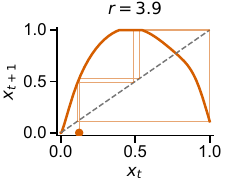} \\[0pt]
\raisebox{0.55cm}{\footnotesize 6k} & \includegraphics[width=0.175\textwidth]{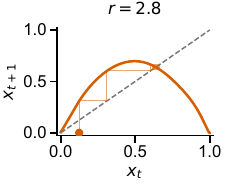} & \includegraphics[width=0.175\textwidth]{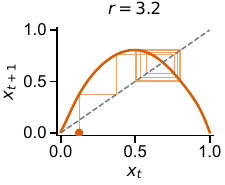} & \includegraphics[width=0.175\textwidth]{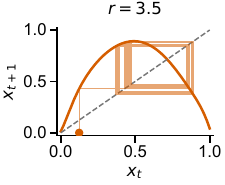} & \includegraphics[width=0.175\textwidth]{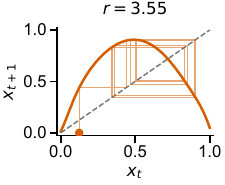} & \includegraphics[width=0.175\textwidth]{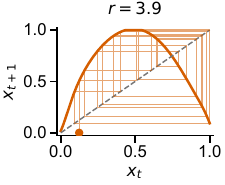} \\[0pt]
\raisebox{0.55cm}{\footnotesize 10k} & \includegraphics[width=0.175\textwidth]{img/cobweb_logistic_010000_model_0.pdf} & \includegraphics[width=0.175\textwidth]{img/cobweb_logistic_010000_model_1.pdf} & \includegraphics[width=0.175\textwidth]{img/cobweb_logistic_010000_model_2.pdf} & \includegraphics[width=0.175\textwidth]{img/cobweb_logistic_010000_model_3.pdf} & \includegraphics[width=0.175\textwidth]{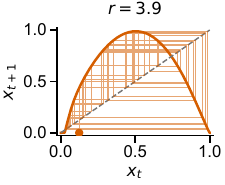} \\[0pt]
\raisebox{0.55cm}{\footnotesize 20k} & \includegraphics[width=0.175\textwidth]{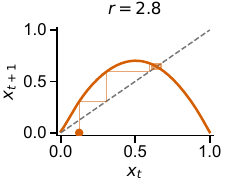} & \includegraphics[width=0.175\textwidth]{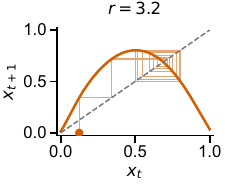} & \includegraphics[width=0.175\textwidth]{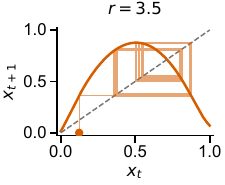} & \includegraphics[width=0.175\textwidth]{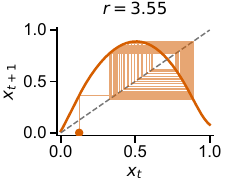} & \includegraphics[width=0.175\textwidth]{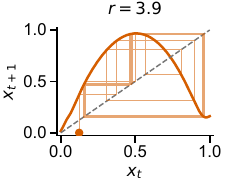} \\[0pt]
\raisebox{0.55cm}{\footnotesize GT} & \includegraphics[width=0.175\textwidth]{img/cobweb_logistic_truth_0.pdf} & \includegraphics[width=0.175\textwidth]{img/cobweb_logistic_truth_1.pdf} & \includegraphics[width=0.175\textwidth]{img/cobweb_logistic_truth_2.pdf} & \includegraphics[width=0.175\textwidth]{img/cobweb_logistic_truth_3.pdf} & \includegraphics[width=0.175\textwidth]{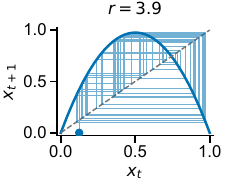} \\[0pt]
\end{tabular}%
}
\caption{\textbf{Logistic cobwebs across training.} Each row shows the Transformer at the labeled stage; the final row is ground truth (GT). Columns keep the same reference regimes. Blue marks ground truth and orange the Transformer.}
\label{fig:cobweb-logistic}
\end{figure*}

\begin{figure*}[!htbp]
\centering
\resizebox{\textwidth}{!}{%
\setlength{\tabcolsep}{2.0pt}
\renewcommand{\arraystretch}{0.75}
\begin{tabular}{@{}rccccc@{}}
 & \footnotesize Stable & \footnotesize Period 2 & \footnotesize Period 4 & \footnotesize Period 8 & \footnotesize Chaotic \\[-2pt]
\raisebox{0.55cm}{\footnotesize Init.} & \includegraphics[width=0.175\textwidth]{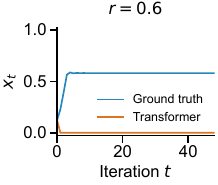} & \includegraphics[width=0.175\textwidth]{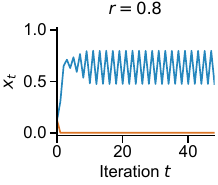} & \includegraphics[width=0.175\textwidth]{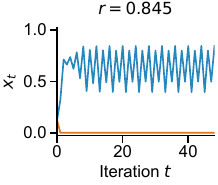} & \includegraphics[width=0.175\textwidth]{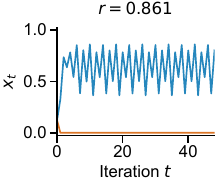} & \includegraphics[width=0.175\textwidth]{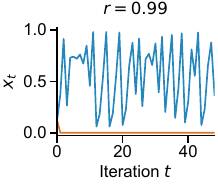} \\[0pt]
\raisebox{0.55cm}{\footnotesize 500} & \includegraphics[width=0.175\textwidth]{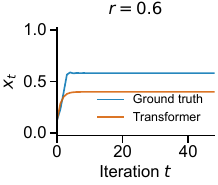} & \includegraphics[width=0.175\textwidth]{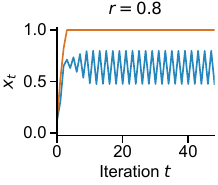} & \includegraphics[width=0.175\textwidth]{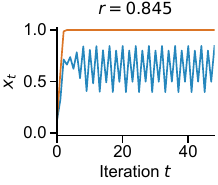} & \includegraphics[width=0.175\textwidth]{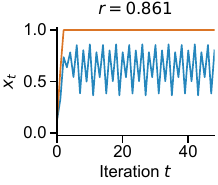} & \includegraphics[width=0.175\textwidth]{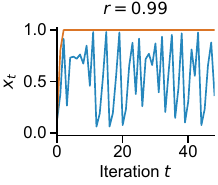} \\[0pt]
\raisebox{0.55cm}{\footnotesize 1k} & \includegraphics[width=0.175\textwidth]{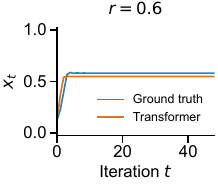} & \includegraphics[width=0.175\textwidth]{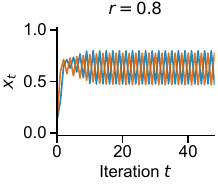} & \includegraphics[width=0.175\textwidth]{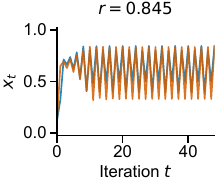} & \includegraphics[width=0.175\textwidth]{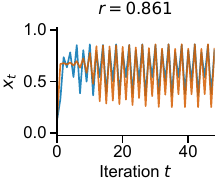} & \includegraphics[width=0.175\textwidth]{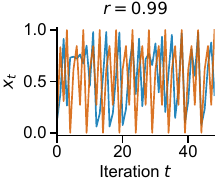} \\[0pt]
\raisebox{0.55cm}{\footnotesize 2k} & \includegraphics[width=0.175\textwidth]{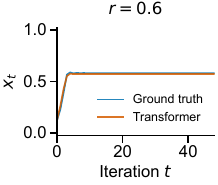} & \includegraphics[width=0.175\textwidth]{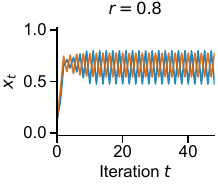} & \includegraphics[width=0.175\textwidth]{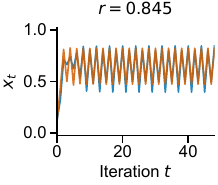} & \includegraphics[width=0.175\textwidth]{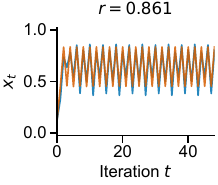} & \includegraphics[width=0.175\textwidth]{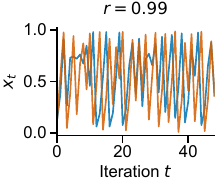} \\[0pt]
\raisebox{0.55cm}{\footnotesize 3k} & \includegraphics[width=0.175\textwidth]{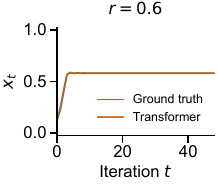} & \includegraphics[width=0.175\textwidth]{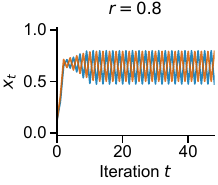} & \includegraphics[width=0.175\textwidth]{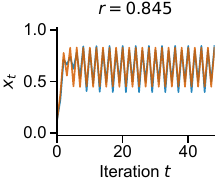} & \includegraphics[width=0.175\textwidth]{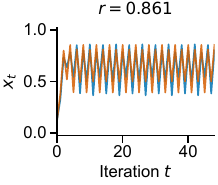} & \includegraphics[width=0.175\textwidth]{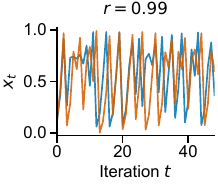} \\[0pt]
\raisebox{0.55cm}{\footnotesize 4k} & \includegraphics[width=0.175\textwidth]{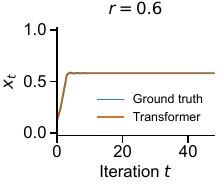} & \includegraphics[width=0.175\textwidth]{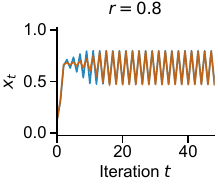} & \includegraphics[width=0.175\textwidth]{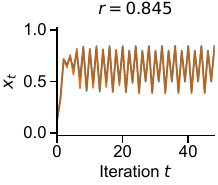} & \includegraphics[width=0.175\textwidth]{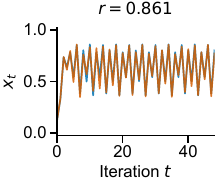} & \includegraphics[width=0.175\textwidth]{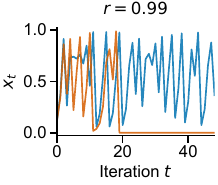} \\[0pt]
\raisebox{0.55cm}{\footnotesize 6k} & \includegraphics[width=0.175\textwidth]{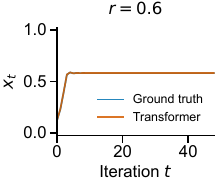} & \includegraphics[width=0.175\textwidth]{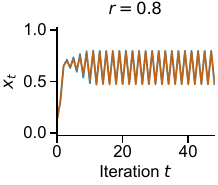} & \includegraphics[width=0.175\textwidth]{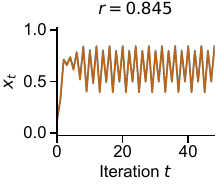} & \includegraphics[width=0.175\textwidth]{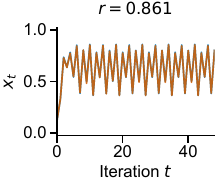} & \includegraphics[width=0.175\textwidth]{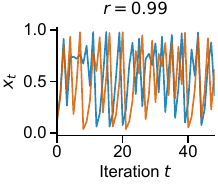} \\[0pt]
\raisebox{0.55cm}{\footnotesize 10k} & \includegraphics[width=0.175\textwidth]{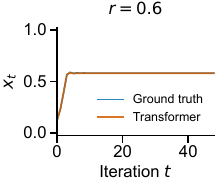} & \includegraphics[width=0.175\textwidth]{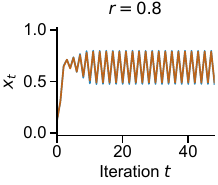} & \includegraphics[width=0.175\textwidth]{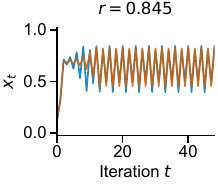} & \includegraphics[width=0.175\textwidth]{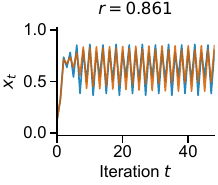} & \includegraphics[width=0.175\textwidth]{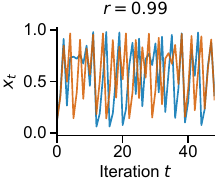} \\[0pt]
\raisebox{0.55cm}{\footnotesize 20k} & \includegraphics[width=0.175\textwidth]{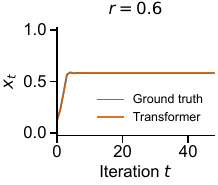} & \includegraphics[width=0.175\textwidth]{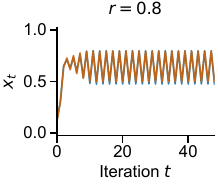} & \includegraphics[width=0.175\textwidth]{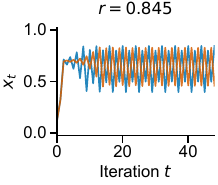} & \includegraphics[width=0.175\textwidth]{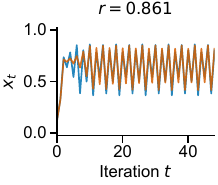} & \includegraphics[width=0.175\textwidth]{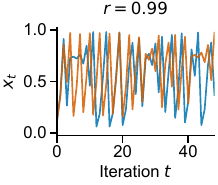} \\[0pt]
\end{tabular}%
}
\caption{\textbf{Sine trajectories across training.} Each row is one checkpoint; columns are the same five reference regimes. Blue denotes ground truth and orange the Transformer, with the legend only in the top-left panel. Regime names describe the reference, not a classification inferred for the checkpoint.}
\label{fig:cobweb-sine-traj}
\end{figure*}

\begin{figure*}[!htbp]
\centering
\resizebox{\textwidth}{!}{%
\setlength{\tabcolsep}{2.0pt}
\renewcommand{\arraystretch}{0.75}
\begin{tabular}{@{}rccccc@{}}
 & \footnotesize Stable & \footnotesize Period 2 & \footnotesize Period 4 & \footnotesize Period 8 & \footnotesize Chaotic \\[-2pt]
\raisebox{0.55cm}{\footnotesize Init.} & \includegraphics[width=0.175\textwidth]{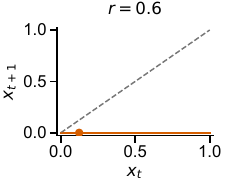} & \includegraphics[width=0.175\textwidth]{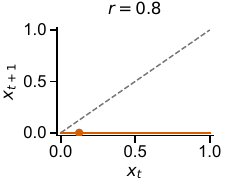} & \includegraphics[width=0.175\textwidth]{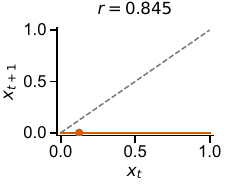} & \includegraphics[width=0.175\textwidth]{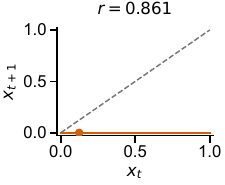} & \includegraphics[width=0.175\textwidth]{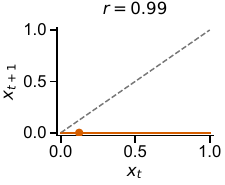} \\[0pt]
\raisebox{0.55cm}{\footnotesize 500} & \includegraphics[width=0.175\textwidth]{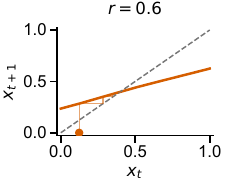} & \includegraphics[width=0.175\textwidth]{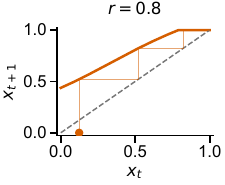} & \includegraphics[width=0.175\textwidth]{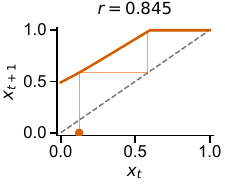} & \includegraphics[width=0.175\textwidth]{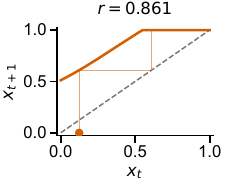} & \includegraphics[width=0.175\textwidth]{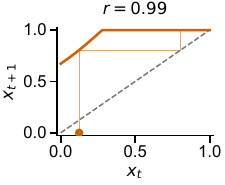} \\[0pt]
\raisebox{0.55cm}{\footnotesize 1k} & \includegraphics[width=0.175\textwidth]{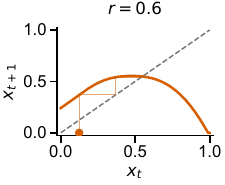} & \includegraphics[width=0.175\textwidth]{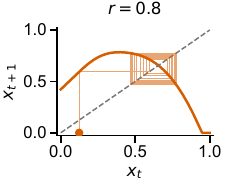} & \includegraphics[width=0.175\textwidth]{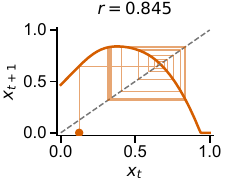} & \includegraphics[width=0.175\textwidth]{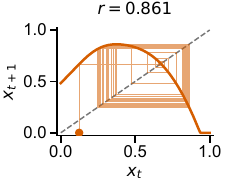} & \includegraphics[width=0.175\textwidth]{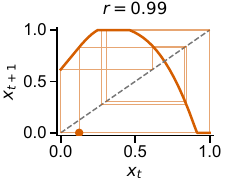} \\[0pt]
\raisebox{0.55cm}{\footnotesize 2k} & \includegraphics[width=0.175\textwidth]{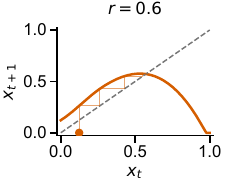} & \includegraphics[width=0.175\textwidth]{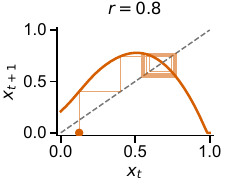} & \includegraphics[width=0.175\textwidth]{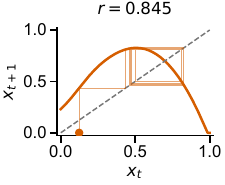} & \includegraphics[width=0.175\textwidth]{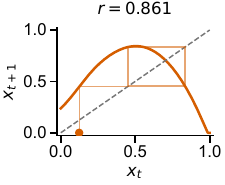} & \includegraphics[width=0.175\textwidth]{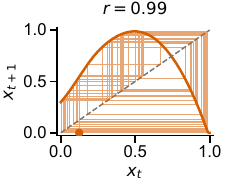} \\[0pt]
\raisebox{0.55cm}{\footnotesize 3k} & \includegraphics[width=0.175\textwidth]{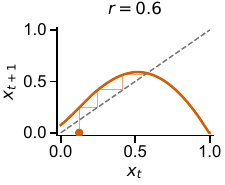} & \includegraphics[width=0.175\textwidth]{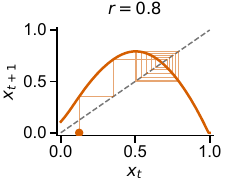} & \includegraphics[width=0.175\textwidth]{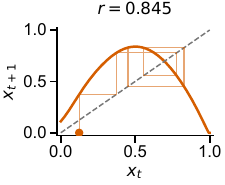} & \includegraphics[width=0.175\textwidth]{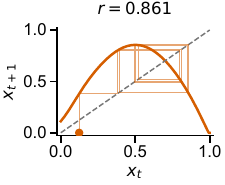} & \includegraphics[width=0.175\textwidth]{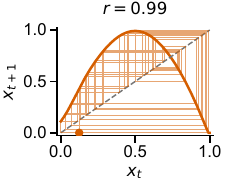} \\[0pt]
\raisebox{0.55cm}{\footnotesize 4k} & \includegraphics[width=0.175\textwidth]{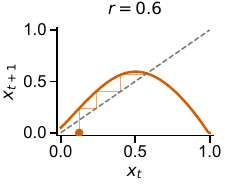} & \includegraphics[width=0.175\textwidth]{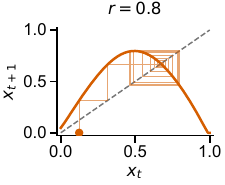} & \includegraphics[width=0.175\textwidth]{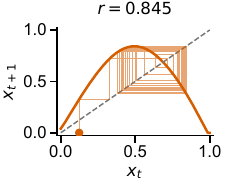} & \includegraphics[width=0.175\textwidth]{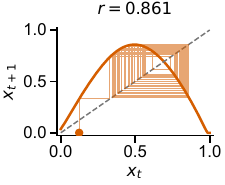} & \includegraphics[width=0.175\textwidth]{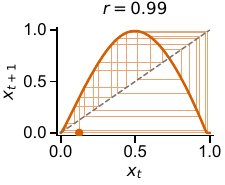} \\[0pt]
\raisebox{0.55cm}{\footnotesize 6k} & \includegraphics[width=0.175\textwidth]{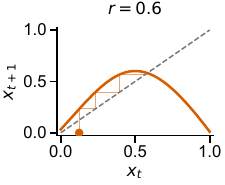} & \includegraphics[width=0.175\textwidth]{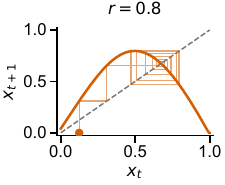} & \includegraphics[width=0.175\textwidth]{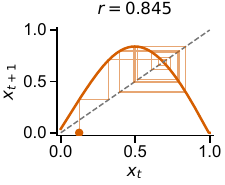} & \includegraphics[width=0.175\textwidth]{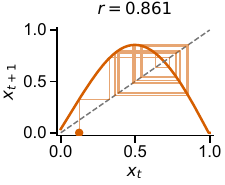} & \includegraphics[width=0.175\textwidth]{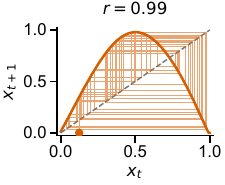} \\[0pt]
\raisebox{0.55cm}{\footnotesize 10k} & \includegraphics[width=0.175\textwidth]{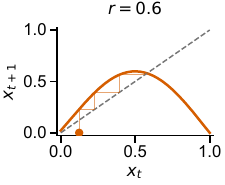} & \includegraphics[width=0.175\textwidth]{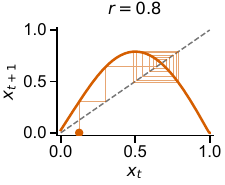} & \includegraphics[width=0.175\textwidth]{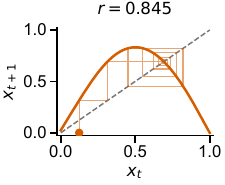} & \includegraphics[width=0.175\textwidth]{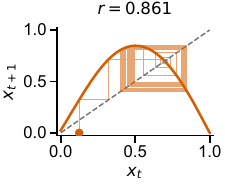} & \includegraphics[width=0.175\textwidth]{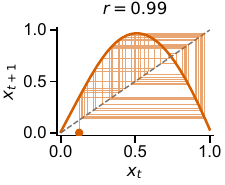} \\[0pt]
\raisebox{0.55cm}{\footnotesize 20k} & \includegraphics[width=0.175\textwidth]{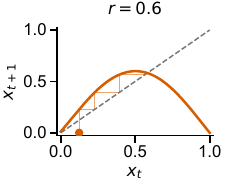} & \includegraphics[width=0.175\textwidth]{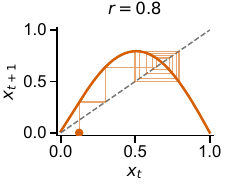} & \includegraphics[width=0.175\textwidth]{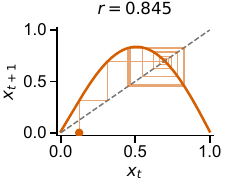} & \includegraphics[width=0.175\textwidth]{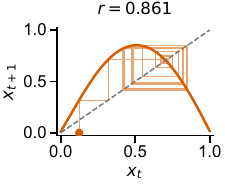} & \includegraphics[width=0.175\textwidth]{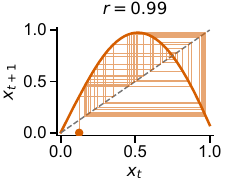} \\[0pt]
\raisebox{0.55cm}{\footnotesize GT} & \includegraphics[width=0.175\textwidth]{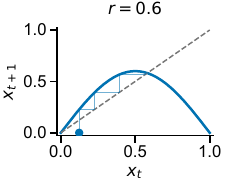} & \includegraphics[width=0.175\textwidth]{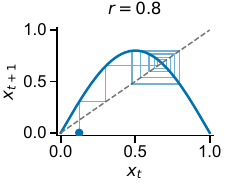} & \includegraphics[width=0.175\textwidth]{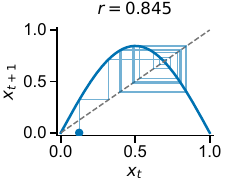} & \includegraphics[width=0.175\textwidth]{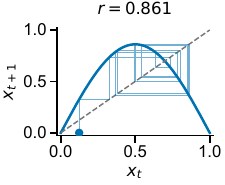} & \includegraphics[width=0.175\textwidth]{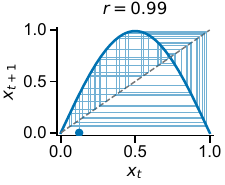} \\[0pt]
\end{tabular}%
}
\caption{\textbf{Sine cobwebs across training.} Each row shows the Transformer at the labeled stage; the final row is ground truth. Columns keep the same reference regimes. Blue marks ground truth and orange the Transformer.}
\label{fig:cobweb-sine}
\end{figure*}

\subsection{Dependence on the Observation Window}
\label{app:window}
The ablation in Section~\ref{sec:window} retrains the same scalar Transformer on four training
parameter windows with everything else fixed: seed 42--44, 20,000 AdamW updates, batch size 512,
the short-sequence length distribution of Appendix~\ref{app:protocol}, and the final checkpoint.
Only the interval from which parameters are drawn changes; no evaluation signal enters training or
checkpoint selection. Each run takes 209--274\,s on a single CPU machine in our reproduction. The
evaluation reuses the 513-point physical grid, three fixed initial states, 1,024 discarded steps and
512 retained states of Appendix~\ref{app:evaluation}.

Table~\ref{tab:window} summarizes the four windows, and Table~\ref{tab:window-arms} lists every seed.

\begin{table}[!htbp]
\centering
\small
\caption{\textbf{Per-seed window ablation.} Same protocol for every row. $W_1$ is the uniform-grid
OOD average; agreement is over reference-periodic parameter--initial-state pairs; coarse flips count
consecutive detected doublings on the 513-point grid.}
\label{tab:window-arms}
\vspace{-1ex}
\begin{tabular}{lrrrrr}
\toprule
Training window & Seed & ID $W_1$ & OOD $W_1$ & Agreement & Coarse flips\\
\midrule
$[2.0,3.0]$ & 42 & 0.00179 & 0.03456 & 70.2\% & 3\\
            & 43 & 0.00050 & 0.02772 & 84.0\% & 3\\
            & 44 & 0.00045 & 0.02108 & 81.3\% & 3\\
$[2.5,3.0]$ & 42 & 0.00077 & 0.03172 & 90.0\% & 3\\
            & 43 & 0.00073 & 0.03940 & 88.2\% & 3\\
            & 44 & 0.00389 & 0.06319 & 76.9\% & 0\\
$[2.9,3.0]$ & 42 & 0.00415 & 0.03756 & 85.1\% & 3\\
            & 43 & 0.00138 & 0.04022 & 86.9\% & 3\\
            & 44 & 0.00149 & 0.03047 & 92.6\% & 3\\
$[2.0,2.5]$ & 42 & 0.00100 & 0.06865 & 63.9\% & 3\\
            & 43 & 0.00022 & 0.16082 & 67.8\% & 3\\
            & 44 & 0.00068 & 0.06104 & 75.5\% & 3\\
\bottomrule
\end{tabular}
\end{table}

\begin{table}[!htbp]
\centering
\small
\caption{\textbf{Observation-window ablation.} All four windows use the locked protocol of Appendix~\ref{app:protocol} and are evaluated on the same 513-point grid. OOD $W_1$ is in units of $10^{-3}$, mean $\pm$ sample SD over three training seeds. ``Verified flips'' counts consecutive flips passing the local root and multiplier checks for seed 42 (three refinement rounds for every window).}
\label{tab:window}
\vspace{-1ex}
\begin{tabular}{lcccc}
\toprule
Training window & OOD $W_1$ & Periodic agreement & First verified flip & Verified flips\\
\midrule
$[2.0,3.0]$ & $27.8\pm6.7$ & 78.5\% & 2.9692 & 7\\
$[2.5,3.0]$ & $44.8\pm16.4$ & 85.0\% & 3.0170 & 6\\
$[2.9,3.0]$ & $36.1\pm5.0$ & 88.2\% & 2.9884 & 6\\
$[2.0,2.5]$ & $96.8\pm55.5$ & 69.1\% & 2.9304 & 6\\
\midrule
Reference & --- & --- & 3.0000 & 7\\
\bottomrule
\end{tabular}
\end{table}

\textbf{Bifurcation Verification across Training Windows.}
Each window was refined with the blind candidate search, paired burn-in budgets and phase-center
agreement of Appendix~\ref{app:evaluation}, followed by the independent closure-plus-multiplier
solve. Three refinement rounds were used for every arm, whereas the original $[2,3]$ training condition used
four levels, so the count of verified flips is budget-limited for the new arms rather than
capability-limited.

\begin{table}[!htbp]
\centering
\small
\caption{\textbf{Verified flips per training window} (seed 42). Parameters are the flip parameters
verified by orbit closure and multiplier $-1$ with two-sided continuation; $\delta_n$ follows
Equation~\eqref{eq:ratio}. The last column shows the largest finite-order ratio reached in the
available levels.}
\label{tab:window-verify}
\vspace{-1ex}
\begin{tabular}{lllr}
\toprule
Training window & Verified flips & First three flip parameters & Largest $\delta_n$\\
\midrule
$[2.0,3.0]$ (published) & 7 & 2.9692235, 3.3927447, 3.4848703 & 4.6687\\
$[2.5,3.0]$ & 6 & 3.0169678, 3.4462607, 3.5237060 & 4.6731\\
$[2.9,3.0]$ & 6 & 2.9884133, 3.3770377, 3.4491587 & 4.6745\\
$[2.0,2.5]$ & 6 & 2.9303844, 3.2847299, 3.3538837 & 4.6739\\
\midrule
Logistic reference & 7 & 3.0000000, 3.4494897, 3.5440904 & 4.6691\\
\bottomrule
\end{tabular}
\end{table}

The early ratios differ between windows because the displaced first flip changes the interval
lengths from which the first ratios are formed; all arms converge towards the same same-definition
reference value as more levels are resolved. The window therefore controls where the cascade sits
and how fast the finite-order sequence approaches the reference, but not whether the cascade exists.

\textbf{Training and Verification Protocols.}
Training applies the locked protocol with the parameter interval overridden, and refinement reuses the candidate search, cross-budget period test, temporal confirmation and multiplier solve of Appendix~\ref{app:evaluation}.

\subsection{Variation across Seeds and Contexts}
\label{app:seed-context}
The four-state Transformer retains substantially better fidelity than the 255-state rollout. Period agreement on accepted reference trajectories is 73.7\% versus 24.5\% on logistic and 78.1\% versus 50.4\% on sine. Since the latter context far exceeds realized training lengths, this contrast measures sensitivity to length extrapolation together with the learned computation. It cannot by itself establish an intrinsic disadvantage of long-context attention, and the 255-state rows show in-distribution $W_1$ above the one-state out-of-distribution values on every seed, so they are best read as a configuration that fails in distribution rather than as a clean context-length probe.

The atlas pairs ground truth with every reported model, seed, and Transformer context. Each displayed panel is a separate PDF with an editable SVG and 600-dpi PNG counterpart; the document assembles panels using LaTeX subcaptions. The full-range plots use the original evaluation arrays and share physical axes, 512 state bins, and logarithmic counts from 1 to 1,536. Every retained observation contributes. Occupied bins use fixed 0.6-point square marks for final-size visibility, with color encoding the original count; no bins are smoothed or omitted. Training parameter intervals are shaded gray. State context $C$ excludes the parameter token; $C=255$ is a length-extrapolation stress test.

\subsubsection{Complete Transformer Trajectories}
Each system is shown once. Columns are ground truth and the three training seeds; rows are Transformer contexts $C=1,4,255$. Panels use the compact density style of Figure~\ref{fig:learning-evolution}.
\begin{figure*}[!htbp]
\centering
\setlength{\tabcolsep}{1.2pt}
\renewcommand{\arraystretch}{0.75}
\begin{tabular}{@{}rcccc@{}}
 & \footnotesize Ground truth & \footnotesize Seed 42 & \footnotesize Seed 43 & \footnotesize Seed 44 \\[-2pt]
\raisebox{0.45cm}{\footnotesize $C=1$} & \includegraphics[width=0.22\textwidth]{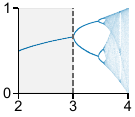} & \includegraphics[width=0.22\textwidth]{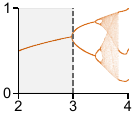} & \includegraphics[width=0.22\textwidth]{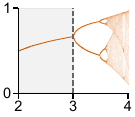} & \includegraphics[width=0.22\textwidth]{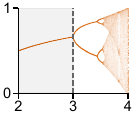} \\[0pt]
\raisebox{0.45cm}{\footnotesize $C=4$} & \includegraphics[width=0.22\textwidth]{img/grid_logistic_truth.pdf} & \includegraphics[width=0.22\textwidth]{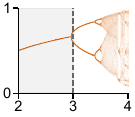} & \includegraphics[width=0.22\textwidth]{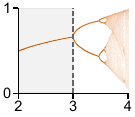} & \includegraphics[width=0.22\textwidth]{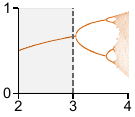} \\[0pt]
\raisebox{0.45cm}{\footnotesize $C=255$} & \includegraphics[width=0.22\textwidth]{img/grid_logistic_truth.pdf} & \includegraphics[width=0.22\textwidth]{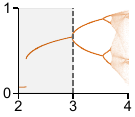} & \includegraphics[width=0.22\textwidth]{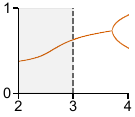} & \includegraphics[width=0.22\textwidth]{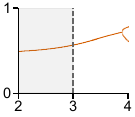} \\[0pt]
\end{tabular}
\caption{\textbf{Logistic Transformer seeds and contexts.} Columns share ground truth with seeds 42--44. Rows mark context length $C$. Gray shading marks the training interval.}
\label{fig:atlas-logistic-transformer}
\end{figure*}

\begin{figure*}[!htbp]
\centering
\setlength{\tabcolsep}{1.2pt}
\renewcommand{\arraystretch}{0.75}
\begin{tabular}{@{}rcccc@{}}
 & \footnotesize Ground truth & \footnotesize Seed 42 & \footnotesize Seed 43 & \footnotesize Seed 44 \\[-2pt]
\raisebox{0.45cm}{\footnotesize $C=1$} & \includegraphics[width=0.22\textwidth]{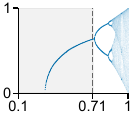} & \includegraphics[width=0.22\textwidth]{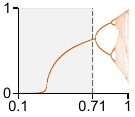} & \includegraphics[width=0.22\textwidth]{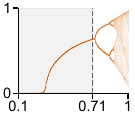} & \includegraphics[width=0.22\textwidth]{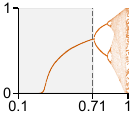} \\[0pt]
\raisebox{0.45cm}{\footnotesize $C=4$} & \includegraphics[width=0.22\textwidth]{img/grid_sine_truth.pdf} & \includegraphics[width=0.22\textwidth]{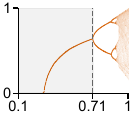} & \includegraphics[width=0.22\textwidth]{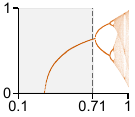} & \includegraphics[width=0.22\textwidth]{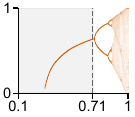} \\[0pt]
\raisebox{0.45cm}{\footnotesize $C=255$} & \includegraphics[width=0.22\textwidth]{img/grid_sine_truth.pdf} & \includegraphics[width=0.22\textwidth]{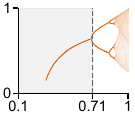} & \includegraphics[width=0.22\textwidth]{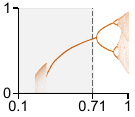} & \includegraphics[width=0.22\textwidth]{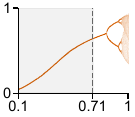} \\[0pt]
\end{tabular}
\caption{\textbf{Sine Transformer seeds and contexts.} Columns share ground truth with seeds 42--44. Rows mark context length $C$. Gray shading marks the training interval.}
\label{fig:atlas-sine-transformer}
\end{figure*}

\subsubsection{Parameter-Resolved Distributional Errors}
\begin{figure}[!htbp]
\centering
\begin{subfigure}[t]{.4\textwidth}
\centering
\includegraphics[width=\linewidth]{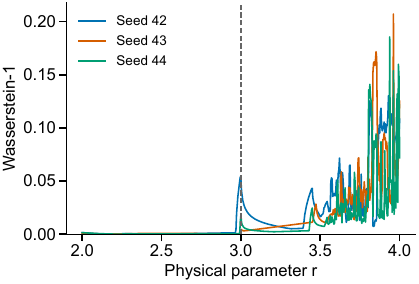}
\caption{Logistic}
\end{subfigure}
\quad
\begin{subfigure}[t]{.4\textwidth}
\centering
\includegraphics[width=\linewidth]{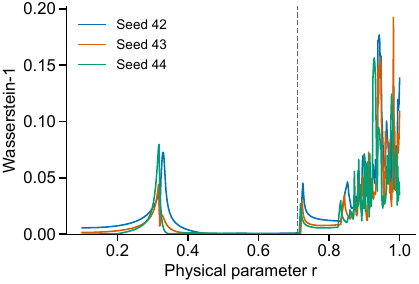}
\caption{Sine}
\end{subfigure}
\caption{\textbf{Parameter-resolved fidelity for Transformer $C=1$.} Lines show each seed separately, using every original uniform-grid Wasserstein-1 measurement. The dashed vertical line marks the training boundary. Vertical ranges adapt to the observed error and are labeled; these curves localize differences summarized by the main OOD table.}
\label{fig:error-c1}
\end{figure}

\begin{figure}[!htbp]
\centering
\begin{subfigure}[t]{.4\textwidth}
\centering
\includegraphics[width=\linewidth]{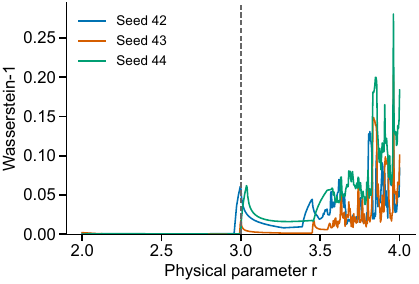}
\caption{Logistic}
\end{subfigure}
\quad
\begin{subfigure}[t]{.4\textwidth}
\centering
\includegraphics[width=\linewidth]{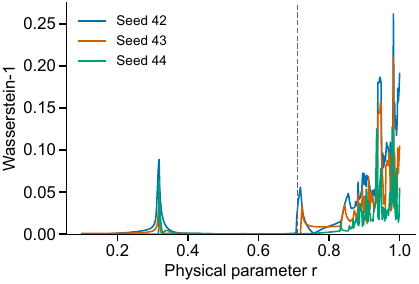}
\caption{Sine}
\end{subfigure}
\caption{\textbf{Parameter-resolved fidelity for Transformer $C=4$.} Lines show each seed separately, using every original uniform-grid Wasserstein-1 measurement. The dashed vertical line marks the training boundary. Vertical ranges adapt to the observed error and are labeled; these curves localize differences summarized by the main OOD table.}
\label{fig:error-c4}
\end{figure}

\begin{figure}[!htbp]
\centering
\begin{subfigure}[t]{.4\textwidth}
\centering
\includegraphics[width=\linewidth]{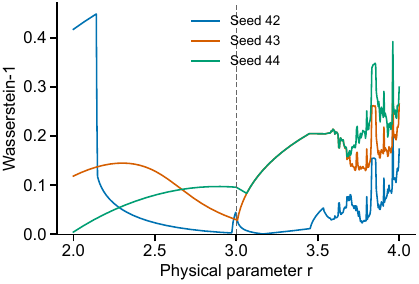}
\caption{Logistic}
\end{subfigure}
\quad
\begin{subfigure}[t]{.4\textwidth}
\centering
\includegraphics[width=\linewidth]{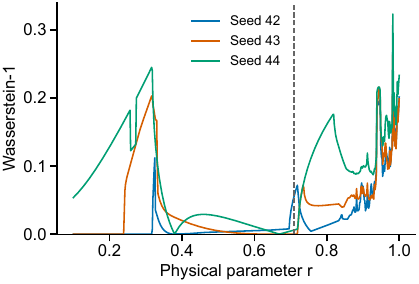}
\caption{Sine}
\end{subfigure}
\caption{\textbf{Parameter-resolved fidelity for Transformer $C=255$.} Lines show each seed separately, using every original uniform-grid Wasserstein-1 measurement. The dashed vertical line marks the training boundary. Vertical ranges adapt to the observed error and are labeled; these curves localize differences summarized by the main OOD table.}
\label{fig:error-c255}
\end{figure}

\subsection{Critical Geometry and Universality}
\label{app:universality}
A finite-order ratio close to $4.6692$ shows membership of a universality class, not identification
of a system: the ratio is fixed by the order of the map's maximum. This appendix separates the two,
using critical-point diagnostics and a quartic-family comparison. Function-level errors are reported in Appendix~\ref{app:map-accuracy}.

\textbf{Critical Exponent and Schwarzian Derivative.}
For each map we locate the maximum $x_c$ numerically, fit $\log|F(x)-F(x_c)|$ against
$\log|x-x_c|$ on both sides of $x_c$ for offsets $10^{-4}$ to $10^{-2}$, and evaluate the Schwarzian
derivative $SF=F'''/F'-1.5\,(F''/F')^2$ on the state interval where $|F'|>0.05$. A negative
Schwarzian together with a single critical point is the classical sufficient condition for a full
period-doubling cascade.

\begin{table}[!htbp]
\centering
\small
\caption{\textbf{Class diagnostics} at an out-of-distribution parameter. $z$ is the fitted order of
the maximum ($z=2$ quadratic, $z=4$ quartic). Derivatives are autograd first derivatives of the
float64 map; second and third derivatives are finite differences of that exact first derivative.}
\label{tab:class-diagnostics}
\vspace{-1ex}
\begin{tabular}{lccc}
\toprule
Map & Fitted $z$ & Reference $z$ & Schwarzian negative\\
\midrule
Logistic Transformer, $C=1$ & 2.000 & 2.000 & 97.3\% (reference 100\%)\\
Logistic MLP & 2.021 & 2.000 & 100\% (reference 100\%)\\
Sine Transformer, $C=1$ & 2.100 & 2.000 & 96.6\%\\
Unimodal $p=4$ Transformer & \textbf{2.005} & 3.932 & 100\%\\
\bottomrule
\end{tabular}
\end{table}

\textbf{Quartic Universality Class.}
We train the same architecture on the family $f(x)=1-\mu|x|^4$ (quartic maximum, pre-bifurcation
interval $[0.1,0.4785]$) and run the identical detection and verification pipeline on the reference
and on the learned map, evaluating the cascade region $\mu\in[0.4,1.7]$.

\begin{table}[!htbp]
\centering
\caption{\textbf{Quartic family.} Reference flips and finite-order ratios obtained with the
manuscript's pipeline; the literature value for the quartic class is $\delta\approx7.2847$, clearly
separated from the quadratic $4.6692$. The learned map is verified at one flip only, and its fitted
critical exponent is quadratic.}
\label{tab:class-z4}
\small
\vspace{-1ex}
\begin{tabular}{lcc}
\toprule
Quantity & Reference & Transformer\\
\midrule
Verified flips & 6 & 1\\
First flip & 0.4882812 & 0.4987569\\
Finite-order ratios ($n=2..4$) & 5.580, 6.963, 7.283 & ---\\
Fitted critical exponent $z$ & 3.932 & 2.005\\
Coarse flips in $\mu\in[0.4,1.7]$ & 3 & 1\\
OOD $W_1$ (three seeds) & --- & 0.0982, 0.3668, 0.0962\\
\bottomrule
\end{tabular}
\end{table}

The learned map has an approximately quadratic critical point despite being trained on a quartic family, and it fails to reproduce the quartic cascade: only one flip is verified. Thus no learned finite-order scaling ratio can be estimated in this experiment. Together with the quadratic-family results, this suggests that matching the Feigenbaum scaling in Section~\ref{sec:verified-cascade} reflects the universality class of the learned map rather than identification of the underlying analytic family.

\subsection{Transition Accuracy and Parameter Alignment}
\label{app:map-accuracy}

The same measurements give a direct reading of how well the learned transition matches the physical
one, beyond the distributional distances of Appendix~\ref{app:results}. Averages are over the
out-of-distribution parameter slice; the reference mean $|\partial_r F|$ is 0.173 for logistic and
0.660 for sine.

\begin{table}[!htbp]
\centering
\small
\caption{\textbf{Map-level diagnostics} (OOD slice, unclipped float64 map except
where noted). $\sup_x|F_\theta-F|$ is in physical state units; $\partial_r^2F_\theta$ is spurious
parameter curvature, identically zero for the reference.}
\label{tab:map-diagnostics}
\vspace{-1ex}
\begin{tabular}{lcccc}
\toprule
Map & $\sup_x|F_\theta-F|$ & mean $|\Delta\partial_xF|$ & mean $|\Delta\partial_rF|$ & mean $|\partial_r^2F_\theta|$\\
\midrule
Logistic Transformer, $C=1$ & 0.041 & 0.166 & 0.036 & 0.073\\
Logistic MLP & --- & 0.048 & 0.024 & 0.041\\
Sine Transformer, $C=1$ & --- & 0.079 & 0.121 & 0.638\\
Window $[2.9,3]$ & 0.102 & --- & 0.143 & 0.138\\
Window $[2,2.5]$ & 0.103 & --- & 0.143 & 0.138\\
Parameter token permuted & 0.255 & --- & 0.164 & 0.012\\
\bottomrule
\end{tabular}
\end{table}

Two features of this table matter for the interpretation of the main results. The parameter
derivative of the learned map differs from the physical one by roughly a fifth of the physical
scale, and the map carries a non-zero parameter curvature that the physical family does not have.
Both are consistent with a learned map that reproduces the organization of the family, including
its universality class, while remaining a smooth surrogate rather than a recovered equation.

\textbf{Parameter Alignment across Seeds.}
At the final checkpoint the three training seeds displace the same reference ladder in different directions, from $-0.060$ to $-0.031$ for seed 42, from $+0.001$ to $+0.036$ for seed 43 and from $-0.020$ to $-0.005$ for seed 44, an across-seed spread of $0.096$ in $r$ (Figure~\ref{fig:scaling}).

\section{Comparisons across Transition Representations}
\label{app:results}
We compare the Transformer with a direct parameter--state MLP and a polynomial
NVAR model, which realize parameter-conditioned transitions through different
computational representations. We evaluate both the dynamical structure produced
by recursive execution and its distributional agreement with the reference
system. These comparisons clarify which observed structural properties extend
beyond token-based computation and provide context for the Transformer-specific
parameter-routing analysis.

\subsection{Direct Neural Transition Model}
The MLP evaluates $m_\theta(r,x_t)$ directly from concatenated numerical inputs with three hidden layers of width 64, SiLU activations, biases, and a scalar output, giving 8,577 parameters. It learns the same transition target without a token communication stage.
The paired sampling and optimization protocol is given in Appendix~\ref{app:protocol}.

\subsection{Polynomial Transition Model}
The NVAR features are all monomials $\tilde r^i\tilde x^j$ with $i+j\leq d$, including a constant. We set $\tilde x=2x-1$ and map the training parameter interval affinely to $[-1,1]$ without clipping test parameters. Candidate degrees are $d\in\{3,5,7\}$ and ridge strengths are $\lambda\in\{10^{-10},10^{-8},10^{-6},10^{-4},10^{-2}\}$. We fit
\begin{equation}
 w=(\Phi^\top\Phi/N+\lambda I)^{-1}\Phi^\top y/N.
\end{equation}
The fitting data use 64 batches from the same stream family as the neural models. Independent validation uses 16 batches with seed increased by 10,000. Selection minimizes ID validation MSE. Each logistic seed selects $(d,\lambda)=(3,10^{-10})$ and each sine seed selects $(7,10^{-10})$. Fitting and rollout use float64, with float32-generated training targets. This current-state polynomial model follows the NVAR representation principle; it is not a reproduction of the delayed features and parameter-channel construction of parameter-aware NG-RC.

\subsection{Distributional Comparisons}
Table~\ref{tab:fidelity} reports the 30 fixed-grid comparisons as means and sample standard deviations over three training seeds, and Table~\ref{tab:allseeds} lists every model--context--seed combination. Error bars use the sample standard deviation of three independently trained models; trajectory samples are not treated as independent training replications.

\begin{table}[t]
\small
\centering
\caption{\textbf{OOD distributional fidelity.} Uniform-grid $W_1$ in units of $10^{-3}$, mean $\pm$ sample SD over three training seeds. $C$ counts state tokens, excluding the parameter token. The polynomial NVAR model uses a separate, smaller fitting budget.}
\label{tab:fidelity}
\vspace{-1ex}
\begin{tabular}{lcc}
\toprule
Model & Logistic & Sine\\
\midrule
TF, $C=1$ & $27.86\pm6.76$ & $34.42\pm6.20$\\
TF, $C=4$ & $34.02\pm16.32$ & $33.29\pm16.39$\\
TF, $C=255$ & $129.77\pm81.24$ & $82.32\pm36.12$\\
MLP & $22.97\pm1.53$ & $46.23\pm6.66$\\
Polynomial NVAR & $2.15\pm0.13$ & $2.24\pm0.16$\\
\bottomrule
\end{tabular}
\end{table}

Finite trajectory sampling contributes to these distances. Repeating the reference calculation with different initial states gives mean OOD distances of 0.00210 on logistic and 0.00224 on sine, and the polynomial errors are on this scale. A separate reference-versus-reference calculation uses 16 independent initial-state ensembles under the same sampling budget, yielding mean OOD $W_1$ of 0.002104 (SD 0.000092) for logistic and 0.002239 (SD 0.000188) for sine. This quantifies sampling variability under the evaluation budget; it is not a strict lower bound on model error.

The scalar $W_1$ ordering measures a different property from the cascade: the MLP leads on logistic and the Transformer on sine, and the degree-3 polynomial reaches the sampling floor on a family whose update it can express exactly. Reproducing the visited-state distribution and reproducing the cascade are therefore separate achievements, as the intervention results of Appendix~\ref{app:interventions} and the class diagnostics of Appendix~\ref{app:universality} make concrete.

\subsection{Structural Comparisons}
\label{app:structural-comparisons}
The same comparison at the level of verified flips is reported here in full. The MLP's cascade reaches the same order, with a final finite-order ratio of $4.668296$ against $4.669132$ for the reference map and $4.668686$ for the Transformer, and its flip parameters are displaced in the opposite direction: the seventh flip lies at $3.586416$ for the MLP against $3.510275$ for the Transformer and $3.569891$ for the reference. Both architectures therefore reproduce the quadratic-cascade organization with a seed-specific displacement.
The complete verified roots appear in Table~\ref{tab:allroots}; the across-seed
Transformer alignment analysis is given in Appendix~\ref{app:map-accuracy}.

\subsection{Complete Results across Seeds}

\subsubsection{Trajectories for Alternative Transition Models}
The full-range rendering conventions are shared with Appendix~\ref{app:seed-context}.
MLP and NVAR comparisons use the same four-column seed layout. Logistic and sine each occupy one figure per model.
\begin{figure}[H]
\centering
\setlength{\tabcolsep}{1.2pt}
\renewcommand{\arraystretch}{0.75}
\begin{tabular}{@{}rcccc@{}}
 & \footnotesize Ground truth & \footnotesize Seed 42 & \footnotesize Seed 43 & \footnotesize Seed 44 \\[-2pt]
\raisebox{0.45cm}{\footnotesize MLP} & \includegraphics[width=0.22\textwidth]{img/grid_logistic_truth.pdf} & \includegraphics[width=0.22\textwidth]{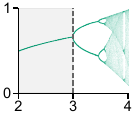} & \includegraphics[width=0.22\textwidth]{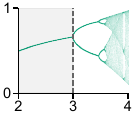} & \includegraphics[width=0.22\textwidth]{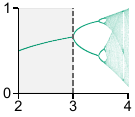} \\[0pt]
\end{tabular}
\caption{\textbf{Logistic MLP comparisons.} Columns are ground truth and seeds 42--44. The MLP directly represents the scalar next-state map.}
\label{fig:atlas-logistic-mlp}
\end{figure}

\begin{figure}[H]
\centering
\setlength{\tabcolsep}{1.2pt}
\renewcommand{\arraystretch}{0.75}
\begin{tabular}{@{}rcccc@{}}
 & \footnotesize Ground truth & \footnotesize Seed 42 & \footnotesize Seed 43 & \footnotesize Seed 44 \\[-2pt]
\raisebox{0.45cm}{\footnotesize NVAR} & \includegraphics[width=0.22\textwidth]{img/grid_logistic_truth.pdf} & \includegraphics[width=0.22\textwidth]{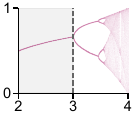} & \includegraphics[width=0.22\textwidth]{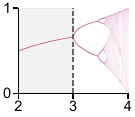} & \includegraphics[width=0.22\textwidth]{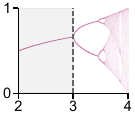} \\[0pt]
\end{tabular}
\caption{\textbf{Logistic NVAR comparisons.} Columns are ground truth and seeds 42--44. NVAR is a parameter-conditioned polynomial model with its separate training protocol.}
\label{fig:atlas-logistic-nvar}
\end{figure}

\begin{figure}[H]
\centering
\setlength{\tabcolsep}{1.2pt}
\renewcommand{\arraystretch}{0.75}
\begin{tabular}{@{}rcccc@{}}
 & \footnotesize Ground truth & \footnotesize Seed 42 & \footnotesize Seed 43 & \footnotesize Seed 44 \\[-2pt]
\raisebox{0.45cm}{\footnotesize MLP} & \includegraphics[width=0.22\textwidth]{img/grid_sine_truth.pdf} & \includegraphics[width=0.22\textwidth]{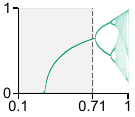} & \includegraphics[width=0.22\textwidth]{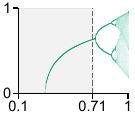} & \includegraphics[width=0.22\textwidth]{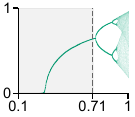} \\[0pt]
\end{tabular}
\caption{\textbf{Sine MLP comparisons.} Columns are ground truth and seeds 42--44. The MLP directly represents the scalar next-state map.}
\label{fig:atlas-sine-mlp}
\end{figure}

\begin{figure}[H]
\centering
\setlength{\tabcolsep}{1.2pt}
\renewcommand{\arraystretch}{0.75}
\begin{tabular}{@{}rcccc@{}}
 & \footnotesize Ground truth & \footnotesize Seed 42 & \footnotesize Seed 43 & \footnotesize Seed 44 \\[-2pt]
\raisebox{0.45cm}{\footnotesize NVAR} & \includegraphics[width=0.22\textwidth]{img/grid_sine_truth.pdf} & \includegraphics[width=0.22\textwidth]{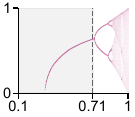} & \includegraphics[width=0.22\textwidth]{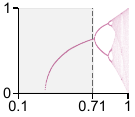} & \includegraphics[width=0.22\textwidth]{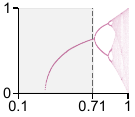} \\[0pt]
\end{tabular}
\caption{\textbf{Sine NVAR comparisons.} Columns are ground truth and seeds 42--44. NVAR is a parameter-conditioned polynomial model with its separate training protocol.}
\label{fig:atlas-sine-nvar}
\end{figure}

\subsubsection{Parameter-Resolved Distributional Errors}
\begin{figure}[h]
\centering
\begin{subfigure}[t]{.4\textwidth}
\centering
\includegraphics[width=\linewidth]{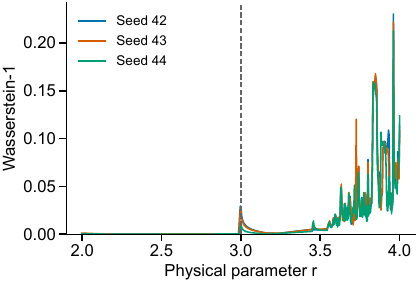}
\caption{Logistic}
\end{subfigure}
\quad
\begin{subfigure}[t]{.4\textwidth}
\centering
\includegraphics[width=\linewidth]{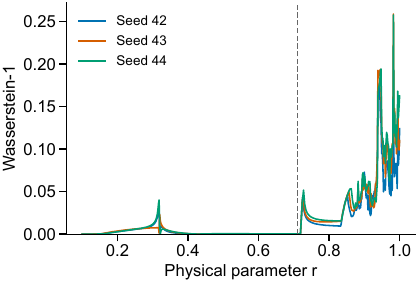}
\caption{Sine}
\end{subfigure}
\vspace{-1ex}
\caption{\textbf{Parameter-resolved fidelity for MLP.} Lines show each seed separately, using every original uniform-grid Wasserstein-1 measurement. The dashed vertical line marks the training boundary. Vertical ranges adapt to the observed error and are labeled; these curves localize differences summarized in Table~\ref{tab:allseeds}.}
\label{fig:error-mlp}
\end{figure}

\begin{figure}[h]
\centering
\begin{subfigure}[t]{.4\textwidth}
\centering
\includegraphics[width=\linewidth]{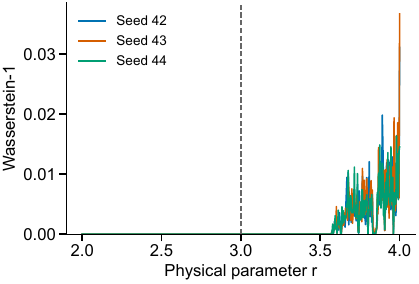}
\caption{Logistic}
\end{subfigure}
\quad
\begin{subfigure}[t]{.4\textwidth}
\centering
\includegraphics[width=\linewidth]{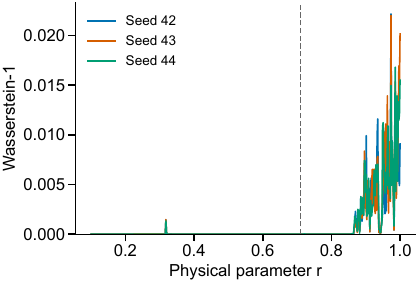}
\caption{Sine}
\end{subfigure}
\vspace{-1ex}
\caption{\textbf{Parameter-resolved fidelity for NVAR.} Lines show each seed separately, using every original uniform-grid Wasserstein-1 measurement. The dashed vertical line marks the training boundary. Vertical ranges adapt to the observed error and are labeled; these curves localize differences summarized in Table~\ref{tab:allseeds}.}
\label{fig:error-nvar}
\end{figure}

\begin{table}[t]\centering
\small
\setlength{\tabcolsep}{4pt}
\caption{All fixed-grid evaluations. ID/OOD columns are mean $W_1$ over the corresponding parameter subset. Period agreement is over accepted reference pairs on the full grid.}\label{tab:allseeds}
\vspace{-1ex}
\begin{tabular}{llrcccr}\toprule
Family & Model / context & Seed & ID $W_1$ & OOD $W_1$ & Period agreement & Coarse flips\\\midrule
Logistic & TF, $C=255$ & 42 & 0.087319 & 0.036792 & 73.6\% & 1\\
Logistic & TF, $C=1$ & 42 & 0.001788 & 0.034595 & 70.2\% & 3\\
Logistic & TF, $C=4$ & 42 & 0.002201 & 0.030639 & 66.4\% & 3\\
Logistic & MLP & 42 & 0.000503 & 0.023623 & 90.0\% & 3\\
Logistic & Polynomial NVAR & 42 & 0.000000 & 0.002137 & 99.8\% & 3\\
Sine & TF, $C=255$ & 42 & 0.004700 & 0.054363 & 78.1\% & 3\\
Sine & TF, $C=1$ & 42 & 0.007186 & 0.041563 & 74.2\% & 3\\
Sine & TF, $C=4$ & 42 & 0.003129 & 0.049795 & 69.9\% & 3\\
Sine & MLP & 42 & 0.002029 & 0.038802 & 73.1\% & 3\\
Sine & Polynomial NVAR & 42 & 0.000005 & 0.002066 & 100.0\% & 3\\
Logistic & TF, $C=255$ & 43 & 0.105557 & 0.165516 & 0.0\% & 1\\
Logistic & TF, $C=1$ & 43 & 0.000497 & 0.027913 & 84.3\% & 3\\
Logistic & TF, $C=4$ & 43 & 0.000487 & 0.019663 & 86.8\% & 3\\
Logistic & MLP & 43 & 0.000440 & 0.024058 & 88.9\% & 3\\
Logistic & Polynomial NVAR & 43 & 0.000000 & 0.002292 & 99.8\% & 3\\
Sine & TF, $C=255$ & 43 & 0.027612 & 0.069479 & 62.4\% & 3\\
Sine & TF, $C=1$ & 43 & 0.003296 & 0.031190 & 78.5\% & 3\\
Sine & TF, $C=4$ & 43 & 0.001037 & 0.033045 & 79.6\% & 3\\
Sine & MLP & 43 & 0.001938 & 0.048183 & 71.0\% & 3\\
Sine & Polynomial NVAR & 43 & 0.000005 & 0.002380 & 100.0\% & 3\\
Logistic & TF, $C=255$ & 44 & 0.069670 & 0.187007 & 0.0\% & 0\\
Logistic & TF, $C=1$ & 44 & 0.000450 & 0.021069 & 81.1\% & 3\\
Logistic & TF, $C=4$ & 44 & 0.000298 & 0.051771 & 67.9\% & 0\\
Logistic & MLP & 44 & 0.000139 & 0.021218 & 87.9\% & 3\\
Logistic & Polynomial NVAR & 44 & 0.000000 & 0.002023 & 99.8\% & 3\\
Sine & TF, $C=255$ & 44 & 0.063694 & 0.123104 & 10.8\% & 3\\
Sine & TF, $C=1$ & 44 & 0.003274 & 0.030495 & 73.1\% & 3\\
Sine & TF, $C=4$ & 44 & 0.000800 & 0.017017 & 84.9\% & 3\\
Sine & MLP & 44 & 0.002315 & 0.051689 & 71.0\% & 3\\
Sine & Polynomial NVAR & 44 & 0.000005 & 0.002289 & 100.0\% & 3\\
\bottomrule\end{tabular}\end{table}

\section{Parameter Computation and Causal Interventions}
\label{app:parameter-analysis}
\label{app:interventions}
\subsection{Representational Analysis}
\label{app:parameter-computation}

The computational question behind our experiments is whether a learned token-space update can approximate the numerical operations of a parameterized dynamical system. We distinguish three questions: whether an architecture can represent such an update, whether optimization discovers it, and whether the learned update preserves dynamics beyond the training region. The analysis below concerns the first question and supplies diagnostics for the other two; it does not assume that successful trajectories establish an exact internal implementation of the governing equation.

\subsubsection{Parameter Dependence in the Scalar Maps}
Both maps in Equation~\ref{eq:maps} have the separable form
\begin{equation}
 F_r(x)=r g(x),\qquad
 g(x)=x(1-x)\ \text{or}\ \sin(\pi x).
 \label{eq:separable-parameter-map}
\end{equation}
Thus a possible computational decomposition is to encode the scalar state, approximate a single nonlinear feature $g$, and multiply it by the parameter. The two scalar families share this parameter dependence even though their state nonlinearities differ. Their success is consequently consistent with an accessible numerical computation, rather than requiring a distinct representation of every attractor. This is a representational hypothesis: the model is not given $g$, and neither training on trajectories nor reproducing a bifurcation diagram alone identifies this decomposition.

The gated feed-forward block has a useful arithmetic capability. For $s(z)=\operatorname{SiLU}(z)=z\sigma(z)$,
\begin{equation}
 s(z)-s(-z)=z.
\end{equation}
Two gated units can therefore form $s(a)b-s(-a)b=ab$ exactly when $a$ and $b$ are available as linear features at the same token. This identity establishes a multiplication primitive in the tested block; it does not show that the trained weights use this construction. Logistic evaluation additionally requires construction of the state polynomial and its product with $r$; sine evaluation requires approximating the trigonometric feature on the state domain. Width, depth, feature preservation through residual updates, and access to the parameter all affect the required implementation.

\subsubsection{Routing Parameters into State Computation}
For a one-state context, our architecture receives two tokens. In the first layer, a state query and parameter key are affine in their respective inputs, so their dot product can contain terms proportional to $rx$, as well as terms linear in $r$ and $x$. The attention weight, however, is a normalized exponential of these logits. A bilinear logit is not itself an exact multiplication output.

More directly, attention can transport parameter features without input-dependent routing. Set the query and key projections to zero, making the state token attend equally to the parameter and state tokens. If the embeddings occupy separate feature subspaces and the value projection selects the parameter subspace, the state update receives a fixed multiple of the parameter feature. The residual path retains the state feature, and an output projection can compensate for the factor of two. This construction requires sufficient feature dimensions and available projection choices, but shows why softmax normalization does not prohibit parameter transport at $C=1$. Subsequent gated computation can combine the two inputs. It also separates \emph{parameter transport through attention} from the stronger claim that \emph{input-dependent attention weights implement the arithmetic}.

The channel alternative embeds the concatenated input through
\begin{equation}
 \widetilde h_t=W_x x_t+W_r r+b.
\end{equation}
With sufficient embedding rank, both inputs are already available in the same residual stream. This removes the need to transmit the parameter across a token boundary. With exactly one state token, attention has a single permitted key, its softmax weight is identically one, and its output is an affine value transformation independent of the query/key logits. Nonlinear computation remains available in the feed-forward blocks. With multiple state tokens, channel-based attention is generally input-dependent; the one-token observation does not apply to the longer-context models studied by \citet{zhai2026transformers}.

These representations provide different routes to the same numerical inputs. A separate token exposes an intervenable communication path, whereas a parameter channel provides direct access at every state position. Neither construction implies a general expressivity advantage. A matched comparison should hold the backbone, parameter budget, training samples, optimizer, and context fixed. Adding a parameter-independent dummy token to the channel model further distinguishes parameter placement from the change in token count. Positional encoding, masking, and the full network must also be considered before making claims about permutation symmetry.

\subsubsection{Diagnostics for Learned Computation}
Equation~\ref{eq:separable-parameter-map} predicts $\partial_rF=g(x)$, $\partial_r^2F=0$, and $\partial_xF=rg'(x)$. These yield function-level tests beyond pointwise prediction: parameter-Jacobian error, spurious parameter curvature, and state-Jacobian error on fixed ID and OOD probes. Derivatives must be measured in physical coordinates, with normalization accounted for; the interior unclipped map is the relevant object when rollout clipping is present. Derivative agreement still does not identify an internal circuit. Routing interventions and matched channel controls address complementary causal questions.

For a scalar period-$k$ orbit, stability depends on the multiplier
\begin{equation}
 \Lambda=\prod_{j=0}^{k-1}\partial_x F_r(x_j).
\end{equation}
Consequently, continued improvement in average one-step loss need not monotonically improve orbit stability, flip parameters, or long-run distributions. An early checkpoint can better preserve these quantities even when a later checkpoint fits observed transitions more closely. We assess this possibility with fixed evaluation probes and matched rollout protocols across training stages. Retrospectively favorable OOD checkpoints characterize learning dynamics; a deployable stopping rule must instead be selected using training-region validation and tested on independent runs. The empirical questions remain whether capacity improves local and multi-step fit, whether derivative structure develops with it, and whether those improvements transfer to autonomous dynamics.

\subsection{Activation Patching}
\begin{figure}[t]
\centering
\begin{subfigure}[t]{.30\textwidth}
\centering
\includegraphics[width=\linewidth]{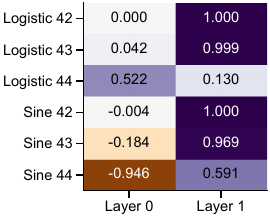}
\caption{Donor-effect transfer}
\end{subfigure}
\hfill
\begin{subfigure}[t]{.30\textwidth}
\centering
\includegraphics[width=\linewidth]{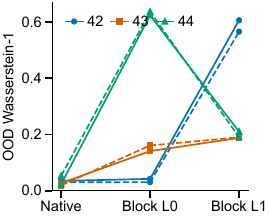}
\caption{Logistic rollouts}
\end{subfigure}
\hfill
\begin{subfigure}[t]{.30\textwidth}
\centering
\includegraphics[width=\linewidth]{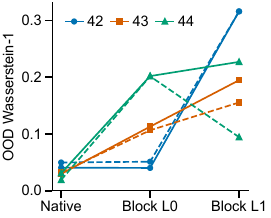}
\caption{Sine rollouts}
\end{subfigure}
\caption{\textbf{Token-level parameter interventions alter closed-loop dynamics.} (a) OOD donor-error reduction after patching final-query attention outputs, for all six Transformers; negative values mean increased donor error. (b,c) OOD $W_1$ after blocking parameter-to-state attention at every step. Lines connect conditions for the same seed; solid/dashed lines indicate one/four state tokens. All 36 conditions share the 129-point grid and trajectory budget.}
\label{fig:mechanism}
\end{figure}

Single-step experiments use 384 ID and 384 OOD parameters per trained Transformer. For each, a 64-state history is generated under the recipient parameter from a random initial condition. The donor parameter is $r'=\min(r+0.25,4)$ for logistic or $r'=\min(r+0.08,1)$ for sine. We keep the recipient history unchanged for the donor pass, so the donor is an explicit counterfactual input rather than a newly generated trajectory. The same physical parameter/history samples are used across seeds within a family.

At each layer, we save the entire final-query attention output before its residual addition. Donor replacement, sham replacement, and a random direction matched to the donor--recipient vector norm are run separately. Table~\ref{tab:patchcontrols} lists the OOD donor-error reductions. Sham replacement has zero effect for every layer and seed. Negative random-control values indicate increased discrepancy from the donor output, rather than evidence of accurate physical prediction.

\begin{table*}[t]\centering\small
\caption{OOD donor-error reduction for complete attention-output patches and equal-norm random controls. Sham reduction is zero throughout. Each entry uses 384 OOD histories from one trained model.}\label{tab:patchcontrols}
\vspace{-1ex}
\begin{tabular}{lrcccc}\toprule
Family & Seed & L0 donor & L1 donor & L0 random & L1 random\\\midrule
Logistic & 42 & 0.000151 & 1.000000 & 0.000022 & -40.688593\\
Logistic & 43 & 0.042383 & 0.999443 & -0.005526 & -0.444243\\
Logistic & 44 & 0.522322 & 0.129718 & -2.708184 & -0.225740\\
Sine & 42 & -0.003656 & 0.999995 & -0.000583 & -0.853852\\
Sine & 43 & -0.184315 & 0.969431 & -0.161228 & -0.314544\\
Sine & 44 & -0.945527 & 0.591043 & -0.155559 & -0.433707\\
\bottomrule\end{tabular}\end{table*}

The layer at which parameter mediation occurs varies across seeds. In logistic seed 44, the first-layer patch yields $R_0=0.522$ and the second yields $R_1=0.130$. Sine seed 44 yields $R_1=0.591$, whereas seeds 42 and 43 yield approximately 1.000 and 0.969. These results support learned, seed-dependent routes of parameter influence. A full attention-output replacement tests mediation at the layer-output level; it does not identify a unique head or feature circuit.

\subsection{Closed-Loop Interventions}
For blocking, all state-query edges to the parameter key are masked in one layer. A single-step rescue restores the recipient attention output at the final query. At the final layer this exactly restores the downstream computation by construction, and is an implementation check. At the first layer, other history positions remain altered, so a final-query rescue need not restore the full later computation. These controls prevent interpreting an algebraic restoration as independent evidence of a unique circuit.

Closed-loop blocking uses all six Transformers, $C\in\{1,4\}$, and native/layer-0-blocked/layer-1-blocked conditions, giving 36 runs on a shared 129-point grid with three initial states, 1,024 discarded steps, and 512 retained steps. Table~\ref{tab:causalall} reports all conditions grouped by model and context. The cascade counts are conservative coarse-grid counts and are not comparable to the adaptive seven-flip result as if they had the same detection power.

\begin{table*}[t]\centering\small
\caption{All closed-loop intervention conditions. Each row contains the three $W_1$ conditions for one model/context. Counts give native/L0-blocked/L1-blocked consecutive flips on the shared 129-point grid.}\label{tab:causalall}
\vspace{-1ex}
\begin{tabular}{lrrcccl}\toprule
Family & Seed & $C$ & Native $W_1$ & Block L0 & Block L1 & Flip counts\\\midrule
Logistic & 42 & 1 & 0.034967 & 0.042183 & 0.606053 & 2/2/0\\
Logistic & 42 & 4 & 0.030323 & 0.030537 & 0.565538 & 2/0/0\\
Logistic & 43 & 1 & 0.027007 & 0.140915 & 0.187707 & 2/0/1\\
Logistic & 43 & 4 & 0.018944 & 0.160885 & 0.190042 & 2/0/1\\
Logistic & 44 & 1 & 0.023330 & 0.623765 & 0.210141 & 2/1/1\\
Logistic & 44 & 4 & 0.052811 & 0.636572 & 0.190833 & 2/0/2\\
Sine & 42 & 1 & 0.040701 & 0.040120 & 0.315712 & 2/2/0\\
Sine & 42 & 4 & 0.049765 & 0.051119 & 0.315712 & 2/2/0\\
Sine & 43 & 1 & 0.029687 & 0.112884 & 0.194603 & 2/2/0\\
Sine & 43 & 4 & 0.032185 & 0.106443 & 0.155418 & 2/2/0\\
Sine & 44 & 1 & 0.030832 & 0.201860 & 0.226855 & 2/1/1\\
Sine & 44 & 4 & 0.019002 & 0.201491 & 0.094361 & 2/0/2\\
\bottomrule\end{tabular}\end{table*}

Blocking also separates finite-grid periodic structure from fidelity: some blocked four-state models retain two detected flips while their distributions change markedly, so a count of detected transitions carries less information than their parameter values and associated trajectories. The 129-point intervention scan is used for matched causal comparisons; the seven-flip result comes from the separate refined scalar analysis.

\subsection{Parameter--Trajectory Correspondence}
The interventions above perturb a trained model. To test whether the parameter route is needed to
build the structure at all, we train the same architecture under the locked protocol with the
parameter label permuted across each batch: every trajectory keeps its dynamics but is labeled by a
parameter drawn independently of it, so the marginal distribution of labels is unchanged while their
pairing with the observed states is destroyed.

\begin{table}[!htbp]
\centering\small
\caption{\textbf{Trained with permuted parameter labels.} Three seeds, identical protocol and
evaluation grid as the main comparison. Periodic agreement is over reference-periodic
parameter--initial-state pairs; the parameter derivative is measured on the out-of-distribution
slice, where the reference mean $|\partial_rF|$ is $0.173$.}
\label{tab:shuffle}
\vspace{-1ex}
\begin{tabular}{lcccc}
\toprule
Seed & ID $W_1$ & OOD $W_1$ & Periodic agreement & Coarse flips\\
\midrule
42 & 0.0417 & 0.1914 & 0.0\% & 0\\
43 & 0.0394 & 0.1912 & 0.0\% & 0\\
44 & 0.0421 & 0.1921 & 0.0\% & 0\\
\midrule
\makecell[l]{Aligned pairs\\(mean of 3)} & 0.0009 & 0.0278 & 78.5\% & 3\\
\bottomrule
\end{tabular}
\end{table}

The cascade disappears completely, and the learned map recovers no systematic parameter
dependence: its mean parameter-derivative mismatch is $0.164$ against a reference scale of $0.173$,
which is as large as the variation the reference family itself carries, and its state critical
exponent remains quadratic ($z=2.03$). Parameter information therefore has to reach the state token
for the extrapolated structure to exist, whereas blocking it after training only degrades a
structure that training had already built.

\section{Supplementary Results and Diagnostics for the Lorenz System}
\label{app:lorenz-results}
\subsection{Joint-Parameter Experimental Setup}
\label{app:joint-lorenz}
The joint experiment varies both $\rho$ and $\sigma$ at $\beta=8/3$ to test periodic extrapolation and recovery of a parameter-dependent regime map. It complements the fixed-$\sigma$ geometry tests in Appendix~\ref{app:lorenz}, using a separately defined training population and evaluation protocol.

\textbf{Numerically Defined Training Support.}
We draw 131,072 scrambled Sobol proposals over $[1,409]\times[0.5,60]$ and integrate each from $(x,y,z)=(0.001,0.001,\rho-1)$ to $t=100$, with RK4 step $0.00125$. A proposal is retained when every coordinate has standard deviation below $10^{-3}$ over the final two time units and its endpoint is within $0.01$ of a nonzero analytic equilibrium. This selects 16,285 parameters by their observed fixed-point destination, without an analytic stability prefilter. The canonical initial condition is shared with the evaluation atlas; training transients are retained and can include cross-wing motion.

A seeded random permutation supplies 8,192 training parameters and 1,024 disjoint validation parameters; the 2,048-trajectory condition uses a nested prefix. Sampling is uniform in parameter area conditional on the numerical fixed-point criterion. All 268 noncritical GT-fixed coarse cells contain training parameters in the larger condition. Each trajectory contributes 512 transitions at $\Delta t=0.01$, spanning 5.12 time units. The smaller and larger datasets therefore contain 1,048,576 and 4,194,304 correlated transition pairs, respectively.

\textbf{Model Architecture and Optimization.}
The 2-layer, 4-head Transformer uses width 64 or 128 (131,840 or 525,824 parameters), a 2-dimensional parameter token, a 3-dimensional state token, and increment prediction with training-only normalization. All four conditions use seed 42, 20,000 AdamW updates, and batch size 512. This fixes transition draws at 10.24 million per condition, corresponding to approximately 9.77 and 2.44 passes through the two datasets. The final checkpoint is used throughout; this is a matched-update comparison, with different effective epoch counts.

\subsection{Regime Structure near Conventional Parameters}

\label{app:lorenz-conventional}
The main flow illustration uses the final 20k checkpoint of the 2,048-trajectory, width-128 condition. After inspecting the original domain, we evaluate the complete $25\times25$ grid $\rho\in[10,50]$, $\sigma\in[5,20]$, including the conventional vicinity of $(28,10)$. The parameters vary jointly; the training data, normalization, and model are unchanged, and the initial state is $(0.001,0.001,\rho-1)$ throughout. Both this model and the 8,192-trajectory, width-128 condition are evaluated on the same grid, and the former is selected for the main display because its long-time geometry is closer to the reference. The display box and the model were chosen after inspecting the wider-domain outcomes, so the figure is an existence illustration rather than a held-out region benchmark.

\textbf{Evaluation on the Coarse Parameter Grid.}
Reference trajectories are generated using float64 RK4 integration with a step size of $0.0025/3$ and recorded every $0.0025$. The model operates in float32 and advances with a step size of $0.01$. For distributional comparisons, we retain the intervals $[50,100]$ and $[150,200]$ and subsample both windows at a spacing of $0.02$. A cell is classified as fixed if the standard deviation of every coordinate over the final two time units is below $10^{-3}$. All 625 cells are included in the analysis, with no removal, smoothing, or interpolation applied. Although short trajectories in the training data may exhibit transient oscillatory behavior prior to convergence, the training set contains no long-horizon trajectories with non-fixed attractors.

\textbf{Adaptive Boundary Refinement.}
Every $\sigma$ row of the coarse grid contains exactly one fixed-to-nonfixed transition, so the transition is bracketed and then refined locally instead of resampling a global fine grid. Each round places nine interior samples inside the current bracket and keeps the adjacent fixed/nonfixed pair, contracting the bracket tenfold. Three rounds reduce the maximum bracket width to $3.7\times10^{-3}$ in $\rho$; the reported boundary is the bracket midpoint. The two refined boundaries differ by a median absolute displacement of $0.95$ in $\rho$ and a mean of $1.55$, with a maximum of $10.7$ at $\sigma=5$, the steepest part of the curve. These are finite-time outcomes from the specified initial state rather than analytically certified bifurcation points.

\textbf{Regime Agreement and Distributional Fidelity.}
Both models remain finite at all 625 cells through $t=200$. The displayed model agrees with the reference fixed/nonfixed classification in 593/625 cells and reproduces repeated wing switching, at least three $x$-sign changes per window, in 362/375 reference cells that switch. Its label is unchanged between the two windows in 600/625 cells. Marginal $W_1$ is computed from the equally spaced physical tails, normalized coordinatewise by the common 8,192-trajectory, width-64 training standard deviations, then averaged across coordinates. Among the reference nonfixed cells the displayed model reaches mean/median $W_1$ of $0.08947/0.06861$ and mean normalized standard-deviation error $0.02270$. The 8,192-trajectory, width-128 model has better fixed/nonfixed agreement (600/625) and switching coverage (369/375) but larger distributional error ($0.14725/0.14473$) and amplitude error ($0.04183$); both complete comparisons are retained.

\textbf{Phase-Portrait Visualization.}
Figure~\ref{fig:lorenz-extension} uses a regular $5\times5$ grid of interior cell centers, $\rho\in\{14,22,30,38,46\}$ and $\sigma\in\{6.5,9.5,12.5,15.5,18.5\}$, chosen independently of the outcomes. Two initial states are released at every pair, $(18,18,\rho-1+6)$ and $(-18,-18,\rho-1-6)$, identical in both panels; the second is deliberately far from the equilibrium so that fixed-state transients remain visible. Trajectories are drawn from $t=0$ for 20 time units, which covers the visual settling of the fixed-state cells. The view is an axonometric rotation of the $(x,y,z)$ state, azimuth $35^\circ$ and elevation $22^\circ$, so that the convergence spirals and the two wings are legible in the same projection. Every glyph is translated so that its own bounding-box center sits on the parameter pair, which makes the two panels directly comparable cell by cell but removes absolute state offsets within a cell. All glyphs share one physical scale, $\rho$ displacement $0.0799$ per state unit, with no per-cell normalization, so amplitudes remain comparable across the panel.

\subsection{Lyapunov Diagnostics}
\label{app:flow-lyapunov}
The flow comparisons of Section~\ref{sec:lorenz-extension} report marginal distributions,
fixed-state agreement and switching. None of those quantities certifies chaos, and the paper's own
detector vocabulary treats an unresolved label as unresolved rather than as chaos. We therefore add
a direct dynamical quantity: the largest Lyapunov exponent of the frozen model and of the reference,
obtained by tangent-vector propagation of the one-step map with renormalization after every step
($\Delta t=0.01$, 3,000 steps after 2,000 burn-in steps, float64).

The reference propagates the RK4 step map of the Lorenz system with a central-difference Jacobian;
the model propagates the increment map
$X \mapsto X + \mathrm{scale}\cdot\mathrm{factor}\cdot m_\theta(p_{\mathrm{norm}},(X-\mathrm{center})/\mathrm{scale})$
with the Jacobian taken by autograd, using the training-only normalization recorded with the
checkpoint. As a self-check, the estimator returns $\lambda=0.9046$ at the classical parameter set
$(\rho,\sigma)=(28,10)$, against the literature value $0.906$.

\begin{table}[t]
\centering\small
\caption{\textbf{Largest Lyapunov exponent} of the reference Lorenz flow and of the frozen
2,048-trajectory, width-128 model of Appendix~\ref{app:joint-lorenz}. Both columns use the
same estimator. Positive values indicate exponential divergence on the attractor; negative values
correspond to fixed-point or periodic destinations.}
\label{tab:lyapunov}
\vspace{-1ex}
\begin{tabular}{lcc}
\toprule
$(\rho,\sigma)$ & Reference $\lambda$ & Model $\lambda$\\
\midrule
28, 10 & 0.9046 & 1.0404\\
32, 10 & 0.9431 & 1.0257\\
30, 10 & 0.9919 & 0.8687\\
26, 10 & 0.9001 & 0.8766\\
28, 9.375 & 0.8916 & 0.8496\\
28, 11.25 & 0.8111 & 0.9837\\
35, 12 & 0.9750 & 1.1131\\
45, 15 & 1.2560 & 1.2582\\
25, 6 & $-0.1062$ & $-0.1683$\\
20, 8 & $-0.1596$ & $-0.0110$\\
\midrule
Mean & 0.7408 & 0.7837\\
\bottomrule
\end{tabular}
\end{table}

The model and the reference agree on the sign of the exponent at all ten parameter pairs and on its
order of magnitude, with individual discrepancies up to a few tenths. The two parameters inside the
fixed-state region are correctly predicted to be non-chaotic. This supports the use of ``chaos'' for
the flow extension in the sense that the frozen model, not only the reference, has a positive
exponential rate at the conventional chaotic parameters, while leaving open whether its chaotic
attractor has the same geometry or dimension as the reference.

\subsection{Regime Recovery over the Extended Parameter Domain}
\label{app:lorenz-extended}
We evaluate every cell of the common $48\times48$ grid for 100 time units, discarding the first 50 for section recurrence. The 48 critical cells at $\rho=1$ remain in the finite-rollout denominator but are excluded from class scores. Reference outcomes define fixed-region membership; all nonfixed reference classes are OOD. Divergent or unresolved model outcomes remain in the scored population.

\textbf{Calibration of Poincar\'e Recurrence Diagnostics.}
Upward crossings of $z=\rho-1$ use degree-seven Lagrange interpolation through eight observations and 24 bisection steps. Signed $(x,y)$ return points are tested for minimal recurrence lag from 1 to 32, requiring at least 40 crossings and four cycles. The 95th-percentile recurrence-distance threshold of $0.1$ is the smallest tested value recovering all 358 P2 and 139 P4 reference cells at model observation resolution. Reference P8 recovery is 19/22; all 22 remain in the primary denominator. Calibration uses a finer reference observer at $\Delta t=0.0025$. An aperiodic label records absence of accepted recurrence; chaos is certified separately by the Lyapunov measurement of Appendix~\ref{app:flow-lyapunov}.

\begin{table}[t]
\centering\small
\caption{Joint Lorenz evaluation on a $48\times48$ parameter grid after training on numerically certified fixed-point parameters. Entries count matching labels against the saved reference; denominators exclude the critical $\rho=1$ line. Finite counts include all cells. All arms use seed 42 and 20,000 updates.}
\label{tab:joint-lorenz}
\small
\vspace{-1ex}
\begin{tabular}{rrr|rrrr|r}
\toprule
$N$ & Width & ID MSE & Fixed/268 & P2/358 & P4/139 & P8/22 & Finite/2304 \\
\midrule
2048 & 64 & 1.56e-06 & 249 & 1 & 2 & 0 & 2155 \\
2048 & 128 & 1.05e-06 & 266 & 9 & 3 & 0 & 2225 \\
8192 & 64 & 1.74e-06 & 227 & 149 & 6 & 0 & 1945 \\
8192 & 128 & 5.96e-07 & 259 & 2 & 5 & 0 & 1173 \\
\bottomrule
\end{tabular}
\end{table}

\textbf{Effects of Dataset Size and Model Capacity.}
Table~\ref{tab:joint-lorenz} separates fixed-region agreement, periodic-regime agreement, and finite-rollout coverage. Increasing data raises P2 agreement from 1 to 149 cells at width 64, but none of the four conditions recovers the full P2/P4/P8 structure. Width 128 attains lower within-condition validation MSE while providing less reliable full-grid stability in the larger-data condition. These outcomes distinguish fitting the training transitions from recovering the joint regime map; they do not identify a unique cause of the extrapolation errors. The single model seed per condition supports these controlled comparisons, without estimating variability over training seeds.

The archived numerical audit also compares 80 saved local transitions against independent DOP853 solves, with maximum absolute discrepancy $2.46\times10^{-5}$. A sensitive long transient can separate pointwise under different integrators despite a matching eventual fixed-point destination. The local-target check and the finite-time destination criterion are therefore recorded separately.

\subsection{Single-Parameter Training-Region Comparisons}
\label{app:lorenz}

\textbf{System Definition and Training Support.}
We use
\begin{equation}
\dot x=10(y-x),\qquad \dot y=\rho x-y-xz,\qquad \dot z=xy-\tfrac83z.
\end{equation}
For these fixed coefficients, the origin loses stability at $\rho=1$. A homoclinic explosion occurs near $13.9265$, producing a chaotic saddle and potentially long chaotic transients while the two nonzero equilibria remain stable. Near $24.0579$, a chaotic attractor appears and coexists with the stable equilibria until their Hopf bifurcation at $470/19\approx24.7368$~\citep{doedel2011preturbulence}. We use these numerical boundaries to define three sampling regions:
\begin{equation}
I_0=(0,1),\qquad I_1=(1,13.9265),\qquad I_2=(13.9265,24.0579).
\end{equation}
The main mixed condition allocates 171, 171, and 170 training trajectories to these intervals, respectively, sampling uniformly within each interval. Independent validation uses 43, 43, and 42 trajectories. Initial coordinates are sampled uniformly from $[-10,10]\times[-10,10]\times[0,30]$. Actual training parameters range from $0.0018139074$ to $23.8602887306$; $24.0579$ is the upper sampling boundary, not the largest realized sample. No basin filtering or convergence-based rejection is applied. The short observations may contain rotations and cross-wing motion. An additional 100-unit oracle continuation is diagnostic only and is not used for learning; finite-time nonconvergence is not evidence of an asymptotic chaotic attractor.

\textbf{Model Architecture and Training Protocol.}
Each trajectory contributes 256 transitions at $\Delta t=0.02$, giving 131,072 training pairs. The oracle uses float64 RK4 with four substeps of size $0.005$. The Transformer has two layers, hidden width 64, four heads, and feed-forward width 256, with one state token plus a parameter token. It predicts a normalized increment added to the current state; the physical vector field is not supplied to the model. State means, state standard deviations, and increment standard deviations are estimated from the training set alone; the parameter input is $(\rho-12.5)/7.5$, a fixed affine encoding across conditions. Training uses seed 42, 10,000 AdamW updates, batch size 128, weight decay $10^{-5}$, gradient clipping at 1, 500 warm-up steps, and cosine decay from $5\times10^{-4}$ to $5\times10^{-5}$. The data seed is 20260917 and the mixed-region assignment seed is 20260919. The final checkpoint is used throughout, without selecting weights using OOD trajectories.

\textbf{Evaluation across Training Regions.}
The primary OOD parameters are $28$ and $32$. At each parameter we evaluate four common positive initial states for 100 time units and score the tail $[50,100]$. Table~\ref{tab:lorenz-regions} holds the architecture, model seed, total trajectory count, initial-state draws, and optimization budget fixed while varying parameter support. Normalization statistics are estimated separately in each condition; scores are converted to shared physical units before comparison. Specifically, marginal Wasserstein distances are divided by the state standard deviations of the original broad $\rho\in[5,20]$ training dataset, averaged over three state coordinates and then over four initial states. This reference dataset defines the metric scale only; it does not contribute training samples to the new conditions. Repeated switching denotes observed changes of the sign of $x$ in the retained window and does not alone certify a positive Lyapunov exponent or identify a parameter-region boundary.

\begin{table}[t]
\centering
\small
\caption{\textbf{Training-region controls at a fixed final checkpoint.} One model seed per condition; 512 training trajectories each. Mixed uses balanced counts over $I_0,I_1,I_2$. Switching counts trajectories across the two OOD parameters and four initial states, not independent model seeds.}
\label{tab:lorenz-regions}
\vspace{-1ex}
\begin{tabular}{lccc}
\toprule
Training support & $W_1(28)$ & $W_1(32)$ & Repeated switching\\
\midrule
$I_0$ & 2.3415 & 2.7005 & 0/8\\
$I_1$ & 1.7427 & 1.9778 & 0/8\\
$I_2$ & 0.1973 & 0.2857 & 8/8\\
Mixed & 0.1022 & 0.2507 & 8/8\\
\bottomrule
\end{tabular}
\end{table}
The mixed and $I_2$ conditions reproduce sustained double-wing motion in these tests. The $I_0$ and $I_1$ conditions do not do so at the fixed checkpoint, showing dependence on the observation region under this protocol rather than establishing an architectural impossibility.

\begin{figure}[h]
\centering
\begin{subfigure}[t]{.49\textwidth}
\includegraphics[width=\linewidth]{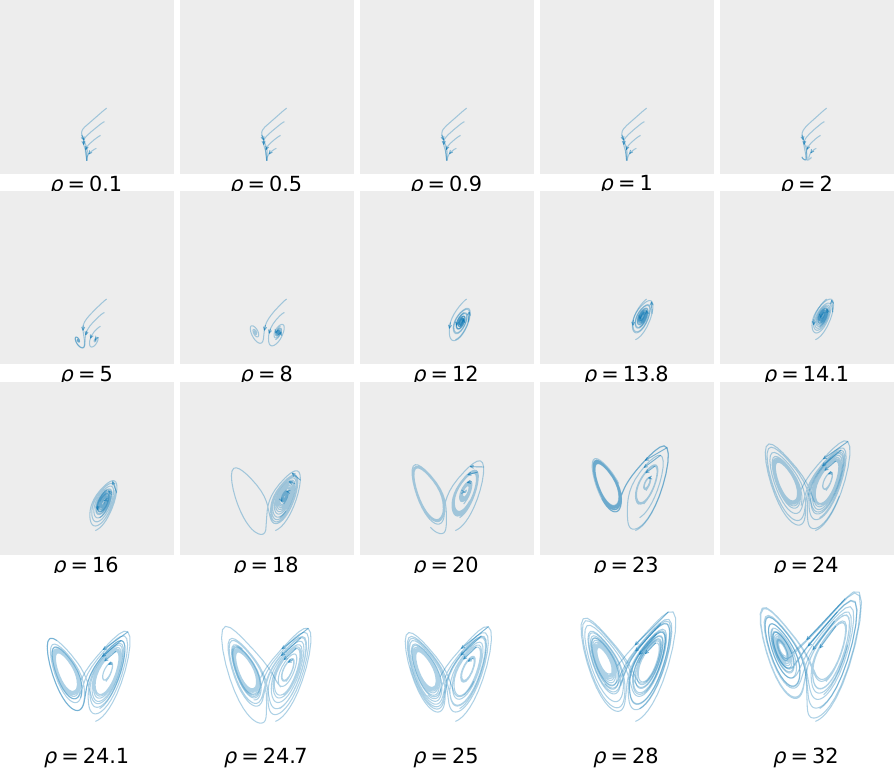}
\caption{Ground truth}
\end{subfigure}
\hfill
\begin{subfigure}[t]{.49\textwidth}
\includegraphics[width=\linewidth]{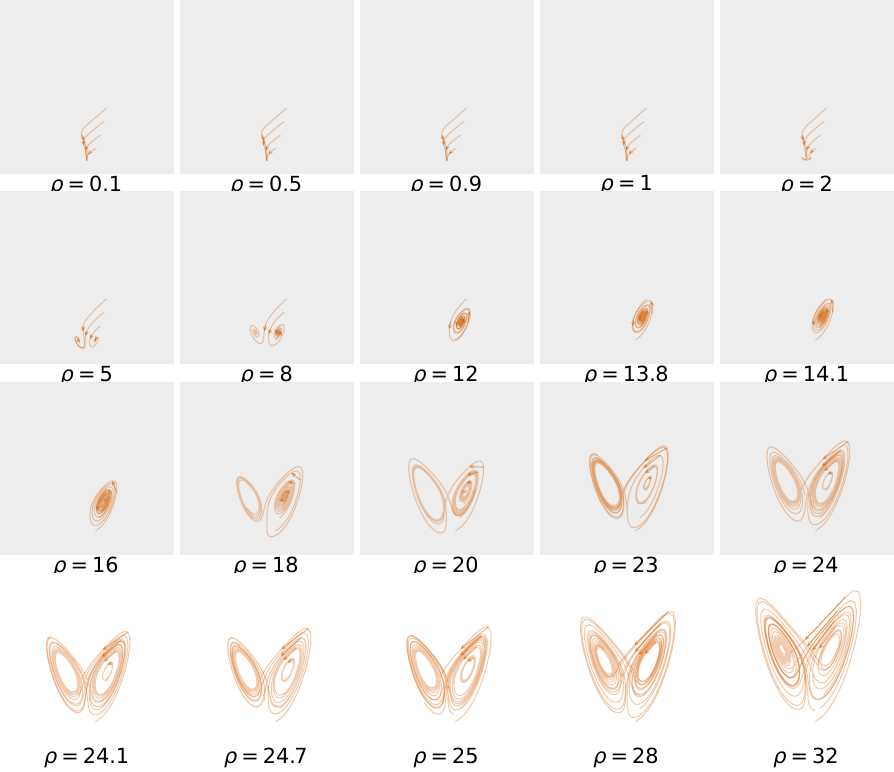}
\caption{Transformer}
\end{subfigure}
\vspace{-1ex}
\caption{\textbf{Lorenz phase portraits across trained and unseen parameter regimes.} Each $5\times4$ atlas overlays four common initial states at each displayed $\rho$, retaining $t\in[0,5.12]$ without burn-in. Curves show the $x$--$z$ projection; arrows indicate motion. Blue: GT; orange: frozen parameter-token Transformer, seed 42, final 10k checkpoint. Gray denotes the training support $0<\rho<24.0579$; white denotes unseen parameters. All cells and both atlases share the same physical scale, with no trajectory normalization. These are test trajectories at new parameter/initial-state combinations. The equal-duration portraits show transients; the separate 100-unit tests quantify sustained double-wing motion.}
\label{fig:lorenz-transients}
\end{figure}

\textbf{Visualization of Transient Trajectories.}
Figure~\ref{fig:lorenz-transients} uses the 20 parameter values listed below. Every panel overlays the same four previously fixed test initial states, approximately $(4.981937,4.941258,9.182749)$, $(6.452753,6.452225,14.340579)$, $(3.399327,3.401437,4.343713)$, and $(7.300586,7.071906,19.377221)$. All panels retain $t\in[0,5.12]$ without burn-in, so this figure compares equal-duration transients rather than final attractors. Arrows follow the displacement from $t=0.3$ to $0.4$. The $x$--$z$ projection has shared limits $[-32.5,32.5]\times[-5,60]$, equal physical aspect ratio, and opacity $0.32$. State axes are suppressed and only $\rho$ is labeled. These are new parameter/initial-state combinations, not training observations. Independent validation remains separate.

\textbf{Parameter Grid and Trajectory Atlas.}
The fixed values, in row-major order, are
\begin{equation}
\begin{gathered}
0.1,\ 0.5,\ 0.9,\ 1,\ 2;\\
5,\ 8,\ 12,\ 13.8,\ 14.1;\\
16,\ 18,\ 20,\ 23,\ 24;\\
24.1,\ 24.7,\ 25,\ 28,\ 32.
\end{gathered}
\end{equation}
This grid spans the origin regime, the pitchfork boundary, the vicinity of the homoclinic transition, the coexistence region, and the standard chaotic probes; it was fixed before computing the new atlas predictions. For mixed training, the first 15 panels are within the sampling support and the final five are OOD. The isolated boundary $\rho=1$ is treated as an in-support interpolation probe, although a continuous sampler has zero probability of observing that exact value. For $I_2$ training, gray panels cover $14.1,16,18,20,23,24$; all remaining displayed values lie outside its training interval. %The finite atlas is a visualization of selected parameters, not a numerical determination of transition values.

In the supplementary atlases below, gray panels overlay 16 trajectories over $[0,5.12]$; white panels show one fixed initial state over $[50,100]$. These supplementary windows deliberately distinguish short-transient fitting from autonomous long-time behavior, unlike the equal-window main figure. These window types differ in horizon and are not comparable as convergence tests. All coordinates retain the same physical scale, with no alignment, centering at equilibria, or amplitude rescaling. Blue denotes GT, orange denotes predictions, and line opacity is 0.35.

\begin{figure}[h]
\centering
\begin{subfigure}[t]{.49\textwidth}
\includegraphics[width=\linewidth]{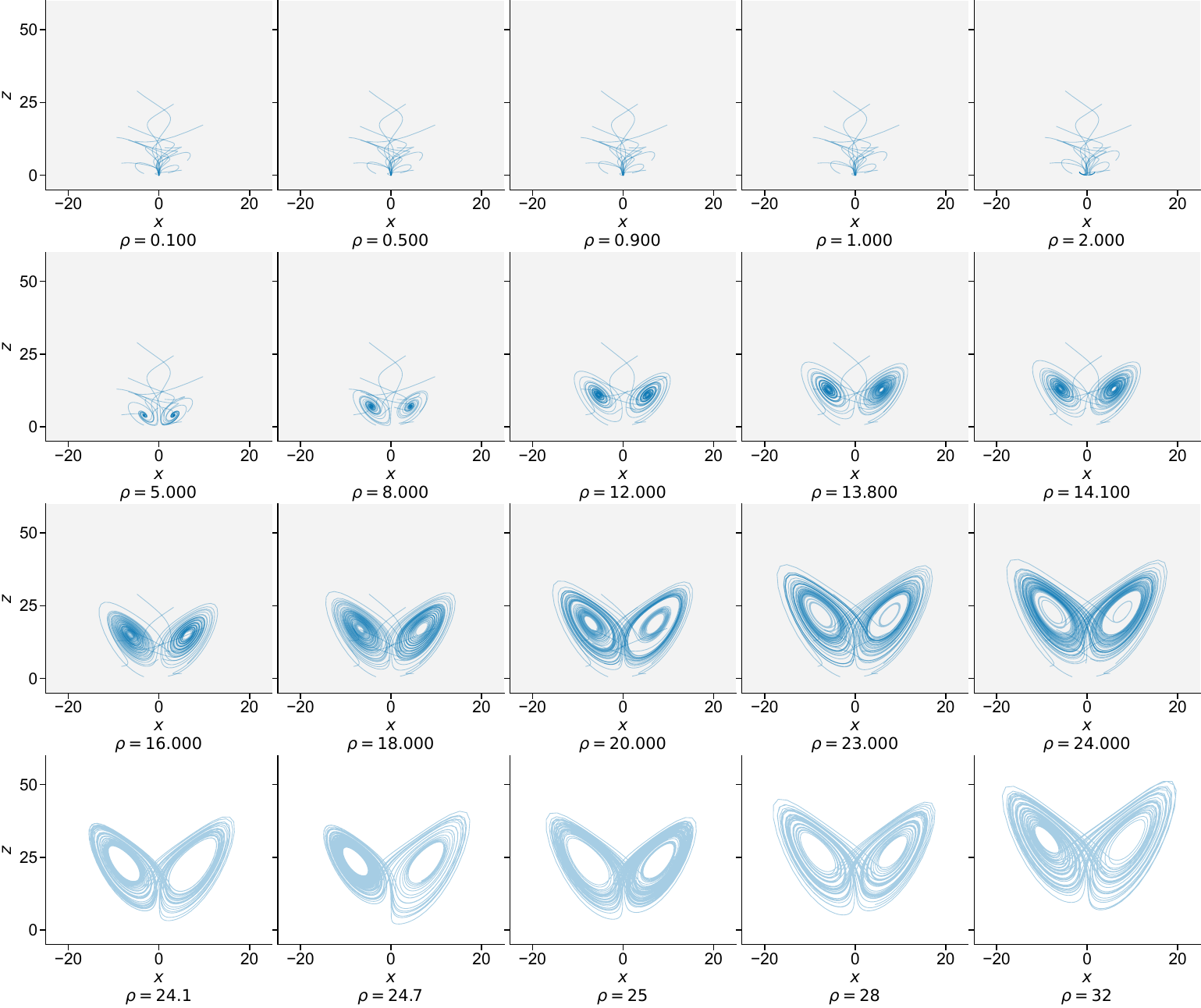}
\caption{Ground truth}
\end{subfigure}
\hfill
\begin{subfigure}[t]{.49\textwidth}
\includegraphics[width=\linewidth]{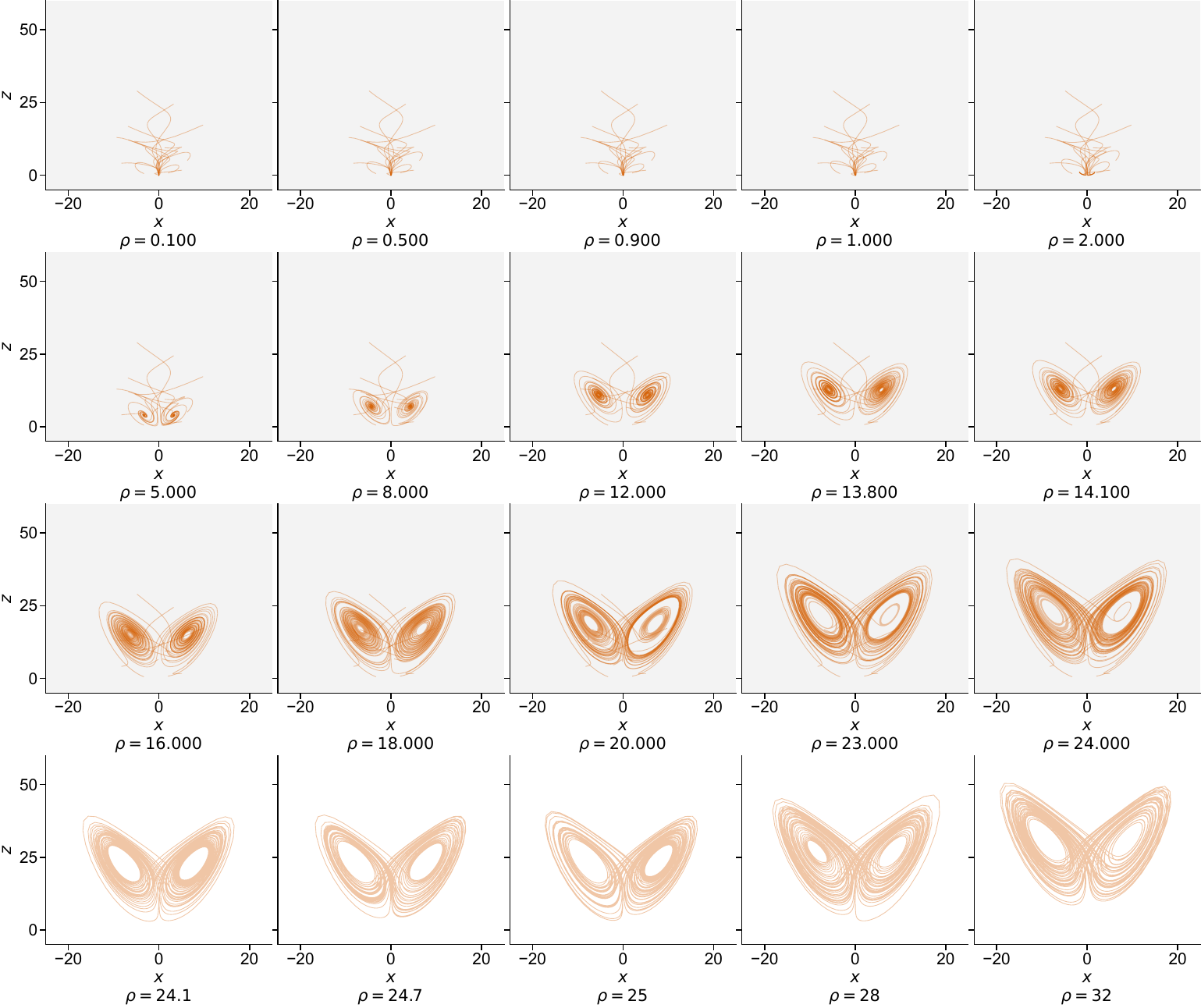}
\caption{Transformer}
\end{subfigure}
\vspace{-1ex}
\caption{\textbf{Lorenz atlas for mixed-region training.}
Twenty parameters are shown in a $5\times4$ layout.
Gray denotes the training support $0<\rho<24.0579$, while white denotes unseen parameters.
Both panels use the same parameter values, initial states, time windows, and axes.
The Transformer result is from the frozen seed-42 model after 10k updates.}
\label{fig:lorenz-mixed-atlas}
\end{figure}

\section{Supplementary Results and Diagnostics for the Generalized Hopf System}
\label{app:generalized-hopf}
\subsection{Experimental Setup and Training Support}
\textbf{System Definition and Bifurcation Boundaries.}
In Cartesian coordinates, the system is
\begin{equation}
\dot x=(\mu_1+\mu_2 r^2-r^4)x-y,\quad
\dot y=x+(\mu_1+\mu_2 r^2-r^4)y,\quad r^2=x^2+y^2.
\end{equation}
The origin changes stability at $\mu_1=0$. Nonzero periodic orbits have squared radii $s_\pm=(\mu_2\pm\sqrt{\mu_2^2+4\mu_1})/2$ when real and positive. For $\mu_2>0$ and $-\mu_2^2/4<\mu_1<0$, the origin and outer cycle are stable, separated by the unstable inner cycle; the cycles meet at $\mu_1=-\mu_2^2/4$~\citep{kuznetsov2004}. These boundaries define the training region before model evaluation. The rectangular sampling limits fix numerical scales and are not additional bifurcations.

\textbf{Data Generation and Model Configuration.}
Parameters are sampled uniformly in area by rejection from $[-1,0]\times[-1,1]$, retaining $\mu_1<-\max(\mu_2,0)^2/4$. Initial states are uniform in a disk of radius $\sqrt{(1+\sqrt5)/2}$, the largest stable-cycle radius in the declared evaluation box $[-1,1]^2$. All 512 training trajectories have strictly decreasing radii; 128 additional trajectories provide independent validation. Each has 256 transitions at $\Delta t=0.05$, spanning 12.8 time units. The oracle integrates the radial equation using ten RK4 substeps per observation and rotates the angle exactly. Against Cartesian DOP853 integration with tolerances $10^{-12},10^{-14}$, the largest one-step discrepancy in 24 independent checks was $2.60\times10^{-9}$.

The two-layer Transformer has width 64, four attention heads, and feed-forward width 256. A linear two-dimensional projection embeds the parameters into one token, separately from the Cartesian state token. Training-only state, parameter, and increment statistics define normalization. The incremental target, AdamW schedule, batch size 128, and 10k-update budget match the single-parameter Lorenz setup in Appendix~\ref{app:lorenz}. The model seed is 42 and data seed is 20260917. The final checkpoint is fixed without OOD selection; its held-out normalized one-step MSE is $2.2243\times10^{-6}$.

\subsection{Attractor Coexistence and Basin Tests}
The nine parameter pairs are $(-0.5,-0.5)$, $(-0.5,1)$, $(-0.1,1)$, $(-0.2,1)$, $(-0.04,0.5)$, $(0.1,-1)$, $(0.1,0)$, $(0.1,1)$, and $(0.5,1)$. Each uses radii $0.1,1.2$ and phases $0,\pi/2$. All 36 rollouts remain finite through $t=300$. Over $t\in[150,300]$, the predicted origin/cycle outcomes match the reference at all nine pairs: 14 trajectories approach points and 22 approach cycles. The operational point threshold is mean radius below $0.03$; cycle labels require mean angular speed within $0.1$ of one and radius standard deviation below $0.05$. Continuous radius and frequency measurements are retained. These are tests of multiple initial conditions, not a quasistatic hysteresis experiment.

\subsection{Parameter Slices and Stability Boundaries}
Two slices fix $\mu_2=-1$ or $1$ and evaluate 41 equally spaced $\mu_1$ values from $-0.5$ to $0.5$, each from radii $0.1,1.2$ at phase zero. All 164 predictions remain finite. Mean absolute radius errors against matched reference windows $t\in[200,400]$ are $0.00894$ and $0.04975$, respectively. At $\mu_2=1$, the large initial state incorrectly collapses at $\mu_1=-0.225$, while small initial states incorrectly approach cycles at $\mu_1=-0.05,-0.025$. Independent fixed-point solves and Jacobians of the learned discrete map locate complex-eigenvalue unit-circle crossings at $\mu_1=0.00720,-0.00652,-0.02872,-0.05615$ for $\mu_2=-1,0,0.5,1$; the reference boundary is zero throughout. Thus the model recovers coexistence but displaces its boundaries. Full unstable-cycle continuation and a dense two-dimensional asymptotic regime classification are not claimed.

\subsection{Trajectory Visualizations}
Figure~\ref{fig:generalized-hopf} uses the 64 cell centers of a uniform $8\times8$ grid over $[-0.5,0.5]\times[-1,1]$. All four initial states are shown at each center for $t\in[0,12.8]$, without burn-in. The glyph transformation is fixed globally, preserving relative amplitudes and the Cartesian aspect ratio. Arrows indicate actual displacements from $t=1$ to $1.4$ on the two outer-start trajectories. Slow convergence near a bifurcation can extend beyond this short display window. Gray marks the exact curved training-support intersection, not a classification inferred from trajectory appearance.

\section{Reproducibility and Artifact Availability}
\label{app:artifacts}
\subsection{Software and Hardware}
Ten neural runs were trained on an NVIDIA RTX 5060 Ti GPU. The two seed-44 MLP runs used CPU execution under the same sampling and optimizer budget. This scheduling choice was made without consulting OOD outcomes; per-run records identify the device. We do not use mixed-device wall-clock times as an efficiency comparison. The implementation used Python 3.13.9, PyTorch 2.9.1 with CUDA 12.8, NumPy 2.3.4, and SciPy 1.16.3.

\subsection{Experiment Records and Integrity Checks}
The experiment archive contains raw rollout arrays, per-parameter diagnostics, checkpoints, configuration records, adaptive-stage tails, and root-verification results. Fixed-grid inputs and reference arrays are identical across compared models. The scalar integrity record covers 12 checkpoints, 30 main evaluations, and 36 closed-loop intervention conditions. Additional flow checks recompute the four joint Lorenz class-count tables, validate the saved training/validation membership, recover all 36 generalized Hopf outcomes, and recompute the mixed-Lorenz switching and marginal distributional distances from saved arrays. The selected $25\times25$ Lorenz confirmation retains all cells and checks representative labels across windows, initial states, model precision, and independent reference integration. Unit tests cover numerical routines and detect missing or changed figure provenance records.

%Every cited figure has a record linking its current asset files to available source data and rendering code, with SHA-256 hashes checked before synchronization. This inventory reconstructs existing file relationships; it does not certify that all figures were rerendered in the latest revision. Byte-identical cached copies additionally identify the overview and flow portrait assets. The validation script checks the saved inventory without silently refreshing it.

\stopcontents[appendices]

\end{document}